\documentclass[sigconf]{acmart}
\acmSubmissionID{473}

\usepackage{booktabs} %
\usepackage{balance}

\usepackage[ruled]{algorithm2e} %
\usepackage{amsfonts} 
\usepackage[export]{adjustbox}
\usepackage{cleveref}
\usepackage{makecell}
\usepackage{multirow}
\usepackage{mwe}
\usepackage{graphbox}
\usepackage{tikz}
\usepackage[normalem]{ulem}

\usetikzlibrary{spy}

\usepackage{xcolor,colortbl}
\usepackage{soul}

\definecolor{GrayTable}{gray}{0.95}
\newcolumntype{a}{>{\columncolor{Gray}}c}

\SetAlFnt{\small}
\SetAlCapFnt{\small}
\SetAlCapNameFnt{\small}
\SetAlCapHSkip{0pt}

\acmJournal{TOG}
\copyrightyear{2023}
\acmYear{2023}
\setcopyright{acmlicensed}\acmConference[SA Conference Papers '23]{SIGGRAPH Asia 2023 Conference Papers}{December 12--15, 2023}{Sydney, NSW, Australia}
\acmBooktitle{SIGGRAPH Asia 2023 Conference Papers (SA Conference Papers '23), December 12--15, 2023, Sydney, NSW, Australia}
\acmPrice{15.00}
\acmDOI{10.1145/3610548.3618180}
\acmISBN{979-8-4007-0315-7/23/12}
\begin{document}
\title{Diffusing Colors: Image Colorization with Text Guided Diffusion}

\author{Nir Zabari}
\orcid{0000-0001-5414-0680}
\affiliation{%
 \institution{Lightricks}
 \city{Jerusalem}
 \country{Israel}}
 \email{nir.zabari@mail.huji.ac.il}

\author{Aharon Azulay}
\orcid{0000-0003-4421-8477}
\affiliation{%
 \institution{Lightricks}
 \city{Jerusalem}
 \country{Israel}}
 \email{aharonchiko@gmail.com}

\author{Alexey Gorkor}
\orcid{0009-0003-4134-3472}
\affiliation{%
 \institution{Lightricks}
 \city{Jerusalem}
 \country{Israel}}
 \email{alexey.gorkor@gmail.com}

\author{Tavi Halperin}
\orcid{0000-0001-9288-5392}
\affiliation{%
 \institution{Lightricks}
 \city{Jerusalem}
 \country{Israel}}
 \email{tavihalperin@gmail.com}

\author{Ohad Fried}
\orcid{0000-0001-7109-4006}
\affiliation{%
 \institution{Reichman University}
 \city{Herzliya}
 \country{Israel}}
 \email{ofried@runi.ac.il}

\renewcommand\shortauthors{Zabari, N. et al}

\def\ShowNotes{}

\begin{abstract}
The colorization of grayscale images is a complex and subjective task with significant challenges. Despite recent progress in employing large-scale datasets with deep neural networks, difficulties with controllability and visual quality persist. To tackle these issues, we present a novel image colorization framework that utilizes image diffusion techniques with granular text prompts.
This integration not only produces colorization outputs that are semantically appropriate but also greatly improves the level of control users have over the colorization process. Our method provides a balance between automation and control, outperforming existing techniques in terms of visual quality and semantic coherence. We leverage a pretrained generative Diffusion Model, and show that we can finetune it for the colorization task without losing its generative power or attention to text prompts.  
Moreover, we present a novel CLIP-based ranking model that evaluates color vividness, enabling automatic selection of the most suitable level of vividness based on the specific scene semantics. Our approach holds potential particularly for color enhancement and historical image colorization.

\end{abstract}



\setlength{\tabcolsep}{1pt}
\newlength{\wwt}
\setlength{\wwt}{0.165\linewidth}
\begin{teaserfigure}
\includegraphics[width=\linewidth]{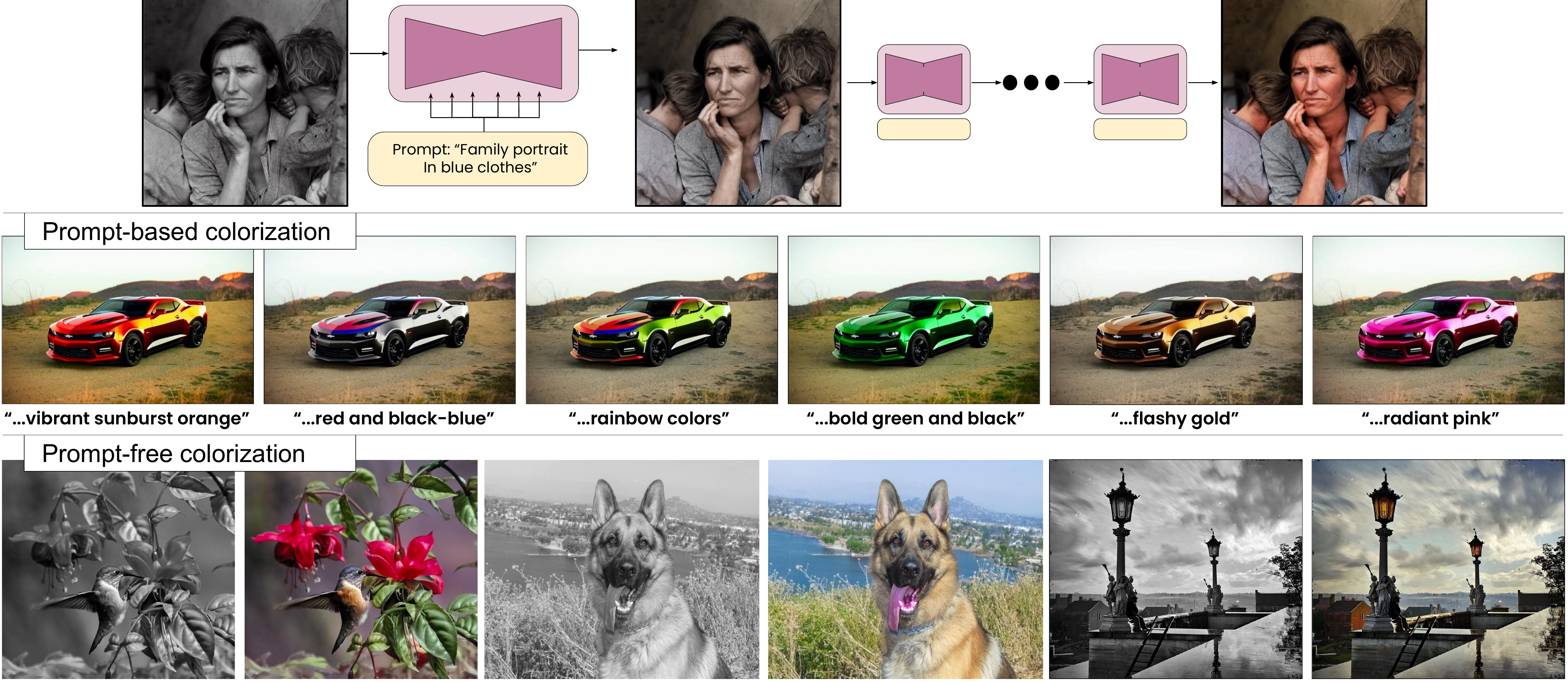}
\caption{\textbf{Diffusion Based Colorization.} We introduce a novel method for image colorization based on diffusion, achieving state of the art results. The first row illustrates our diffusion-based color sampling technique, highlighting the incremental colorization outputs of the iterative process. Next, we exhibit the capability of our method to achieve text-based colorization (second row), which showcases a range of diverse colorization outcomes. Finally, we visualize pairs of grayscale input images alongside the colorized outputs produced by our approach. Image credits (left to right, top to bottom): `Migrant Mother' by Dorothea Lange, Unsplash ©Stefan Rodriguez, Unsplash ©Bryan Hanson, Unsplash ©Sarah Sheedy.}
\label{fig:teaser}
\end{teaserfigure}
\maketitle

\section{Introduction}
\begin{figure*}

\includegraphics[width=\textwidth]{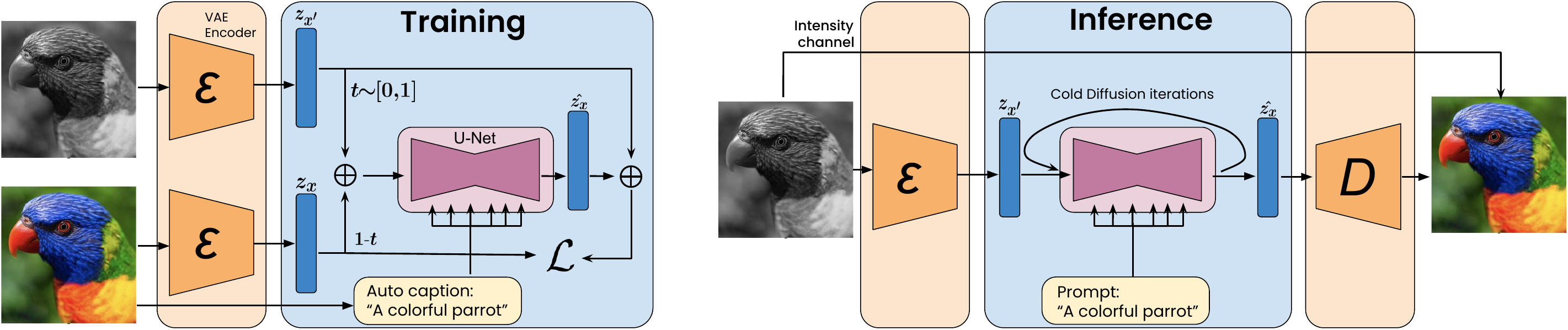}

\caption{
\textbf{Overview.} During training we encode the RGB and grayscale images into the latent space and feed the U-Net with a random convex combination of the two, together with an auto generated caption of the color image, and the timestep $t$. At inference time we encode the input grayscale image as $z_{x}$ and iteratively colorize it to get $\hat{z_{x}}$ which we decode, and combine with the grayscale as a luma channel. Image credit: Unsplash © David Clode.
}
\label{fig:system_overview}
\end{figure*}
Image colorization is a long standing computer graphics problem, where the goal is to add color to a grayscale image. The addition of color can significantly enhance the visual appeal and information content of the image. Current image colorization methods are often trained on large scale color image datasets, where paired grayscale and color images are trivially created by removing the colors from natural images.
However, the colorization task is inherently ill-posed, and each grayscale image has an infinite amount of matching color images. Therefore, naively training a model to colorize an image will suffer from regression to the mean and will result in desaturated images.
One approach to solve this involves color sampling from a projected distribution, such as by using a GAN-based model, which effectively selects a solution from a set of potential modes. However, GANs suffer from the well-known mode collapse problem and from instability in training. 
Another way to get more vivid results is by incorporating user input, for instance in the form of scribbles or text prompts. This reduces the set of potential solutions and can lead to improved results.
However, it comes with the cost of requiring the user to always provide additional inputs, which limits usability. In this paper, we propose a novel image colorization framework that leverages image diffusion techniques and \emph{optional} fine-grained text prompts to produce more vivid and semantically meaningful colorization results, while also being easy to use. Our method enables fine-grained control over the colorization process by providing guidance in the form of a text description of the desired colorization outcome, but also allows for a default parameter-free inference that requires no user interaction. 
Experiments on standard benchmark datasets show that our method outperforms previous methods in terms of colorization quality, visual appeal, and semantic consistency, even in the fully automatic settings. 
To tune the results even further, we propose a colorization ranker that can automatically choose the most suitable level of colorfulness for a given colorized image, resulting in a more natural and aesthetically pleasing outcome.
Finally, an extensive user study demonstrates the  strengths of our method compared to prior art. Due to the high computational resources and time requirements of Diffusion Models, even at inference time, we turned to utilize Latent Diffusion. This technique operates on a lower dimensional latent space where we encode images using a pretrained VAE. Fortunately, the VAE was trained on a large and diverse corpus of images, including grayscale ones, making it suitable to compute latent encodings both for the input and the GT images, allowing us to train exclusively in the latent space. This approach enables us to efficiently fine-tune a Stable Diffusion model for colorization on a relatively small set of images and their corresponding auto-generated captions.
In our experiments, we found that sampling points along the high dimensional line between an RGB and the corresponding grayscale images in latent space is equivalent to varying the levels of colorfulness and saturation of the image in pixel space. We used this to adapt the Cold Diffusion \cite{ColdDiffusionBansal2022ColdDI} framework, by defining a degradation operator as the conversion from RGB to grayscale within this latent space.
Differently from image generation models operating in lower dimensional space, which often suffer from high-frequency decoding artifacts, our colorization approach has an advantage in mitigating the impact of degradation caused by the VAE decoder on the quality of results. This is because we replace the luma channel (in CIE-Lab space) of the decoded image from that of the input, thereby ignoring the impact of luma decoding artifacts.

Moreover, since chroma channels in natural images tend to contain lower frequencies, they are more resistant to decoding artifacts.

Our proposed method has the potential to be used for color enhancement and historical image colorization, where old grayscale images can be restored by adding color information. However, a major challenge in this domain is the absence of large-scale datasets with ground-truth color information, making it difficult to train models with high accuracy.

Unlike modern images where color can easily be removed to create paired datasets, historical images often lack true color reference due to color fading over time. However, our method provides a way to overcome this limitation by using text prompts as a control knob that can regulate the colors of the result. While text prompts cannot solve the data scarcity issue directly, they offer a alternative mechanism for controlling the output of the colorization process. This approach helps bypass the need for exhaustive ground-truth data by allowing the model to be guided by semantic cues, therefore bridging the gap where data is insufficient. 
The control provided by these text prompts can be adjusted to match the desired colorization style, improving visual appeal and semantic coherence. Furthermore, due to the iterative nature of Diffusion Models, our model is able to utilize images with partially vanished colors and enhance them while keeping the original tones. This makes our method a valuable tool for those working with historical images, as it can produce controllable and vivid colorization results, even for these out-of-distribution photographs.
To summarize, our contributions are as follows:
\begin{itemize}
\item A state of the art colorization method based on Diffusion Models.
\item Superior control through textual cues for color guidance and a CLIP-based ranking model for optimal saturation level.
\item An analysis of the color properties of the VAE latent space.
\item A demonstration of the effectiveness of Cold Diffusion within latent spaces.
\end{itemize}

\section{Related Work}
Colorization methods generally fall into two categories: fully automatic colorization methods and user-guided methods. We will review some notable examples of both categories. Subsequently, we will briefly touch upon the recent advancements in Diffusion Models that are relevant to our task.

\subsection{Automatic Colorization}
Automatic Image Colorization involves adding color to grayscale images without user interaction or other hints. Due to the ill-posed nature of the problem, fully automatic methods that use vanilla regression losses suffer from desaturated outputs, or introduce other artifacts, as they ``regress to the mean''. 
This problem has been addressed using various approaches such as distribution prediction \cite{ML_Color_Larsson2016LearningRF, DISCOXia2022DisentangledIC}, novel learning-based paradigms, e.g. CIColor \cite{CICCOLOR_Zhang2016ColorfulIC}, and most recently with generative models such as GANs \cite{CHROMAGAN_Vitoria2019ChromaGANAP, BIG_COLOR_Kim2022BigColorCU, wu2021towards} and diffusion models \cite{saharia2022palette}. 

\begin{figure}
\setlength{\tabcolsep}{1pt}

\newlength{\wwb}
\setlength{\wwb}{0.5\linewidth}

\begin{tabular}{cc}
  \centering

\includegraphics[width=\wwb, height=0.7\textheight,keepaspectratio]{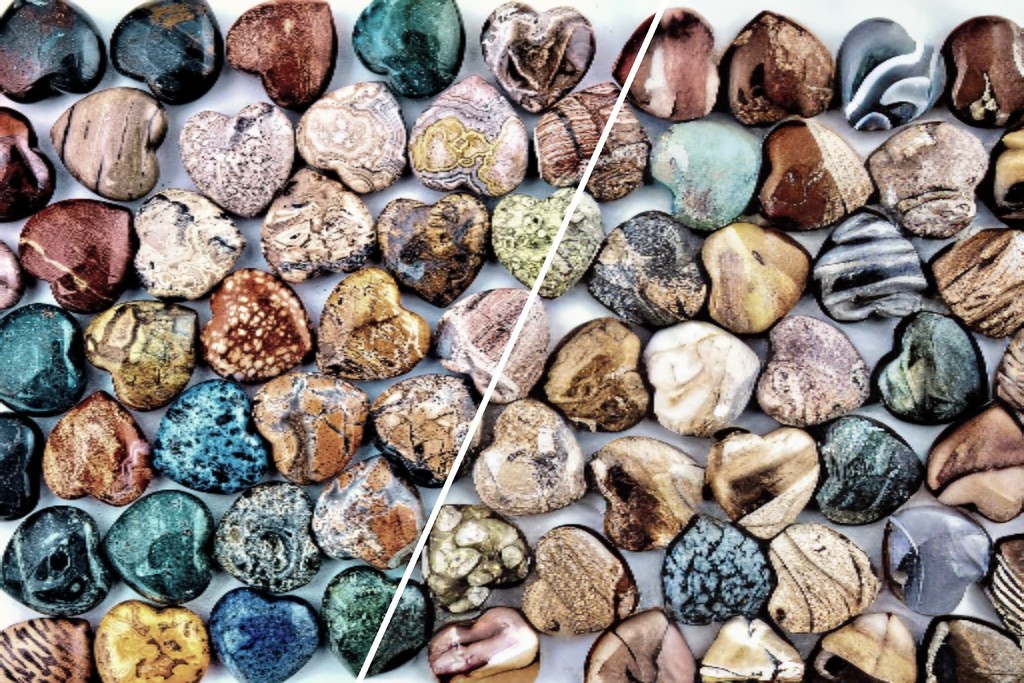} &
\includegraphics[width=\wwb,keepaspectratio]{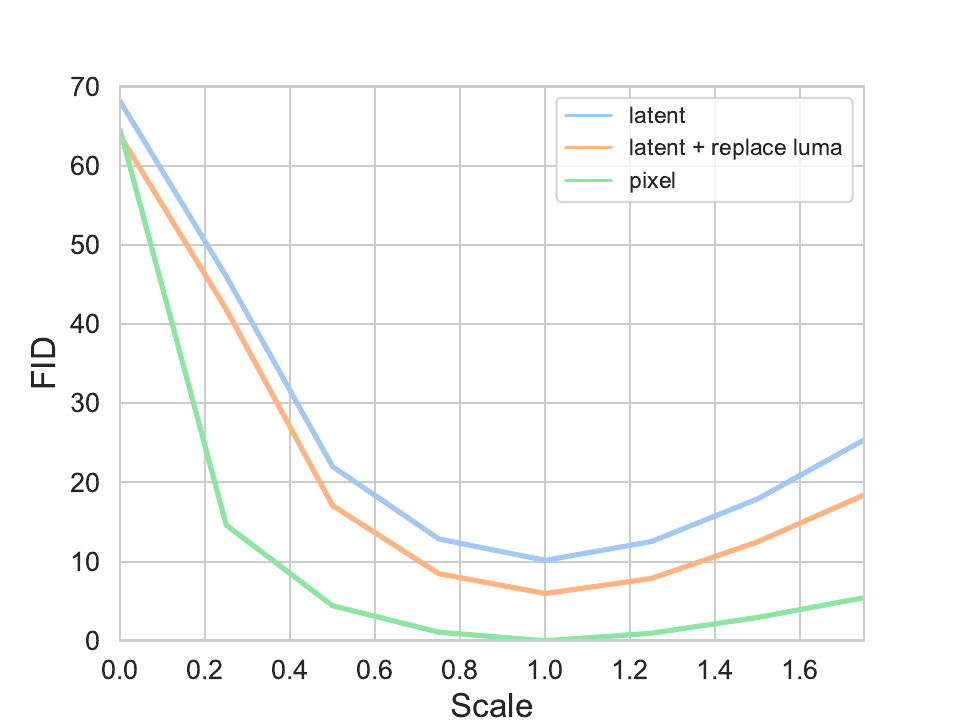} \\

\includegraphics[width=\wwb, height=0.7\textheight,keepaspectratio]{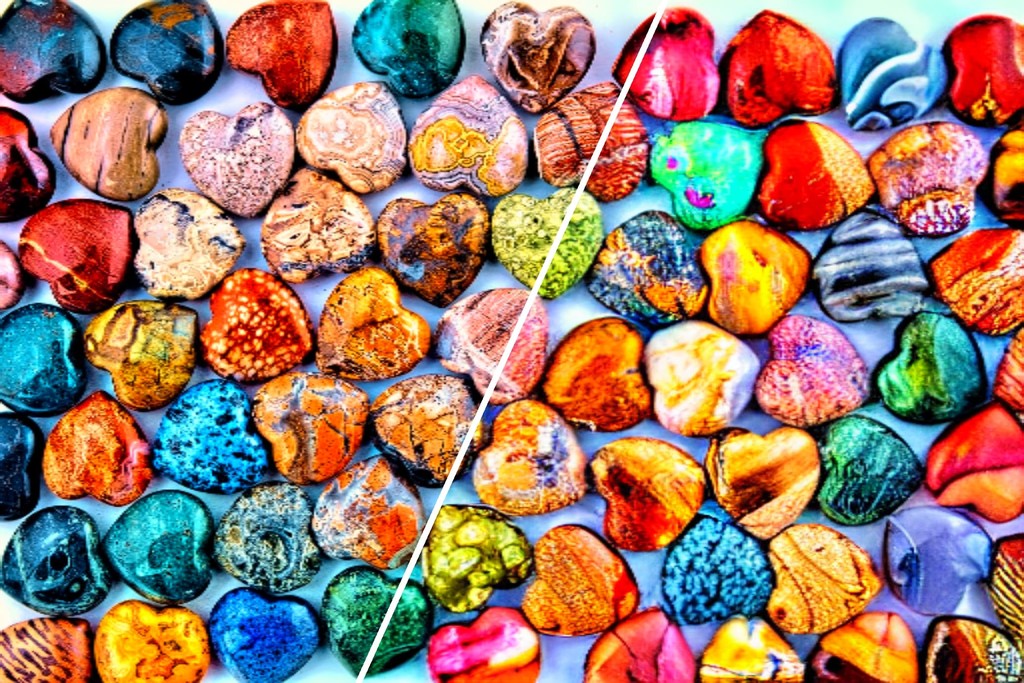} &
\includegraphics[width=\wwb, height=1\textheight,keepaspectratio]{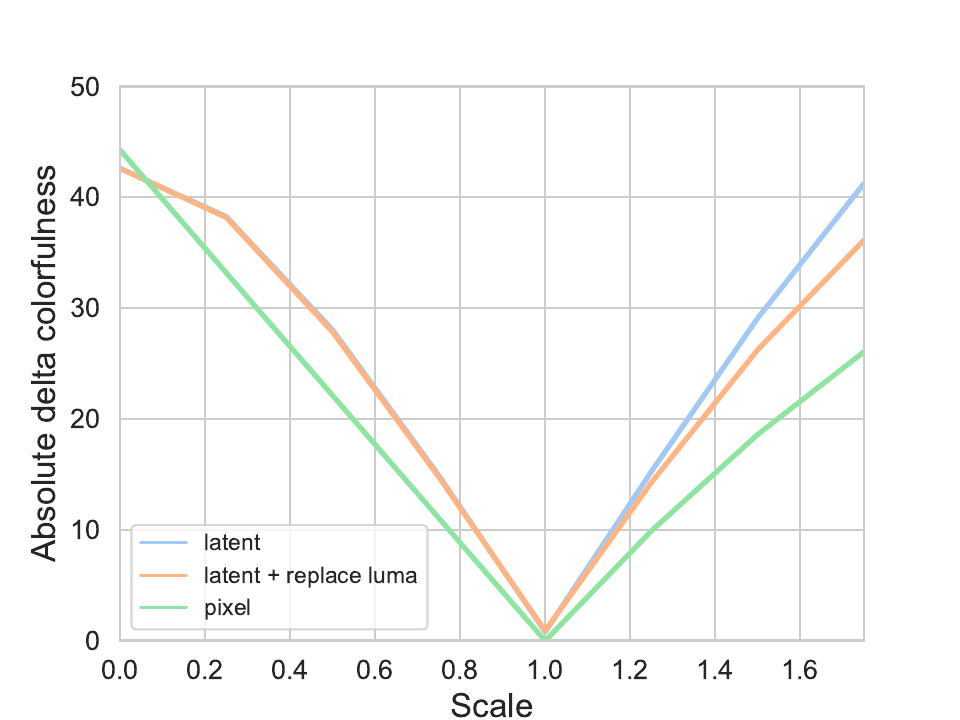} \\

\end{tabular}

\caption{
\textbf{Latent space motivation.} 
We compare scaling colors in image space and the VAE latent space. The images are generated by scaling the colors to $0.5x + 0.5x'$ (top) and $1.5x - 0.5x'$ (bottom), where $x$ is the input image and $x'$ is its grayscale counterpart. The images are split such that the left and right parts exhibit scaling in pixel and latent space, respectively. We observe that scaling color in latent space behaves similarly to doing so in pixel space.
We performed a quantitative comparison (right column) between images color-scaled to different levels, calculated over 500 images from the ImageNet train set.
Again, we observed only small difference between scaling in pixel and latent space, with the standard FID and $\Delta$-colorfulness metrics. The small gap can be further reduced by replacing the luma channel of the images decoded from scaled latents, with the luma of the original image, thus reducing artifacts introduced by the VAE decoder. Image credit: Unsplash ©jccards.
}
\label{fig:latent_space_viz}
\end{figure}
In order to facilitate control over the colorization results, some of the automatic methods allow interactive colorization by incorporating user input, usually given as clicks or color scribbles, as demonstrated by Levin et al. \shortcite{Levin2004ColorizationUO} and UGColor \cite{UGCOLOR_Zhang2017RealtimeUI}. They propagate that input to the rest of the image using classic methods \cite{Levin2004ColorizationUO} or with learned models \cite{UGCOLOR_Zhang2017RealtimeUI, DISCOXia2022DisentangledIC}.
Vitoria~et~al.~\shortcite{INST_Color_Su2020InstanceAwareIC} use instance detection for individual color prediction in order to enhance object-level semantic representation, followed by a fusion module to predict the final colorization based on the object and image features. Vitoria~et~al.~\shortcite{CHROMAGAN_Vitoria2019ChromaGANAP} employ a dual-branch structure for joint image classification and color prediction to enhance global image level representation with adversarial training. Kumar~et~al.~\shortcite{COLTRAN_Kumar2021ColorizationT} use an encoder-decoder conditional transformer for low-resolution coarse color auto-regression, followed by additional spatial and color up-sampling networks to predict the final colorization. While utilizing adversarial training setups decreases the regression to the mean problem, it introduces other challenges, namely unstable training and mode collapse. Attempts to benefit from the advantages while mitigating the drawbacks include Deoldify \cite{DeOldify} which utilizes a scheduled adversarial loss, with careful tuning, termed NoGAN. In contrast, our method is simpler to train as it uses a regression loss which is relatively stable, together with an iterative diffusion process that is robust and expressive, while not suffering from mode collapse. We compare our method to the previously discussed approaches in \Cref{exp:automatic_colorization}.

\begin{figure}
\setlength{\tabcolsep}{2pt}
\centering
\begin{tabular}{ccc}
\includegraphics[ width=0.3\linewidth]{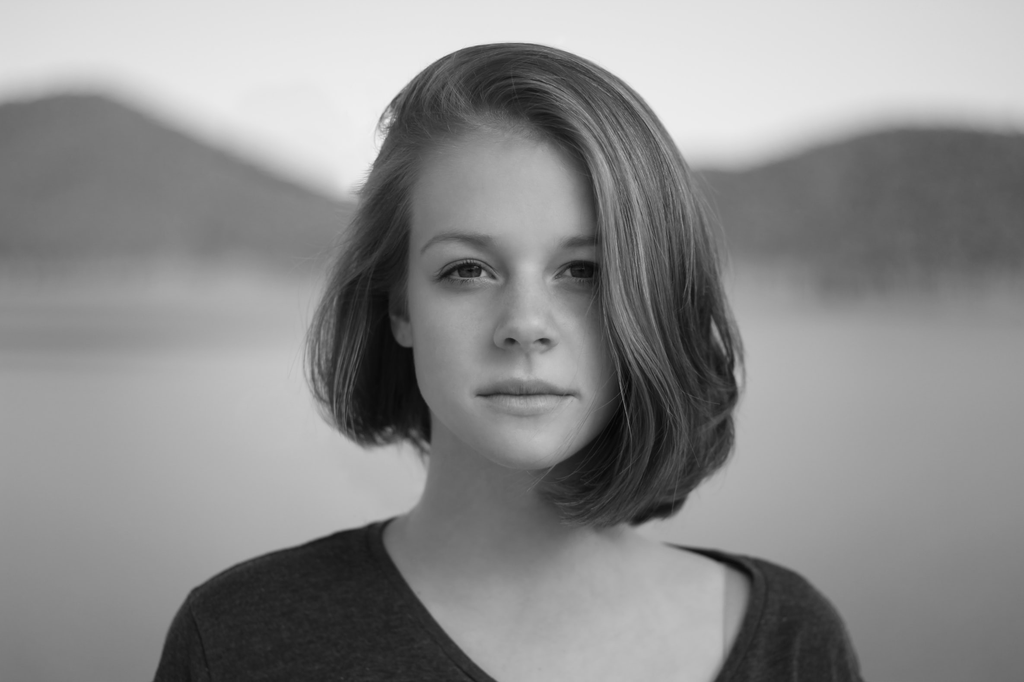}  &
\includegraphics[ width=0.3\linewidth]{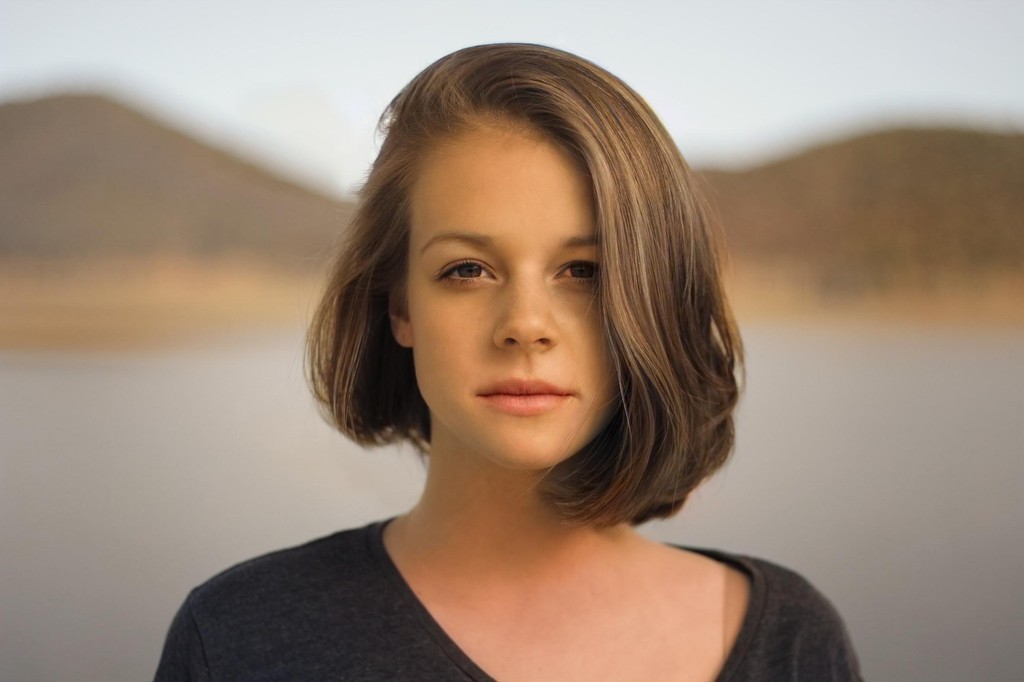} &
\includegraphics[ width=0.3\linewidth]{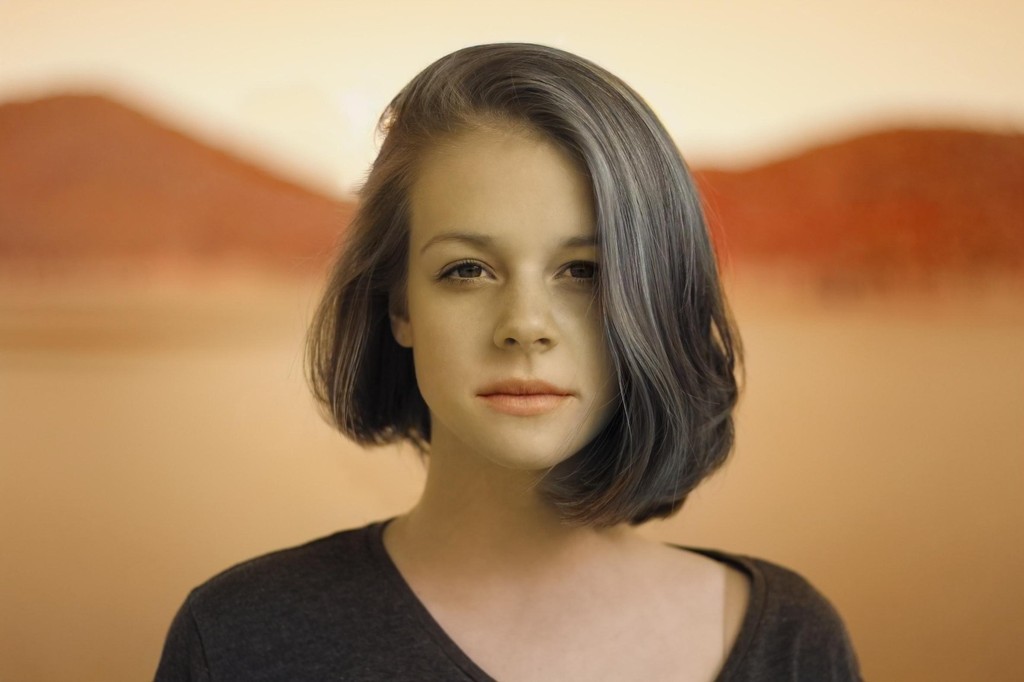}  \\
Input & ``Harvest time'' & ``Gothic Ambience'' \\
\includegraphics[ width=0.3\linewidth]{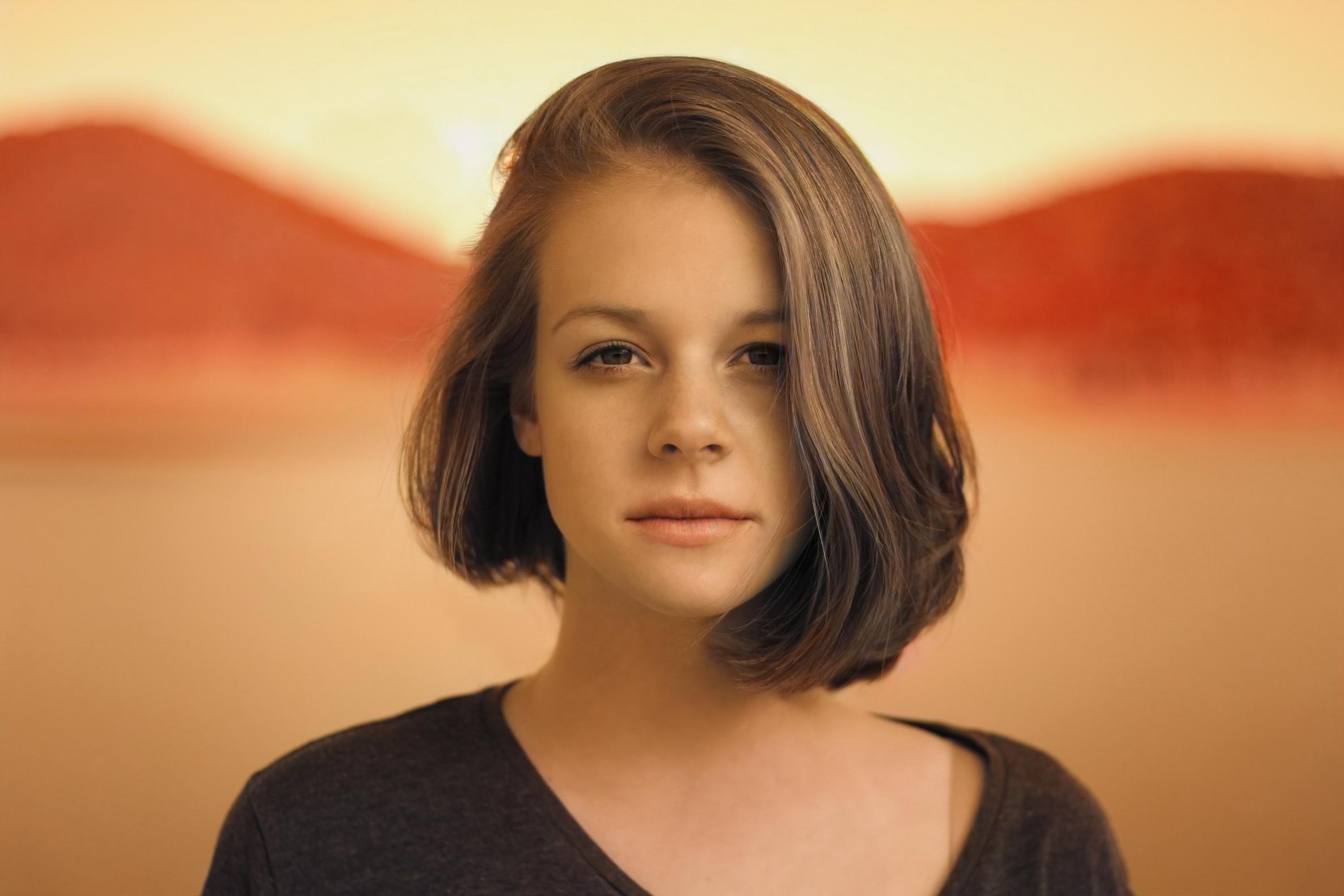} &
\includegraphics[ width=0.3\linewidth]{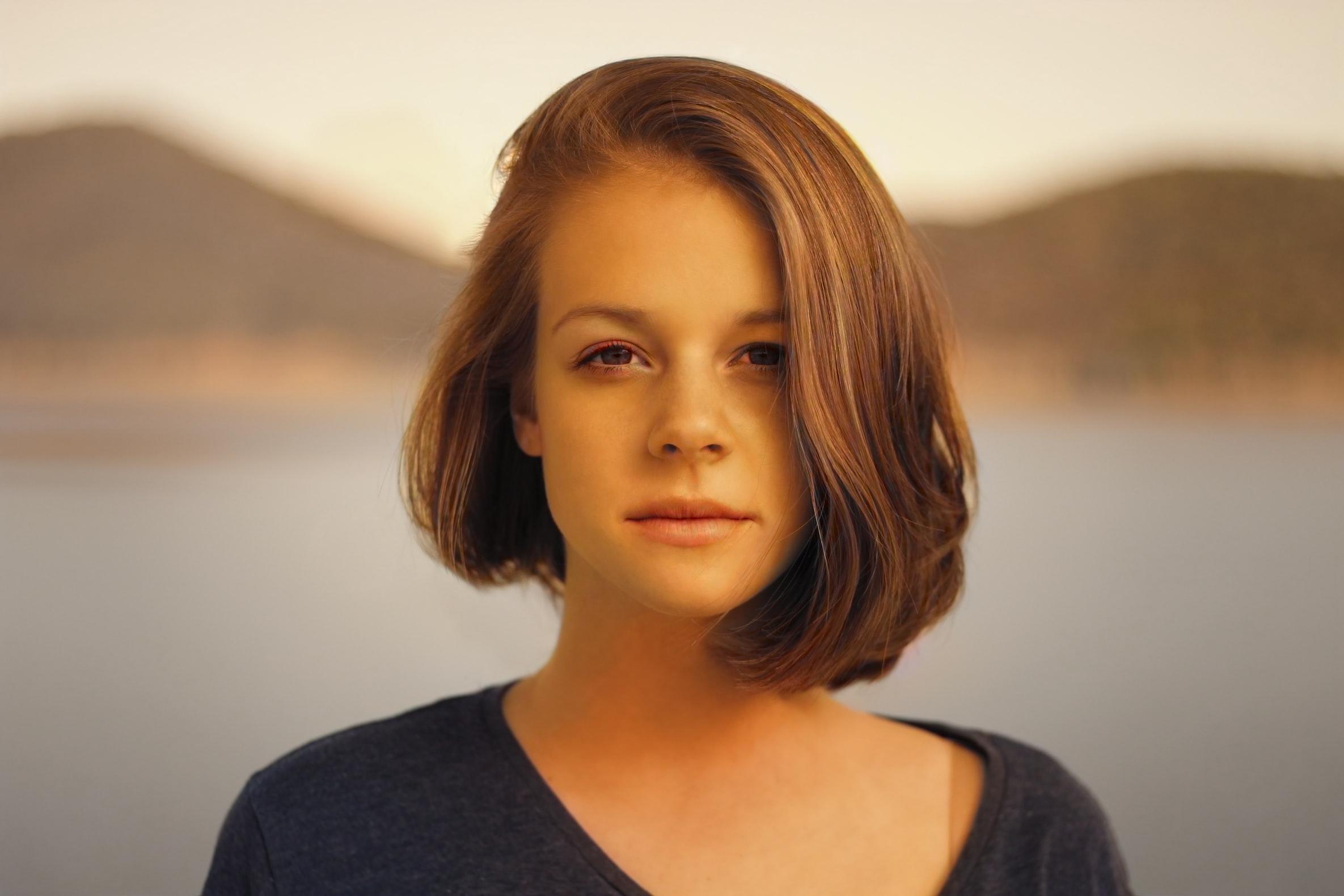}  &
\includegraphics[ width=0.3\linewidth]{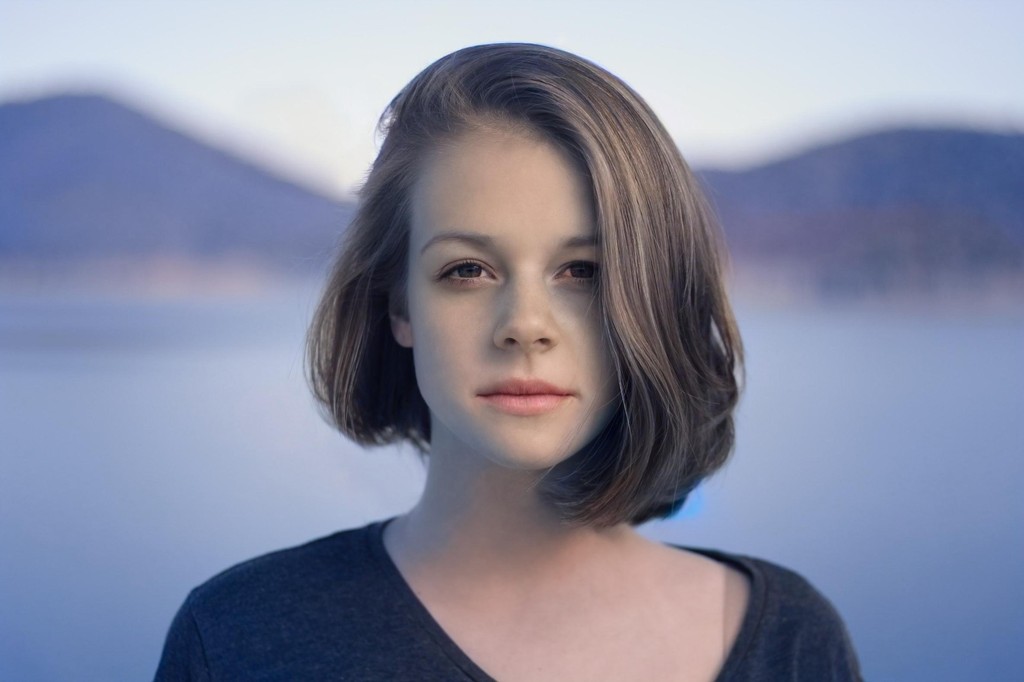} \\
``Romantic Candlelit'' & ``Golden Hour'' & ``Arctic Expedition'' \\
\end{tabular}

\caption{\textbf{Stylized Colorization Driven by Descriptive Text.} We visualize our model's capability to adapt colorization according to various thematic prompts, thereby exhibiting greater text-guided control that extends beyond basic "color-object" descriptions. Image credit: Unsplash ©Christopher Campbell.}

\label{fig:text_based_style_words}
\end{figure}
\subsection{Text-Guided Colorization}
Text prompts offer a straightforward way to guide the colorization process. It does not require interaction with the canvas as scribble based models require. Instead, it utilizes user input in the form of text descriptions, where the most na\"ive way is incorporating color words associated with specific objects such as ``blue jacket'' \cite{LCFL_Manjunatha2018LearningTC, Chang2022LCoDerLC, Weng2022LCoDeLC, huang2022unicolor} as shown in \Cref{fig:text_qualitative_comparison}. Direct color text instructions can mitigate the inherent color uncertainty of the task, providing the ability to render more accurate colorization results as per user guidance. While this may not always adhere to realism, it certainly offers users a greater degree of control over the process. The early work of \cite{LCFL_Manjunatha2018LearningTC} presented architectures that enabled basic textual guidance, by integrating textual guidance from LSTM and image features from FCNN layers. As the spatial control was limited, follow up work presented the L-CoDe(r) models \cite{Weng2022LCoDeLC, Chang2022LCoDerLC}, decoupled language conditions into a color space and an object space, thus solving the problem of color-object coupling as well as color-object mismatch. Although these methods offer different levels of control over how objects are colored, they all have limited expressiveness and focus on the simple case of color words (i.e., ``color-object'' coupling) using a small predefined vocabulary. As we fine-tune a powerful Diffusion Model with descriptive prompts, our model is free from this limitation. Instead, our approach allows for a broad range of more expressive global styles, including options such as ``film grain'', ``golden hour'', or ``in the winter'', demonstrated in \Cref{fig:text_based_style_words}. 

\subsection{Latent Diffusion Models} In recent years, Diffusion Models have been gaining prominence due to their impressive capabilities in a wide range of generative applications, including cutting-edge image \cite{Dalle_Ramesh2021ZeroShotTG, IMAGEN_Ho2021CascadedDM}, text \cite{GRAD_Popov2021GradTTSAD}, audio \cite{Kong2020DiffWaveAV}, and video \cite{IMAGEN_VIDEO_Ho2022ImagenVH,Video_Diffusion_Ho2022VideoDM} generation. These models have demonstrated their applicability in numerous use cases, including image restoration \cite{DDNM_Wang2022ZeroShotIR}, image editing \cite{P2P_Hertz2022PrompttoPromptIE,Kawar2022ImagicTR,Avrahami_2022_CVPR,avrahami2022blended_latent,Avrahami_2023_CVPR}, and unpaired image-to-image translation \cite{saharia2022palette}. The flexibility of diffusion models enables them to handle both unconditional and conditional tasks \cite{Controlnet_Zhang2023AddingCC}, thereby making it possible to apply a single model to a variety of requirements. Although traditional diffusion models tend to have high computational costs and slow inference speeds, recent advancements in Latent Diffusion Models (LDMs) \cite{StableDiffusionRombach2021HighResolutionIS} have addressed these limitations, leading to more efficient training and faster inference without substantially sacrificing synthesis quality. LDMs transition the diffusion process to operate in a lower-dimensional latent space. Unlike traditional diffusion models \cite{DIFF_BEAT_GAN_Dhariwal2021DiffusionMB,DalleRamesh2022HierarchicalTI,Imagen_Saharia2022PhotorealisticTD} that operate directly in pixel space, LDMs perform perceptual image compression in the first stage using an auto-encoder such as a VAE. In the second stage, a diffusion model is used to operate on the lower-dimensional latent space, which enables faster training and inference on limited computational resources. Rombach et al.~\shortcite{StableDiffusionRombach2021HighResolutionIS} demonstrated that LDMs can be conditioned on various modalities, creating realistic images. We take advantage of this latent space to apply our customized diffusion process for image colorization. Differently from the method of Rombach et al., which uses this latent space for image generation, we repurpose it to implement a customized diffusion process for image colorization. 

In a concurrent work, Liu at al. \shortcite{IMPDIFFPIGGY2023ImprovedDI} leveraged the color prior in LDMs using an LDM guider and a lightness-aware VQ-VAE model.

\section{Method}
\label{sec:method}
The objective of our approach is to generate a three channel colorized version $x \in \mathbb{R} ^ {H\times W\times 3}$ of a single channel gray-scale image $x' \in \mathbb{R} ^ {H\times W}$, potentially guiding this process with a text prompt. First, we explore the behaviour of the VAE of Latent Diffusion \cite{StableDiffusionRombach2021HighResolutionIS} when encoding grayscale and colorized images (\Cref{sec:color_analysis}). The findings motivated us to exploit the linear behaviour of the color in latent space for image colorization within this space. 
We trained a customized U-Net to predict residual latent vectors (\Cref{sec:latent_cold_diffusion}) representing the colors of images. This approach offers the advantages of accelerated training and inference run-times (\Cref{sec:color_sampling}). Moreover, 
motivated by the linear nature of color vectors in latent space, we present a CLIP-based ranker that can automatically rank these vectors and select the one with the most appropriate and visually appealing colorization (\Cref{sec:color_ranker}).
\subsection{Analysis of Color Properties in Latent-Space}
\label{sec:color_analysis}
Initiating our image colorization approach, we utilize a significantly expressive latent space. We conducted experiments using the VAE encoder $E(x)$ from Latent Diffusion \cite{StableDiffusionRombach2021HighResolutionIS}, a renowned generative model operating within a latent space that spatially compresses images by a factor of $8$. Our investigation focuses on the VAE's encoding of grayscale $z_x' = E(x')$ and colorized $z_x = E(x)$ images, as fortunately the VAE was trained on a vast dataset of images, including grayscale images, so $x'$ is within the span of this model.
We computed the residual $\Delta = z_x-z_x'$, referred to as the ``color-latent'', that represents the color aspect within the latent space. 
Our analysis reveals a linear correlation between the scaling of the color-latent vector and the resulting image's colorfulness and saturation. This relationship is demonstrated in \Cref{fig:latent_space_viz}, where we showcase it qualitatively and quantitatively. First, we demonstrate the similarity between scaling chromaticity channels in color images (left side of the image split) and scaling of the color vector $\Delta$ in latent space (right).
In addition, we quantitatively measure the similarity between scaled color images and images decoded after scaling in the latent space.
We use FID \cite{FID} and Absolute Delta Colorfulness \cite{colorfulness_metric}, common metrics which are typically used to assess the quality of image colorization.
Based on our observation, we apply our colorization method in the latent space, thus benefiting from its advantages of faster training and inference times.

\begin{figure}
\setlength{\tabcolsep}{2pt}
\centering
\includegraphics[width=\linewidth]{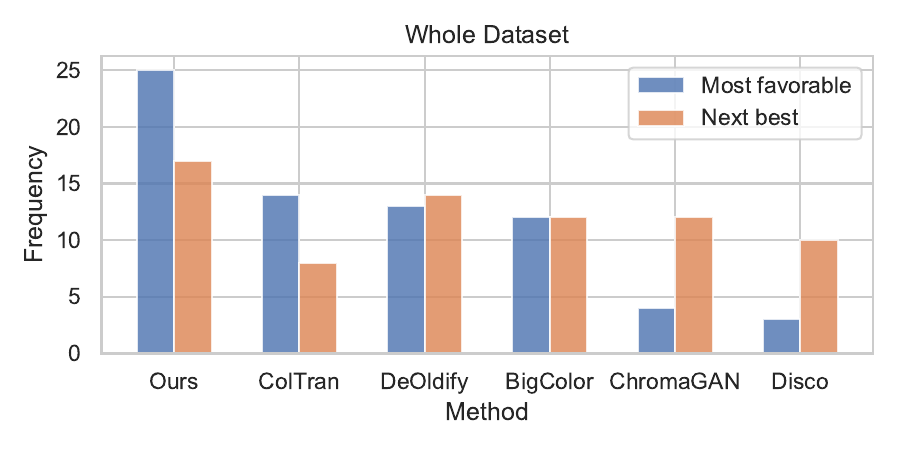}
\begin{tabular}{lcccccc}
\toprule
\textbf{Method} & Ours & BigColor & DeOldify & ColTran & ChromaGAN & Disco \\
\midrule
\textbf{\#Votes} & 676 & 596 & 585 & 573 & 457 & 410 \\
\textbf{Elo} & 86 & 65 & 61 & 59 & 18 & 0 \\
\bottomrule
\label{tab:user_study_total_picks}
\end{tabular}
\caption{\textbf{User Study Results.} A comparison of colorization methods based on the results of a user study. The chart displays the number of times each method was chosen as the most favorable and as the next-best choice by participants, over the entire curated dataset. Our method was selected as the most favorable choice by a large margin (top chart) and was chosen more frequently than the other methods (bottom table).}
\label{fig:user_study_main}
\end{figure}

\subsection{Latent Cold-Diffusion Model (LCDM)}
\label{sec:latent_cold_diffusion}
Cold-Diffusion \cite{ColdDiffusionBansal2022ColdDI} is a generative framework whose training consists of two basic operations: a fixed degradation operator $D$, and a learned restoration operator $R$. For completeness, we will briefly present the essence of Cold Diffusion. During training, $D$ is used to create a degraded image $x_t$ for a random timestep $t\in \left[ 0,...,T \right]$ by applying a pre-defined degradation level to a clean training image $x_0$, i.e., $D(x_0, t)=x_t$. 
For instance, $D$ can be a blurring operator. The main objective of the restoration operator $R$, which is implemented as a neural network, is to reverse the effects of $D$ and approximate the original clean image $x_0$, from a given degraded image $x_t$, namely $R(x_t, t)=\hat{x_0}$.

\begin{figure*}

\setlength{\tabcolsep}{2pt}

\begin{tabular}{c@{\hspace{1pt}}c@{\hspace{1pt}}c@{\hspace{1pt}}c@{\hspace{1pt}}c@{\hspace{1pt}}c@{\hspace{1pt}}c}
  \centering

\includegraphics[width=0.14\linewidth]{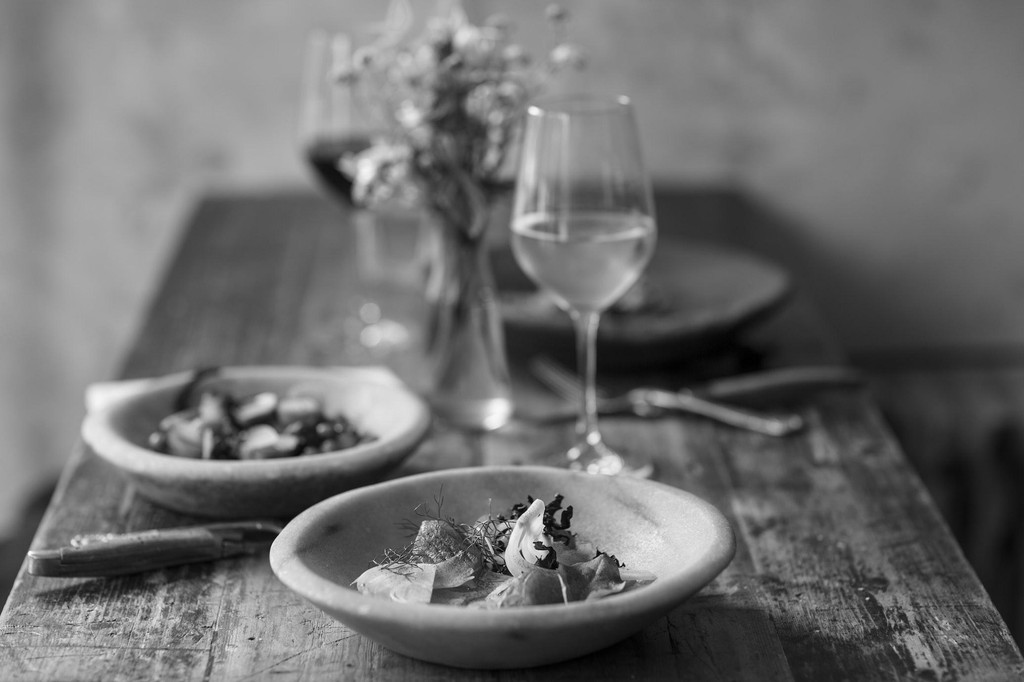} &
\includegraphics[width=0.14\linewidth]{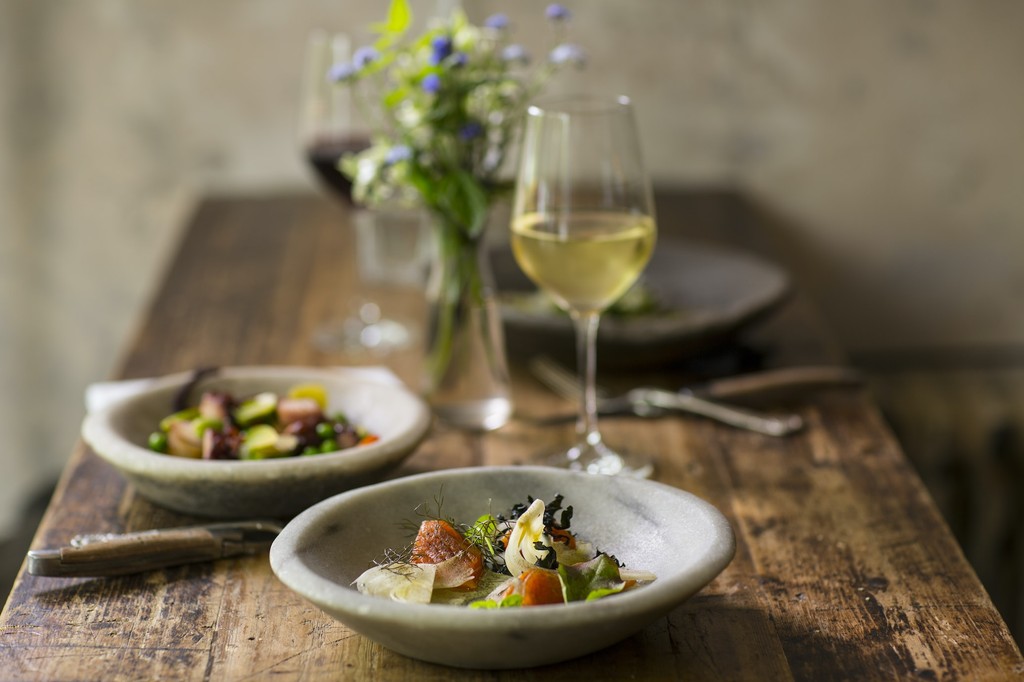} &
\includegraphics[width=0.14\linewidth]{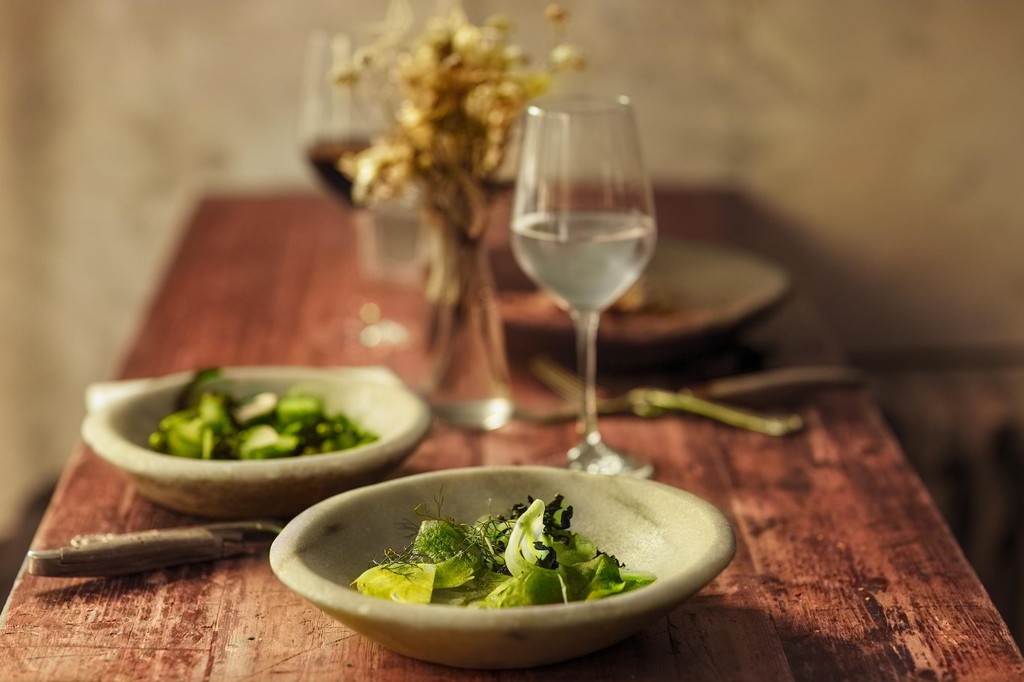} &
\includegraphics[width=0.14\linewidth]{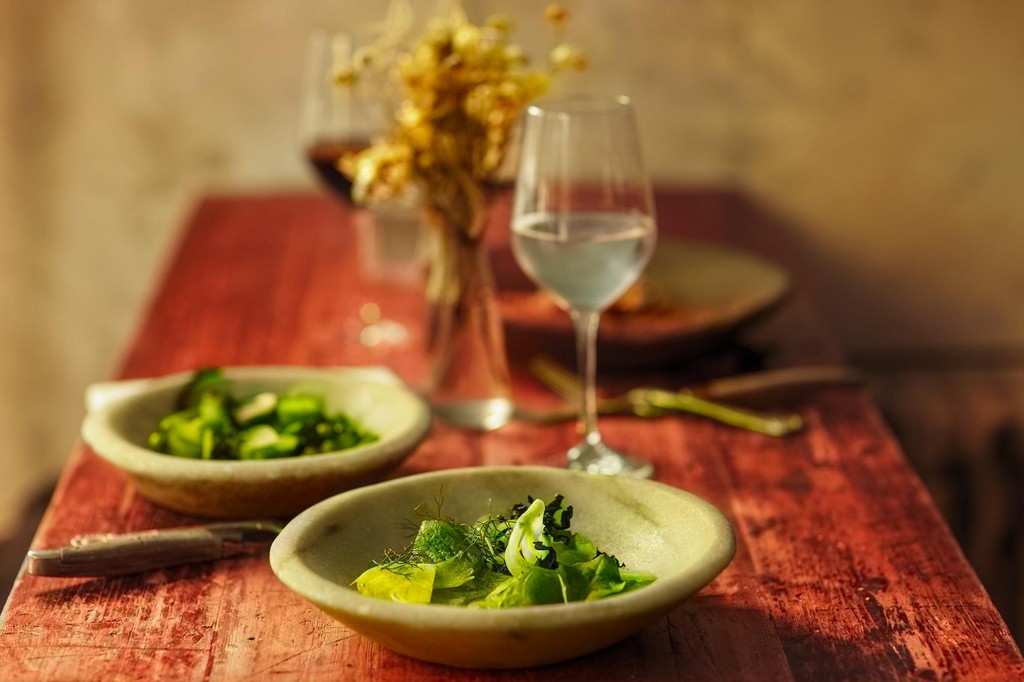} &
\includegraphics[width=0.14\linewidth]{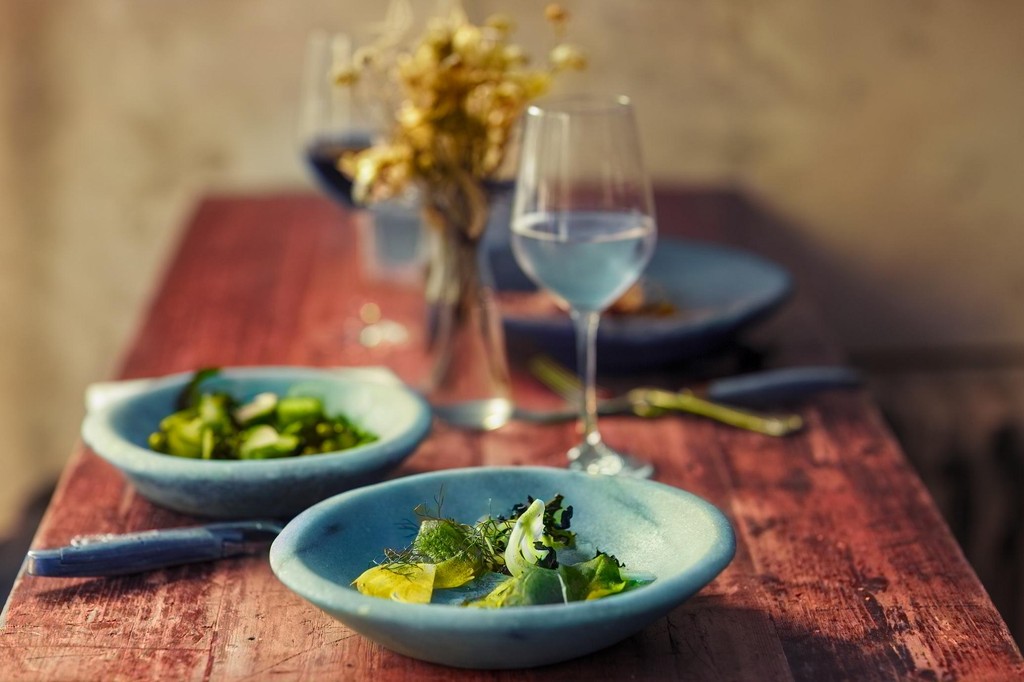} &
\includegraphics[width=0.14\linewidth]{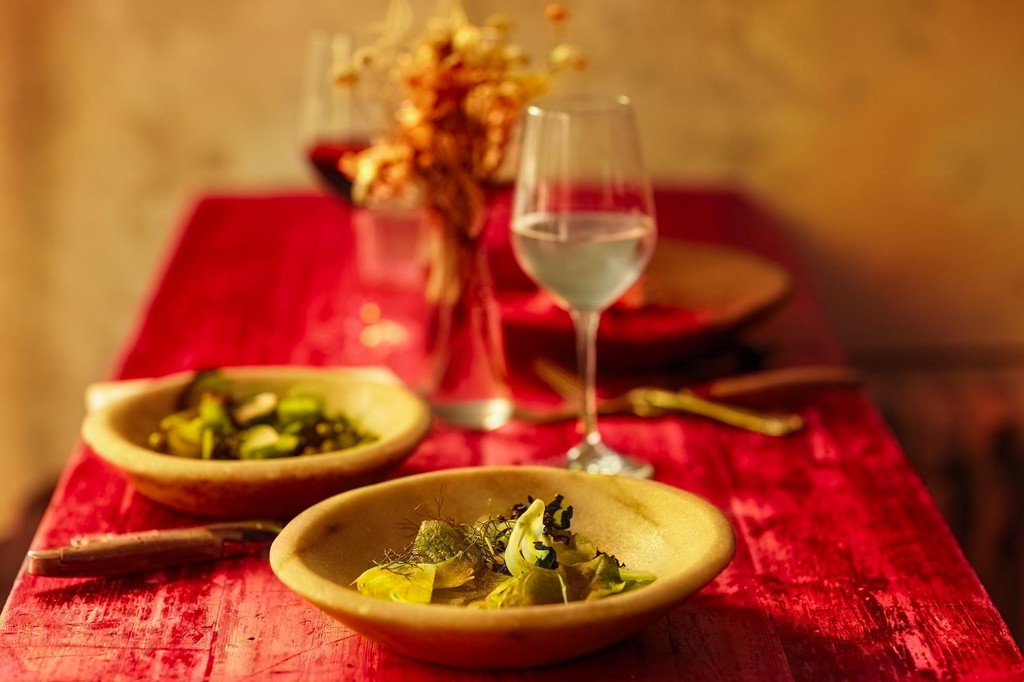} &
\includegraphics[width=0.14\linewidth]{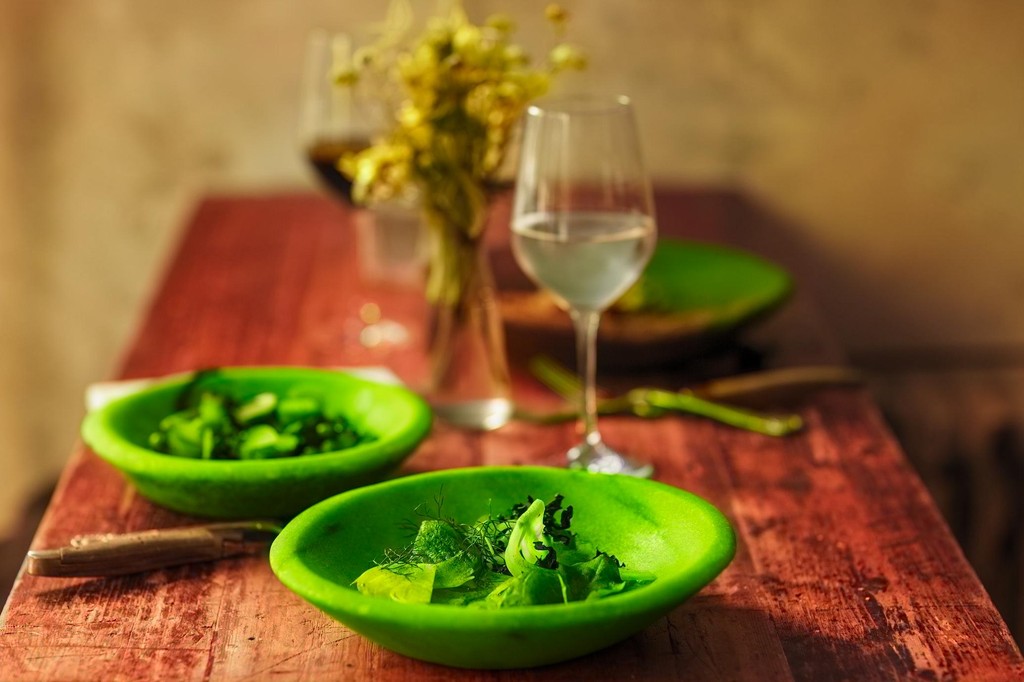} \\

\includegraphics[width=0.14\linewidth]{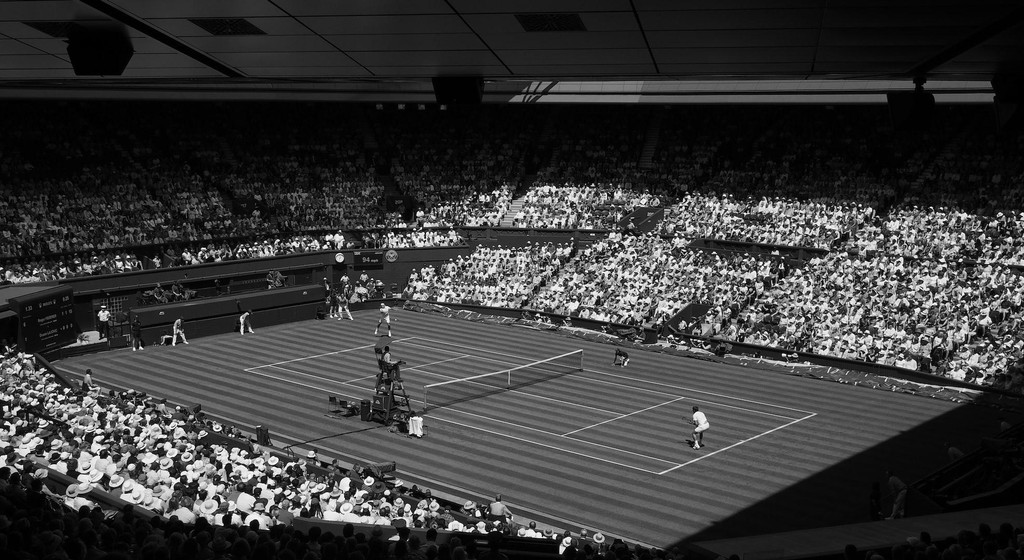} &
\includegraphics[width=0.14\linewidth]{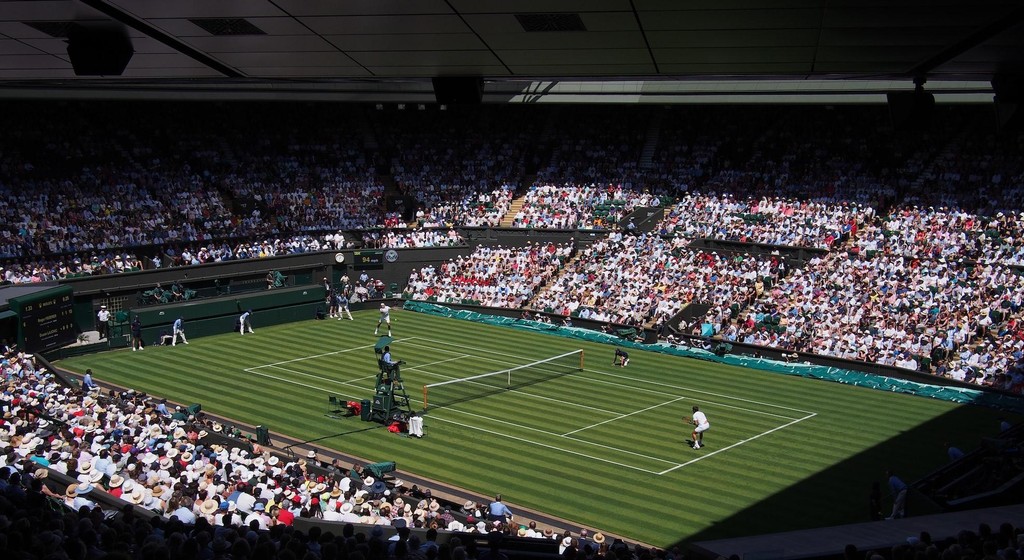} &
\includegraphics[width=0.14\linewidth]{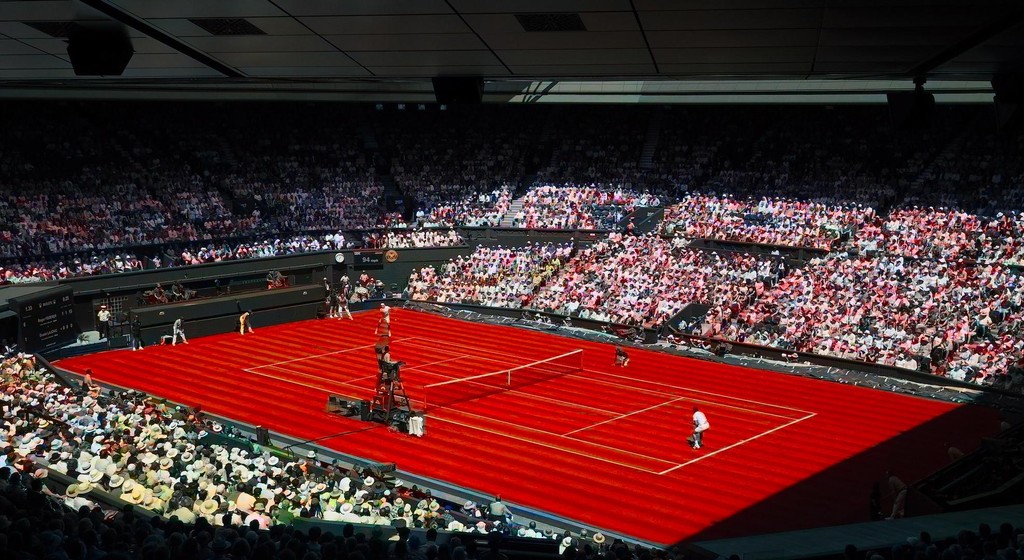} &
\includegraphics[width=0.14\linewidth]{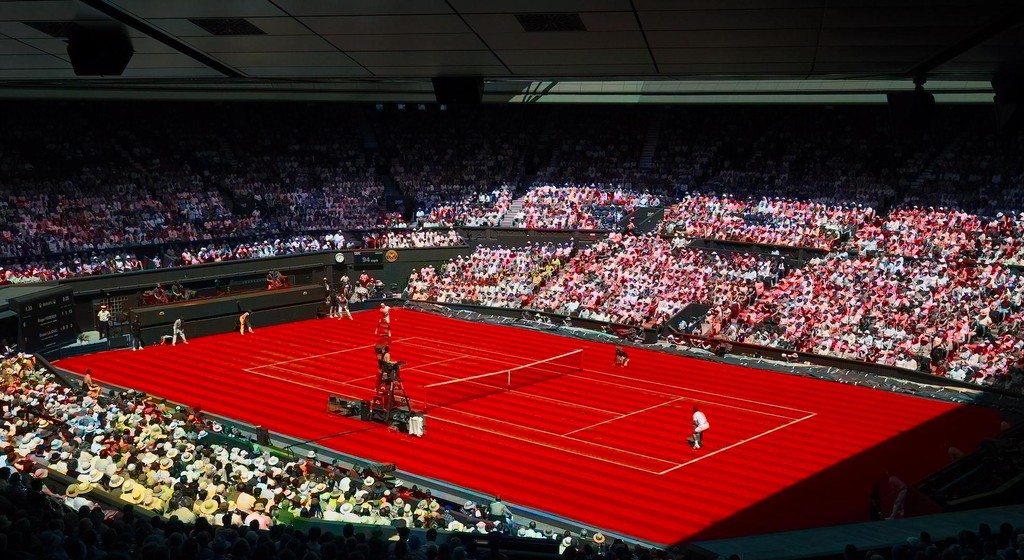} &
\includegraphics[width=0.14\linewidth]{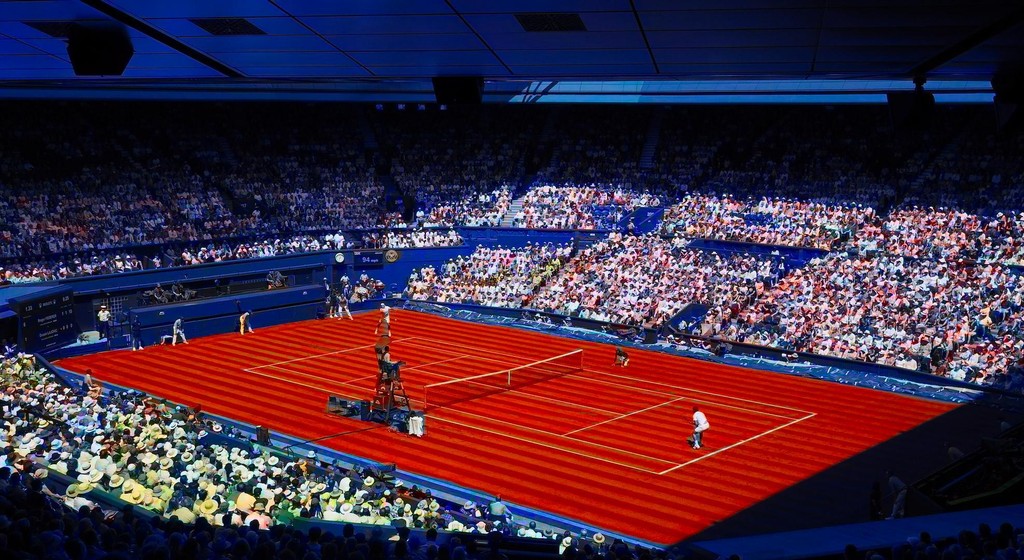} &
\includegraphics[width=0.14\linewidth]{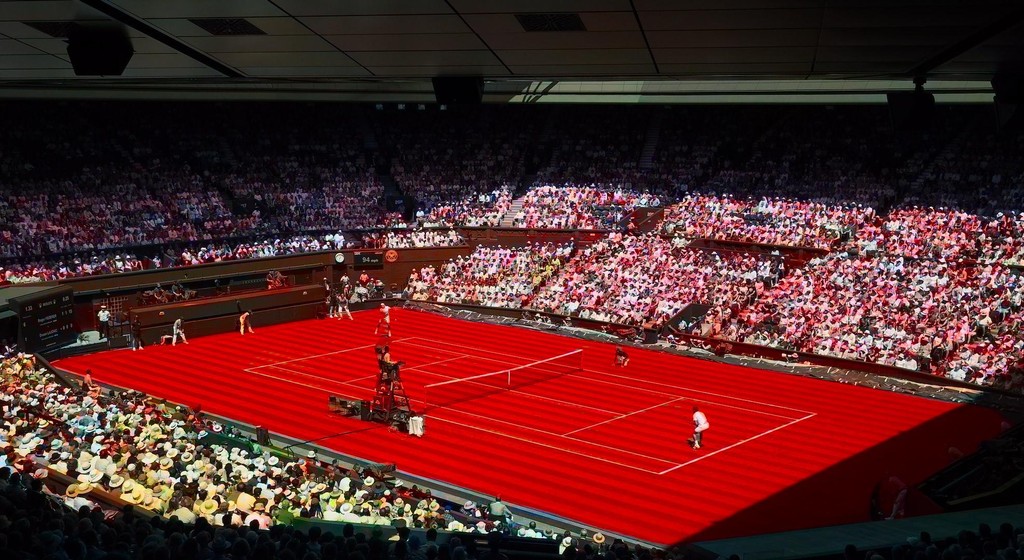} &
\includegraphics[width=0.14\linewidth]{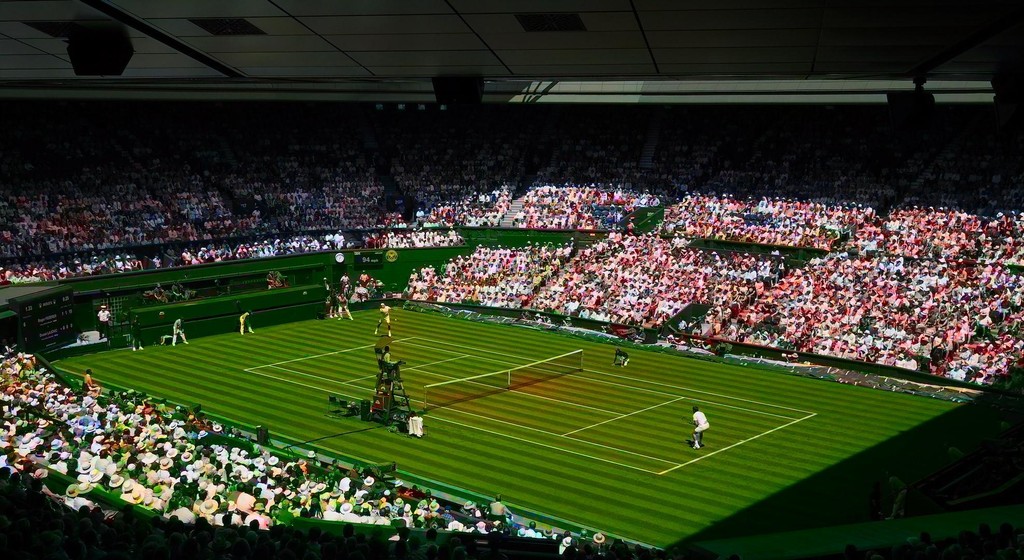} \\

\includegraphics[width=0.14\linewidth, height=2.6cm]{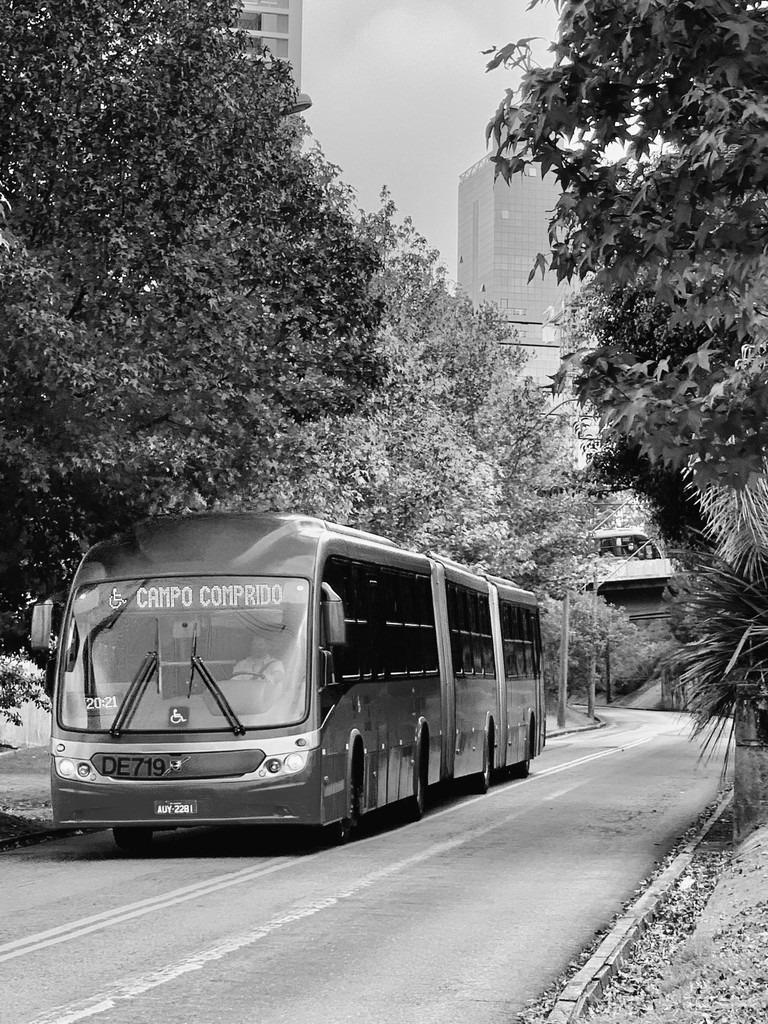} &
\includegraphics[width=0.14\linewidth, height=2.6cm]{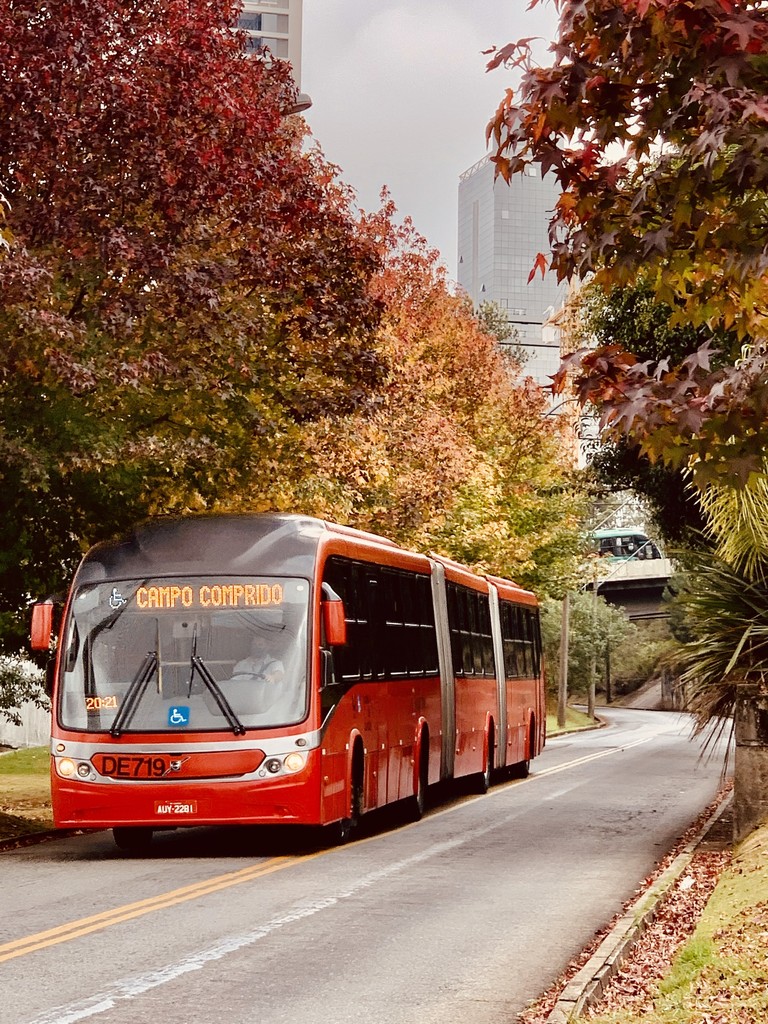} &
\includegraphics[width=0.14\linewidth, height=2.6cm]{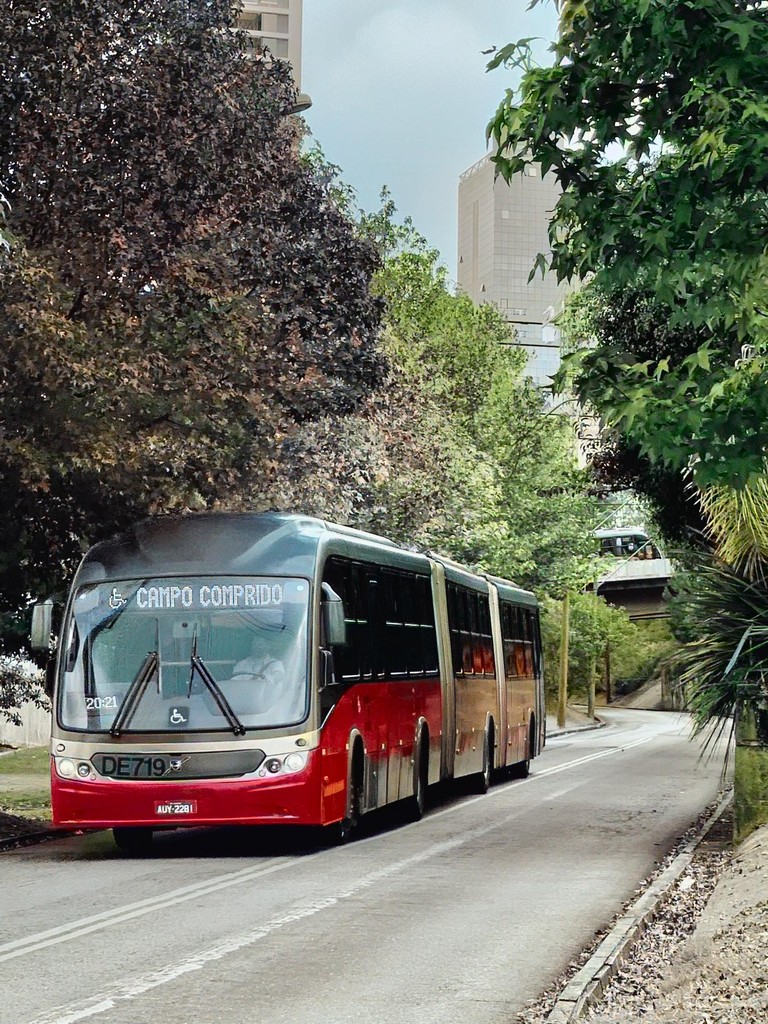} &
\includegraphics[width=0.14\linewidth, height=2.6cm]{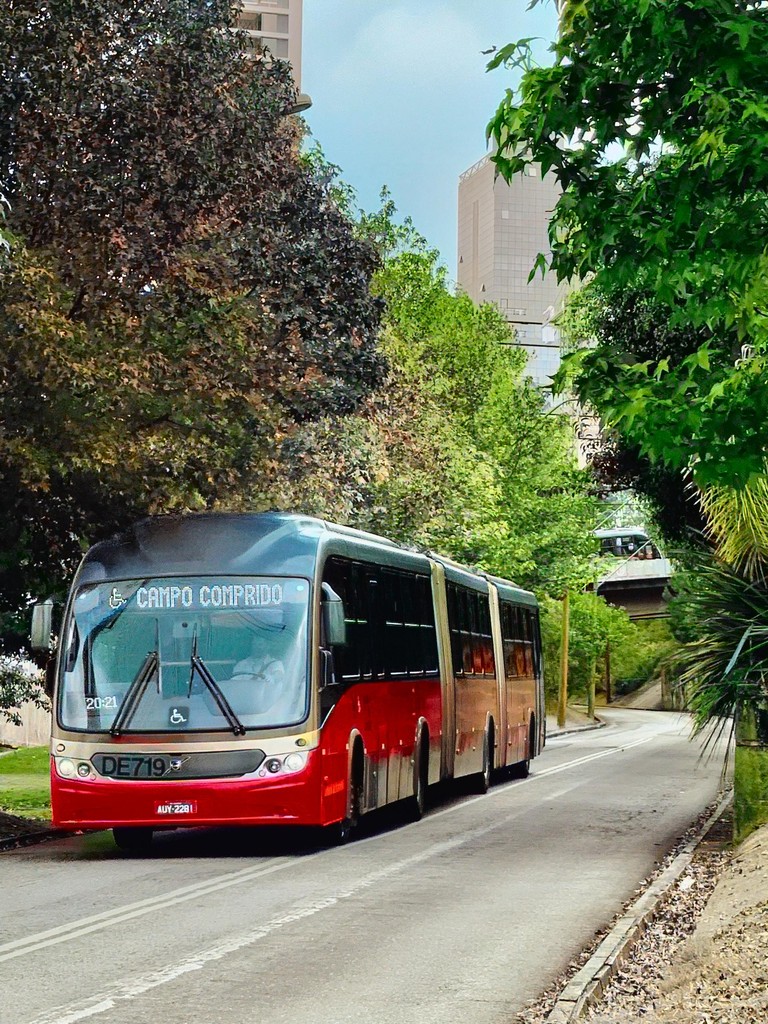} &
\includegraphics[width=0.14\linewidth, height=2.6cm]{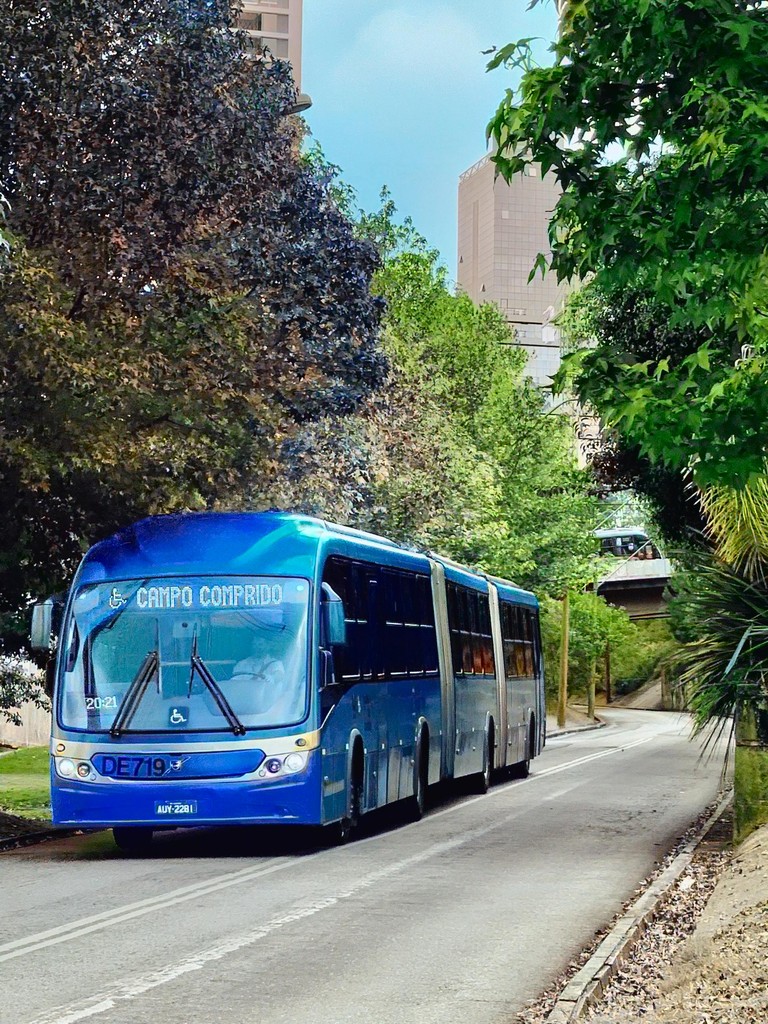} &
\includegraphics[width=0.14\linewidth, height=2.6cm]{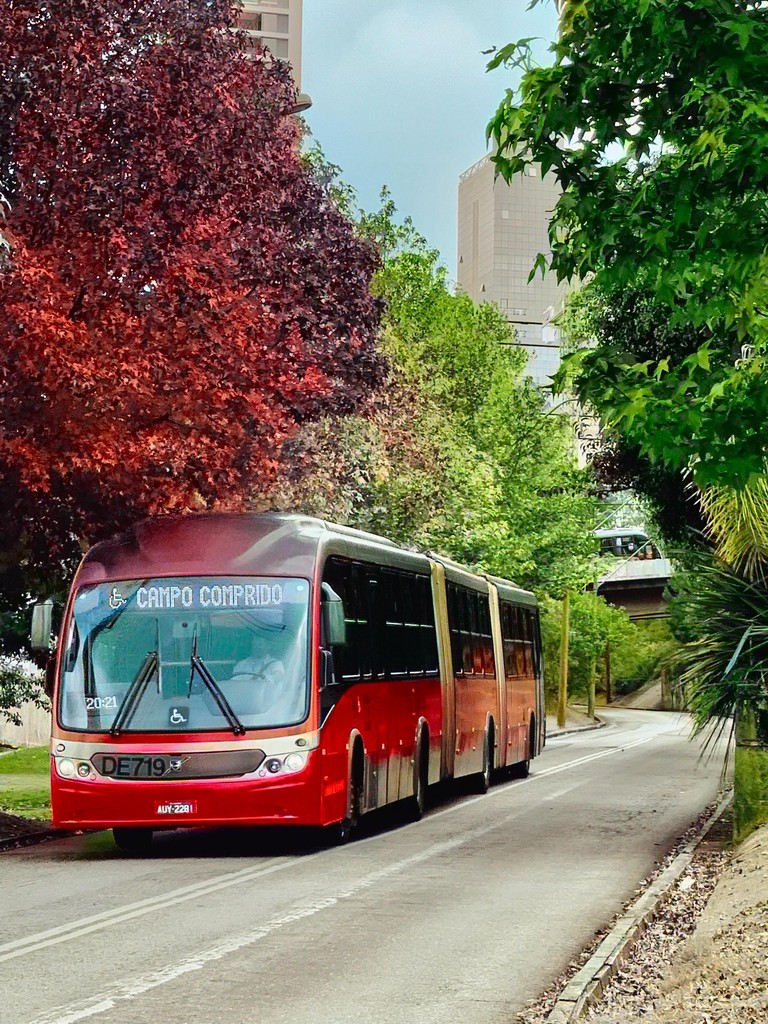} &
\includegraphics[width=0.14\linewidth, height=2.6cm]{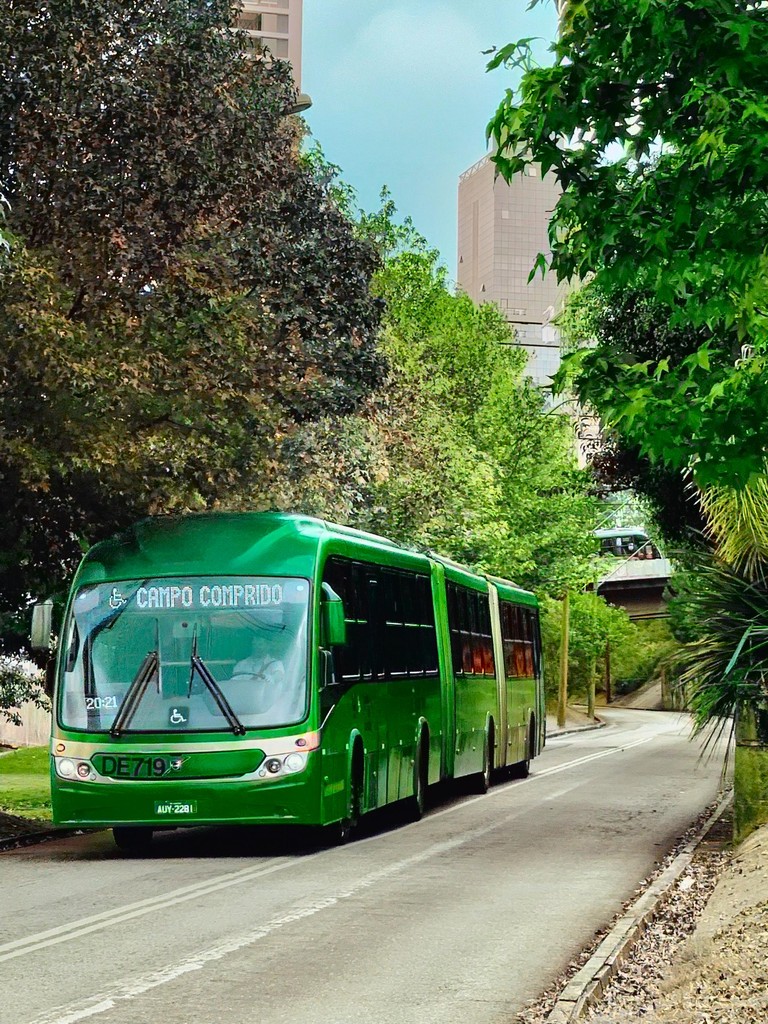} \\

\includegraphics[width=0.14\linewidth]{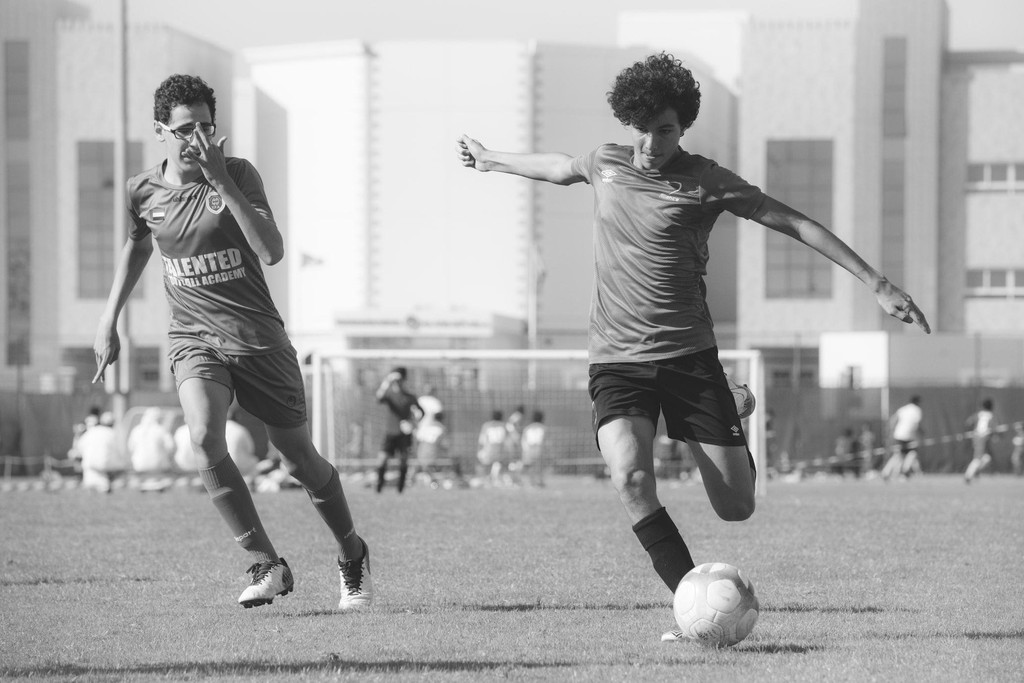} &
\includegraphics[width=0.14\linewidth]{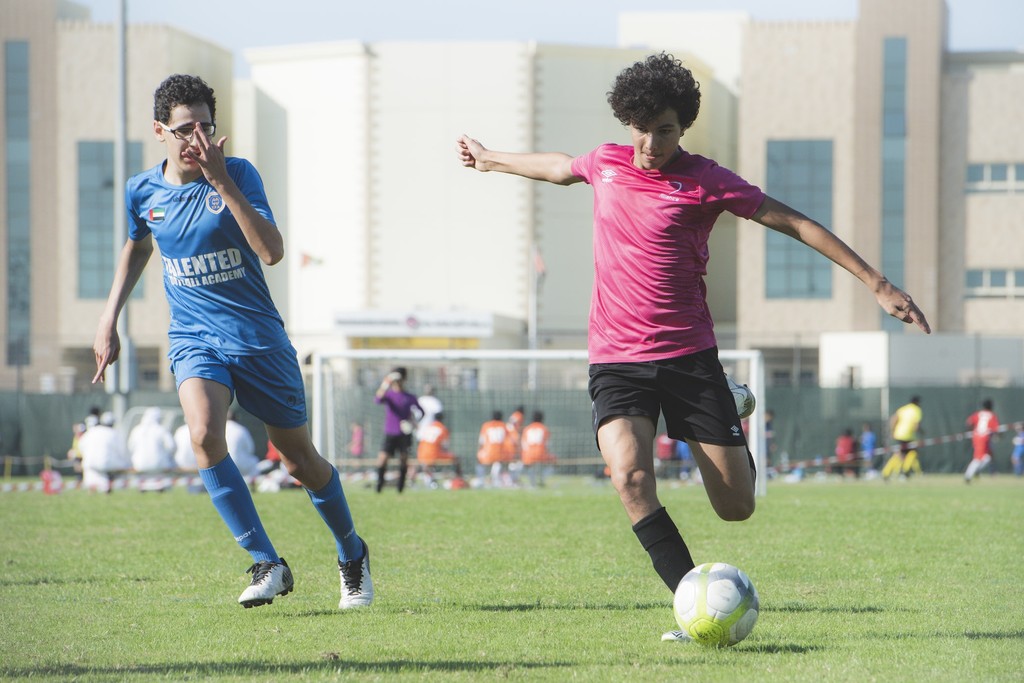} &
\includegraphics[width=0.14\linewidth]{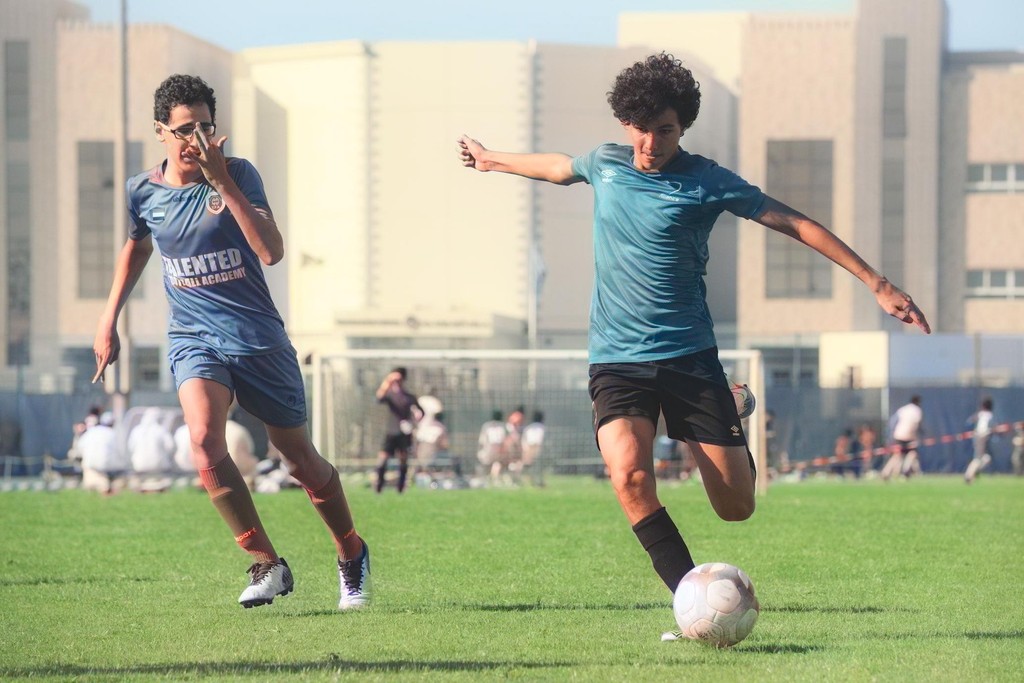} &
\includegraphics[width=0.14\linewidth]{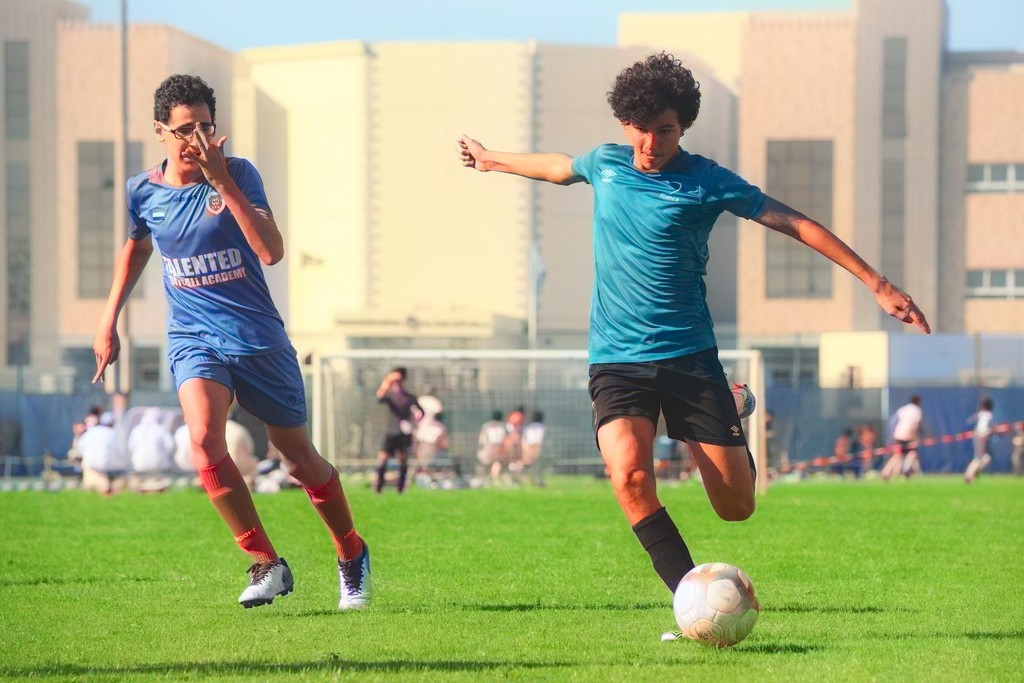} &
\includegraphics[width=0.14\linewidth]{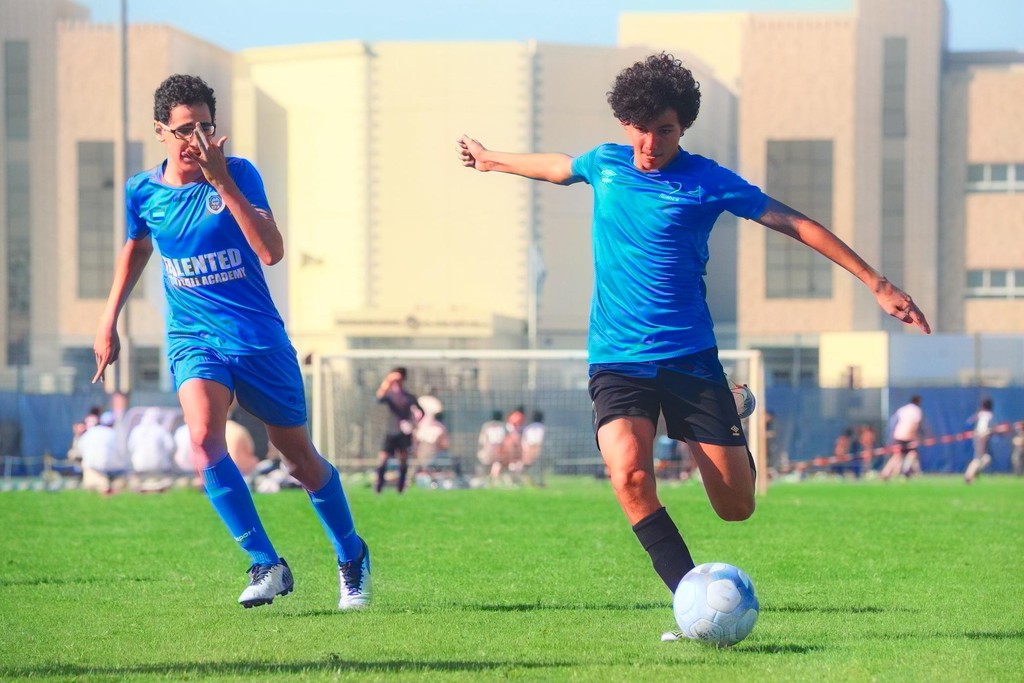} &
\includegraphics[width=0.14\linewidth]{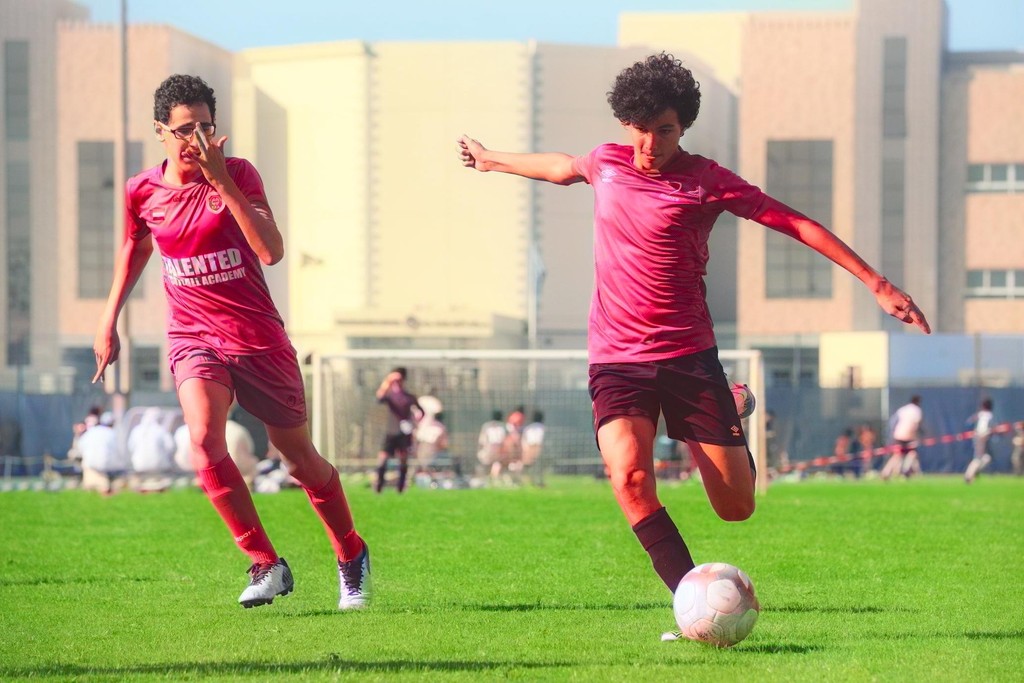} &
\includegraphics[width=0.14\linewidth]{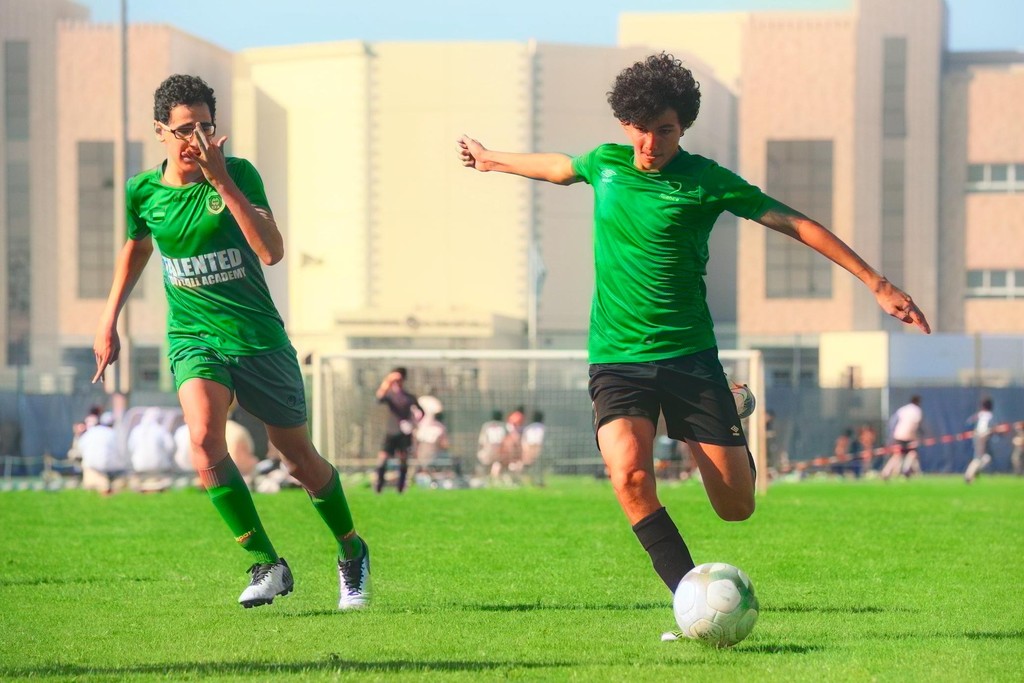} \\

Input & GT & Null Prompt & +Negative Prompt & +\textcolor{blue}{"Blue Tones"} & +\textcolor{red}{"Red Tones"} & +\textcolor{green}{"Green Tones"} \\ 

\end{tabular}

\caption{
\textbf{Prompt Effect on Image Colorization.} A qualitative visualization of image colorization is showcased through the use of diverse prompts. The input grayscale image and its corresponding ground truth colorized image are shown in the first and second columns.
In the third column, we visualize the colorization with the null prompt. Fourth column is generated using negative prompts of ``grainy black-and-white photo, photo taken on an old box camera, grayscale photography''. Finally, positive prompts are used to direct the colorization model to produce specific color tones in the remaining columns. Image credits (top to bottom): Unsplash ©Stefan Johnson, Unsplash ©Shep McAllister, Unsplash ©3Três Consultoria e Criação, Unsplash ©Alliance Football Club (unsplash). 
}
\label{fig:text_effect}

\end{figure*}

In our case, the degradation operator $D$ removes the chroma information from an RGB image, and the recovery model $R$ introduces new chroma information. Unlike Cold Diffusion, we operate in the latent space of the LDM instead of the pixel space. As evident from \Cref{sec:color_analysis}, affine combinations of the image pair (RGB, grayscale) in this latent space are equivalent to varying the colorfulness and saturation level in the corresponding pixel space. Therefore, any convex combination $\alpha D(z_x) + (1-\alpha) z_x, \alpha \in [0,1]$ in latent space, corresponds to a natural image within the span of $E$, and linear degradation in latent space will generate realistic samples for training our model in different timesteps.

The Latent Diffusion architecture gets an additional textual input, encoded using CLIP \cite{CLIP_radford2021learning}; for simplicity, we will refer to both the text and its embedding as $c$, where the meaning should be clear from the context. 
We define the color residual at timestep $t$ to be $\Delta_{t}=t \cdot \Delta$,
and train a time-conditional UNet by using the encoded pair of image and caption, to predict $\hat{\Delta_{t}}=\text{UNet}(z_{x}^{t}, t, c)$. Specifically, our objective is:

\begin{equation}
\begin{aligned}
&  \underset{\substack{(x, c) \sim p_{data} \\ t \sim \mathbb{U}[0,1]}}{\min} \mathbb{E} \Big[ ||z_{x} - (z_{x}^{t} + \hat{\Delta_{t}}) || \Big]
\end{aligned}
\end{equation}

During training, we sample an (image, caption) pair, and a random timestep (i.e. degradation level), and train the UNet to predict the Color Residual at that specific timestep (see \Cref{fig:system_overview}). We keep the VAE frozen throughout training. 
We emphasize that we trained the UNet to predict the color-vector $\Delta$ given the degraded latent, rather than predicting the latent itself (i.e., $z_{x}\overset{!}{=}\text{UNet}\left(z_{x}^{t}, t, {z_c}\right)$), which showed early in our experimentation to produce inferior results. This is  known as $\epsilon$ prediction in Diffusion Model literature \cite{DDPM_Ho2020DenoisingDP}.

\subsection{Color Sampling}
\label{sec:color_sampling}
Our iterative color sampling procedure draws inspiration from Cold Diffusion's inference algorithm, in that we restore a given degraded image $x'=x_T$ in multiple steps. The resulting image from the initial restoration $\hat{x} = R(x_T, T)$ undergoes further degradation (i.e. ``re-degraded'') with decreasing severity levels given by $x_t = D(\hat{x}, t), t=T-1,...,1$. This process is 

repeated to progressively reconstruct the color image.
We first encode the grayscale image $z_{x'} = E(x')$, and the UNet predicts the initial color residual %
$\hat{\Delta_{t}} = \textit{UNet}(z_x', T, c)$

\begin{figure}

\begin{tabular}{lcccc}
\centering
\rotatebox{90}{\shortstack{\textcolor{orange}{Orange} Dog \\ \textcolor{blue}{Blue} Couch }} &
\includegraphics[ width=0.22\linewidth, height=2cm]{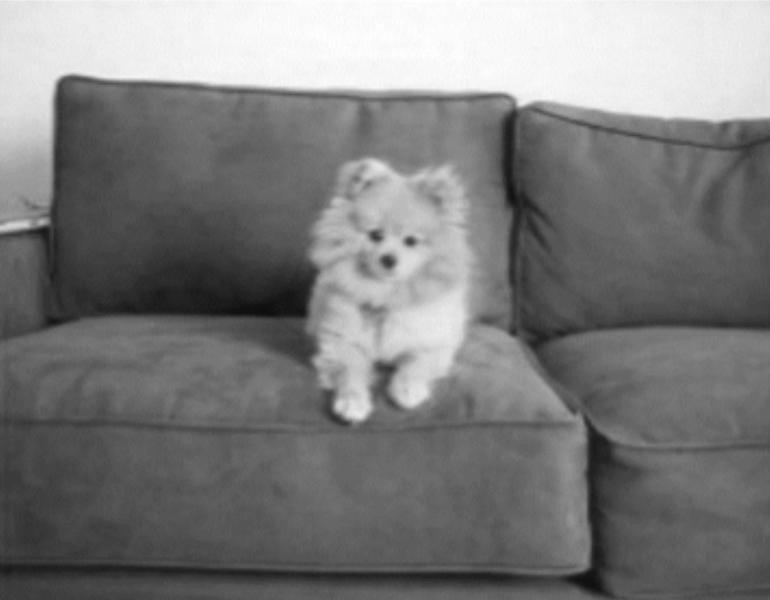} &
\includegraphics[width=0.22\linewidth, height=2cm]{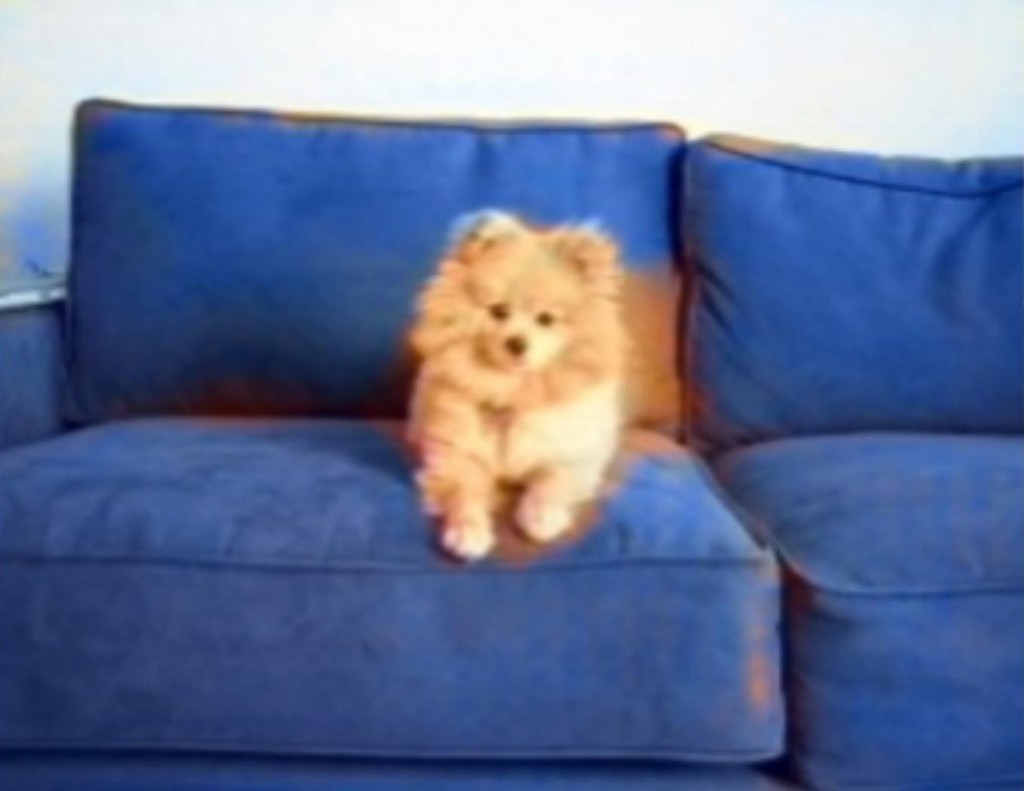} & 
\includegraphics[width=0.22\linewidth, height=2cm]{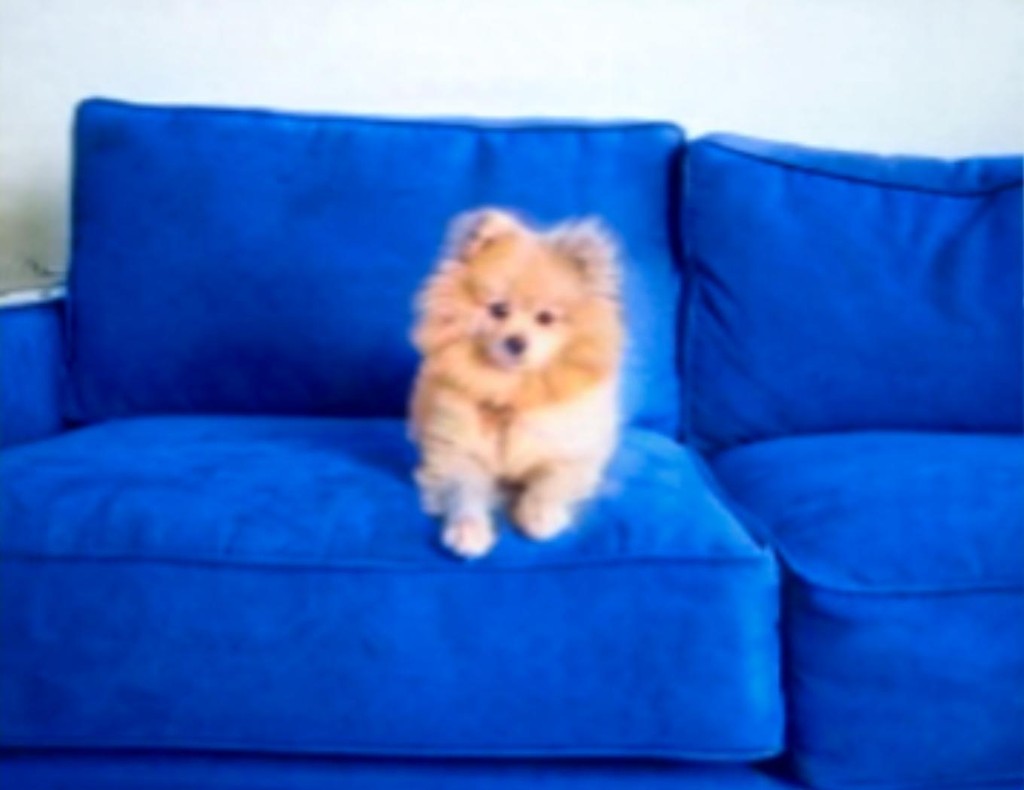} &
\includegraphics[width=0.22\linewidth, height=2cm]{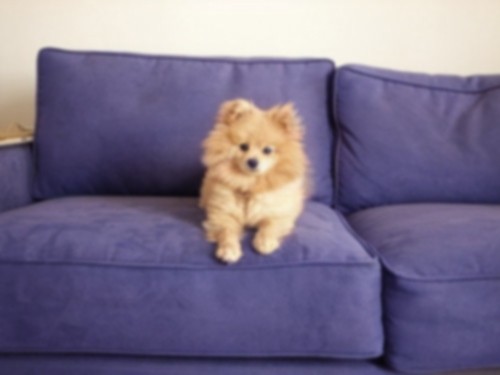} \\

\rotatebox{90}{\textcolor{gray}{Gray} Horse} &
\includegraphics[ width=0.22\linewidth, height=2cm]{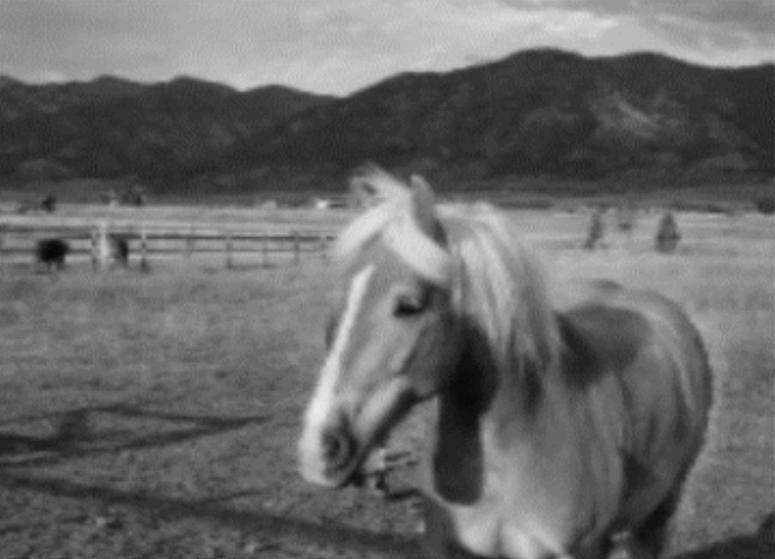} &
\includegraphics[width=0.22\linewidth, height=2cm]{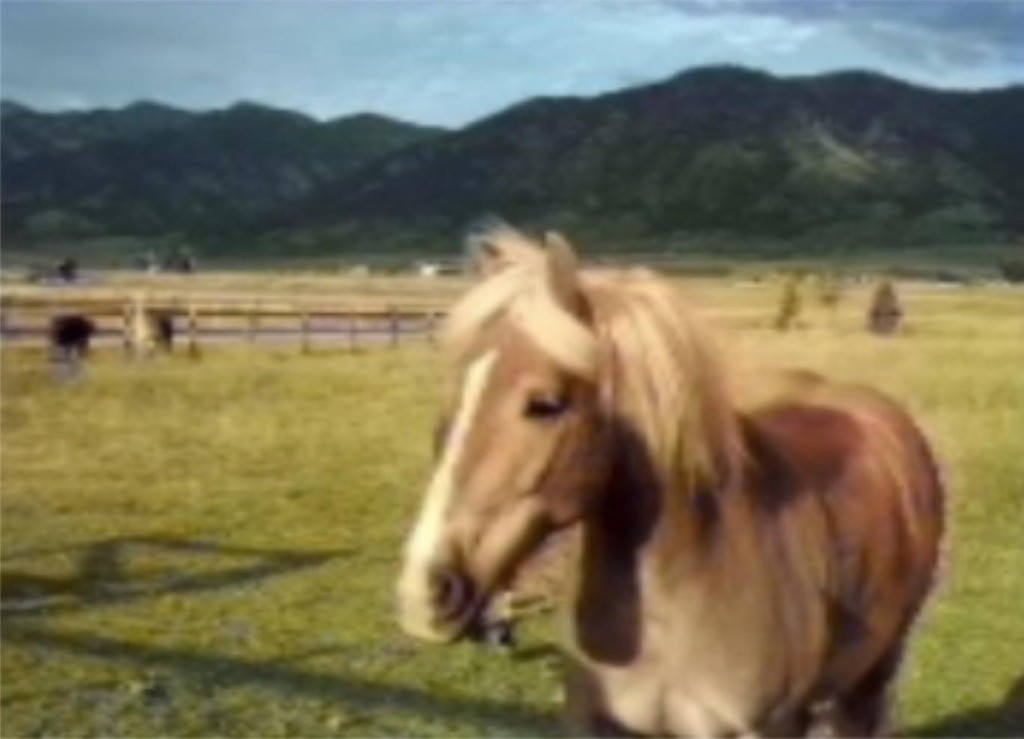} & 
\includegraphics[width=0.22\linewidth, height=2cm]{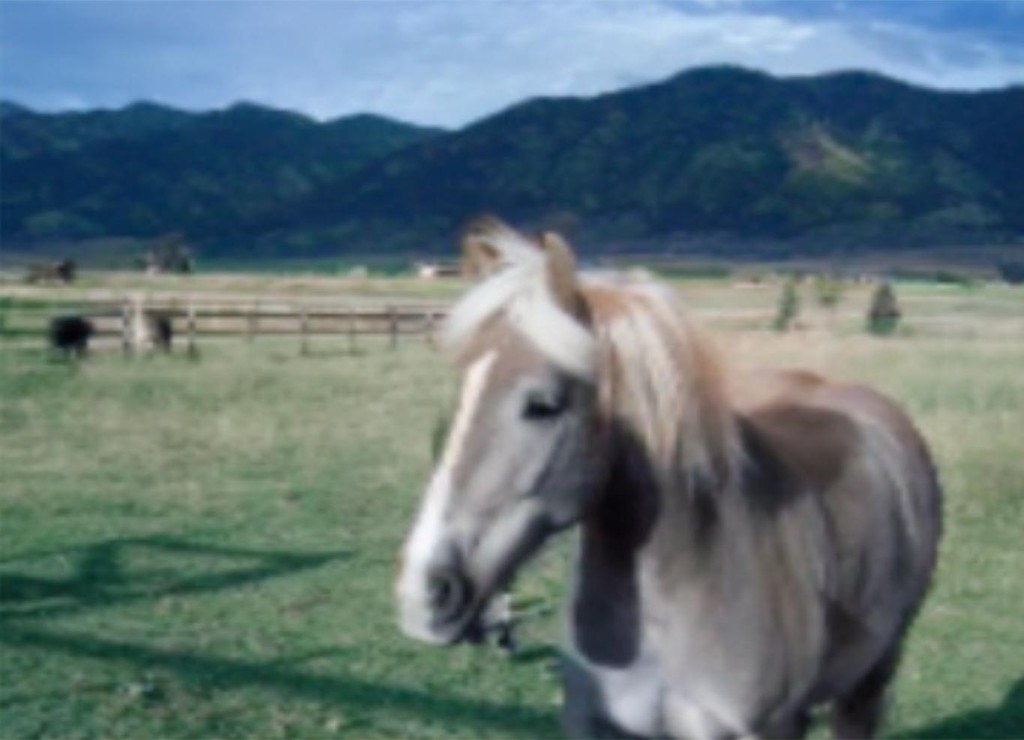} &
\includegraphics[width=0.22\linewidth, height=2cm]{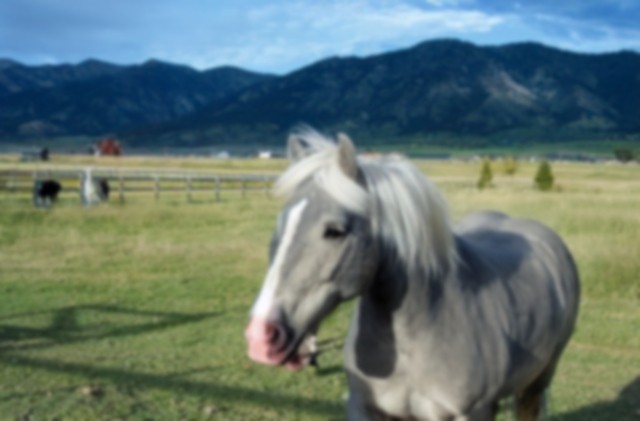} \\
& Input & L-ColorLang & UniColor & Ours
\end{tabular}
\caption{\textbf{Qualitative comparison with other text-based colorization methods}. We illustrate a side-by-side comparison with "L-ColorLang" \cite{LCFL_Manjunatha2018LearningTC} and UniColor \cite{huang2022unicolor}. Images were taken from UniColor \cite{huang2022unicolor}. }
\label{fig:text_qualitative_comparison}
\end{figure}
    
Then, we repeatedly degrade the results back to an earlier timestep $t-1$, and the UNet predicts $\hat{\Delta_{t-1}} = \textit{UNet}(z_x^t, t-1, c)$ for a predefined schedule of time-steps. We show that concepts such as negative-prompt and classifier-free-guidance \cite{Ho2022ClassifierFreeDG} can be used also in the context of Cold Diffusion, as visualized in \Cref{fig:text_effect}. When the number of diffusion steps increases, we generally get better results quantitatively, as showed in the supplementary material (SM), and we demonstrate qualitatively in \Cref{fig:stride_effect_fig} the increased color vividness. Similarly to other colorization works we replace the luma channel of the output with that of the input in order to retain the details of the original image and only manipulate color information.

\subsection{Ranking of Image Colorfulness} 
\label{sec:color_ranker}
Motivated by our findings about the color separation of the latent space, we exploited it in another way. We scale the final predicted residual $\hat{\Delta_{0}}$ to tune the saturation up or down. Clipping artifacts due to over-saturated images is not a primary concern because the VAE decodes natural images even from slightly distorted latents. Thus, we experimented with dynamically scaling the color residual at inference time, namely $\hat{z_{x}}=z_{x'}+s \cdot \Delta_{0}$ with a free parameter $s$,
where $s=1$ corresponds to the default setup. In the interactive setup, the user can tune the color intensities to their liking using a single slider, without a large computational overhead, as the color residual is calculated only once and scaled according to the user preference.
To support the fully automatic setup, we trained a ranker to predict the best scale for each image. Given a pair of possible outputs generated with different scales, we optimized a linear regressor over the CLIP \cite{CLIP_radford2021learning} representations of the images, to give a higher score to the image with the preferable colorization. To this end, we manually labeled $500$ pairs of training images and $100$ pairs of test images, with a binary label indicating the image with better colors in each pair. The regressor is trained to maximize the difference between the score given to the  preferred image and the score of the other. We opted for this loss over a classification loss to allow inference per scale rather than per pair of scales. We found that the ranker agrees with our preferences in $92$ out of the $100$ test image pairs. At inference, we utilize the VAE to convert the scaled latents into colorized images in pixel space. The color-ranker assigns a score to each colorized variant, and the rendition with the highest score is chosen as the final output. Visual examples of this process are shown in the supplementary.

\section{Experiments}
\subsection{Datasets}
In accordance with previous practice \cite{DISCO_Ji2022ColorFormerIC}, we  evaluate our method on \textit{ImageNet ctest10k} \cite{ImageNet_Ctests_Larsson2016LearningRF}, a subset of ImageNet \cite{IMAGENET_5206848} that provides a balanced distribution of ImageNet categories, free of grayscale images. We tuned our method hyper-parameters (such as number of steps, prompts, etc) using \textit{ImageNet cval1k} \cite{ImageNet_Ctests_Larsson2016LearningRF} as a validation set (which is disjoint from imagenet ctest10k).  We additionally evaluate it on the COCO-Stuff \cite{COCO_STUFF_aesar2016COCOStuffTA} validation set of 5k images.
\begin{table}[t]
  \caption{Comparative analysis of different methods on the validation sets. Best performing in bold. Baseline numbers are from Disco \cite{DISCOXia2022DisentangledIC}, and UniColor \cite{huang2022unicolor}. Ours (Ranker) indicates using the auto ranker to choose colorfullness scale. *~denotes using the auto-generated captions from the ground truth color images and is provided for reference.}
  \label{tab:quantiative_comparison}
  \begin{tabular}{lcccc}
  
    \toprule
    \textbf{Method} & \multicolumn{2}{c}{\textbf{ImageNet (10k)}} & \multicolumn{2}{c}{\textbf{COCO-Stuff (5k)}} \\
    \cmidrule(lr){2-3} \cmidrule(lr){4-5}
     &  FID ($\downarrow$)  & $\Delta$ CLR ($\downarrow$)  &  FID ($\downarrow$) & $\Delta$ CLR ($\downarrow$) \\
    \midrule

    CIColor &  11.58 & \textcolor{white}{0}1.28 &  21.44 &  \textbf{\textcolor{white}{0}1.43}  \\
    UGColor &  \textcolor{white}{0}6.85 &  13.70 &  14.74 &   12.38   \\
    Deoldify &  \textcolor{white}{0}5.78 &  18.20 &  12.75 &   17.40  \\
    InstColor &  \textcolor{white}{0}7.35 &  16.07 & 12.24 & 11.64   \\
    ChromaGAN&  \textcolor{white}{0}9.60 &  12.27 &  20.57&   11.68 \\
    ColTran &  \textcolor{white}{0}6.37 &  \textcolor{white}{0}2.97 &   11.65 & \textcolor{white}{0}2.07  \\
    UniColor &  \textcolor{white}{0}9.43 &  \textcolor{white}{0}2.60 &   11.16 & \textcolor{white}{0}1.91  \\
    Disco &  \textcolor{white}{0}5.57 &  \textcolor{white}{0}9.81 &   10.59 &  11.82 \\
    BigColor &  \textcolor{white}{0}3.58 &  \textcolor{white}{0}1.41 &   \textcolor{white}{0}8.23 & \textcolor{white}{0}2.12 \\
    Ours (Scale=0.8) &  \textbf{\textcolor{white}{0}3.52} &  \textcolor{white}{0}8.35   & \textcolor{white}{0}8.17 & \textcolor{white}{0}9.59 \\
    Ours (Ranker) &  \textcolor{white}{0}3.69  &  \textbf{\textcolor{white}{0}0.21} &   \textbf{\textcolor{white}{0}8.05} &  \textcolor{white}{0}1.68  \\
    \midrule
    Ours (Ranker)* &  \textcolor{white}{0}2.60   & \textcolor{white}{0}1.62 &  \textcolor{white}{0}6.62 & \textcolor{white}{0}0.74 \\

    \bottomrule
  \end{tabular}
\end{table}

\subsection{Fully Automatic Colorization}
\label{exp:automatic_colorization}

Our LCDM model uses an input text prompt to better control the colorization result and for personalization. However, we also want to allow a fully automated prompt-free pipeline. Therefore, we explore various strategies to automatically create the most effective prompts, those with best performance on standard benchmarks.
We allow using prompts for this goal as long as they are computed fully automatically given only the input grayscale image. One way is to use image captioning methods (e.g. CLIP-interrogator\footnote{\url{https://github.com/pharmapsychotic/clip-interrogator}}). This approach risks encouraging the model to generate yet another grayscale image as the caption would often include phrases such as ``black and white''. To mitigate this we explore various approaches:
(1) \textit{Colorization CLIP Direction:} Auto generate captions from grayscale and color images, and compute mean residual vector between text CLIP embeddings. Add this residual to auto generated text embedding at inference time.
(2) \textit{Remove Grayscale Hints:} Identify phrases that indicate grayscale photos; remove these phrases from the auto generated captions.
(3) \textit{LLM Rephrasing:} Utilize a Large Language Model \cite{GPT_Brown2020LanguageMA} for rephrasing captions, where the LLM was prompted to exclude any hints related to grayscale images.
(4) \textit{Optimal Negative Prompt:} Identify a constant optimal negative prompt while disregarding the grayscale image caption. Note that this approach does not require any processing at inference time as the negative prompt is constant and the positive prompt is the null-text prompt. 

\begin{figure}
\setlength{\tabcolsep}{1pt}
\newlength{\wwc}
\setlength{\wwc}{0.23\linewidth}

\begin{tabular}{cccc}
  \centering
\includegraphics[width=\wwc]{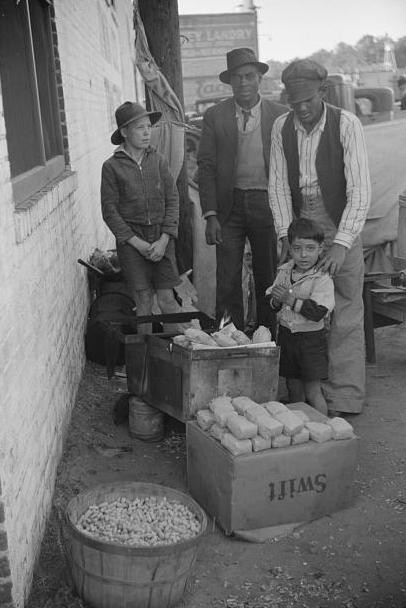} &
\includegraphics[width=\wwc]{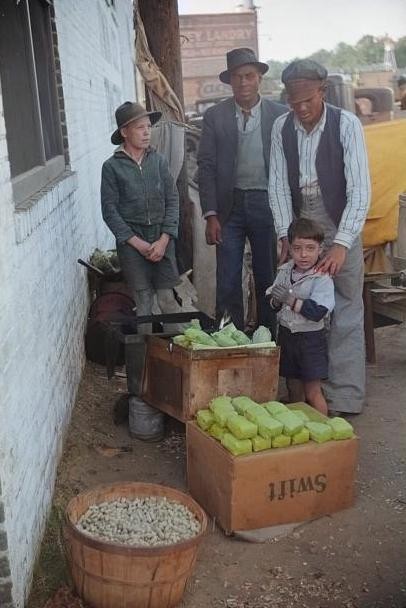} &
\includegraphics[width=\wwc]{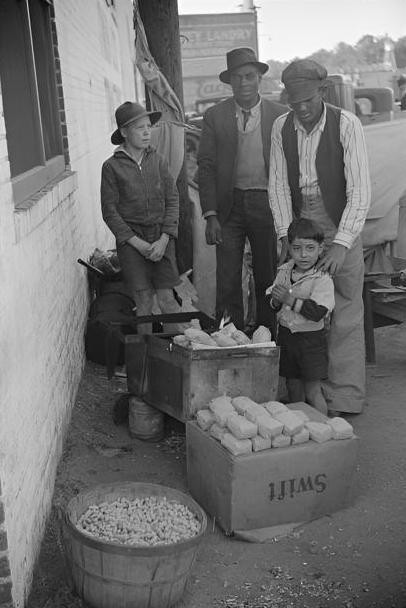} &
\includegraphics[width=\wwc]{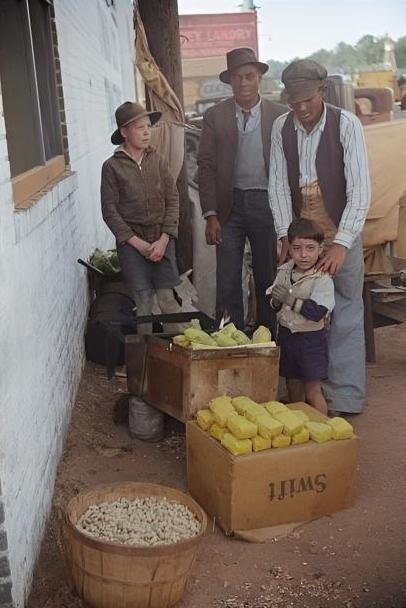} \\
\includegraphics[width=\wwc]{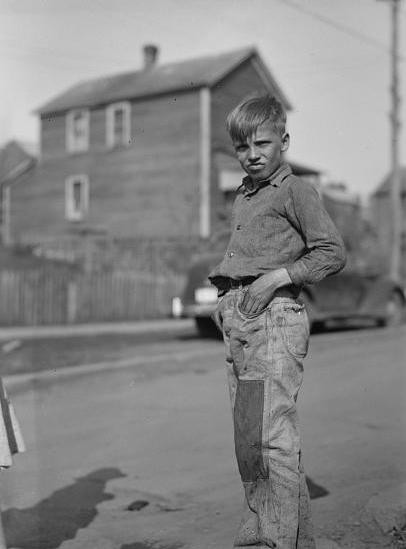} &
\includegraphics[width=\wwc]{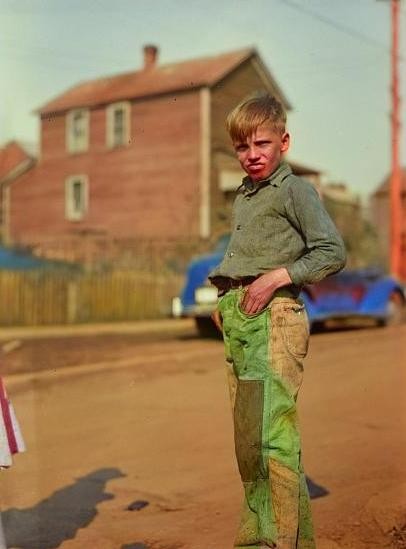} &
\includegraphics[width=\wwc]{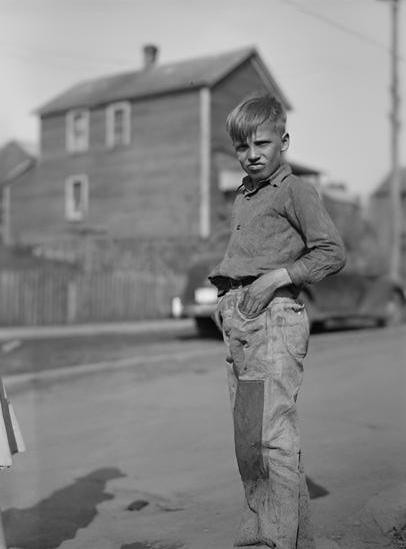} &
\includegraphics[width=\wwc]{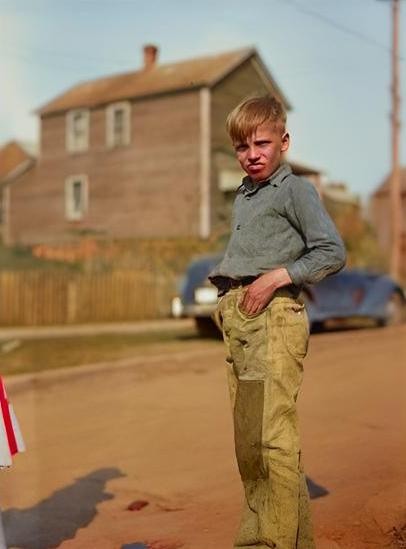} \\
\includegraphics[width=\wwc,height=2cm]{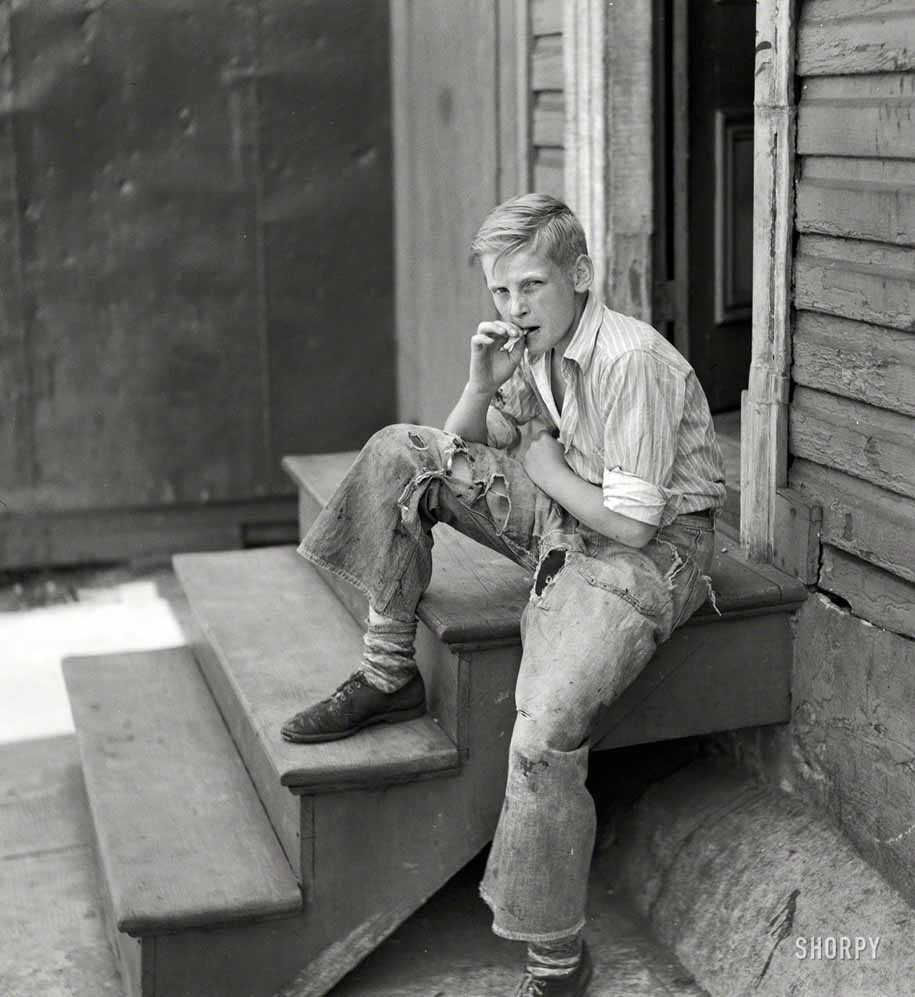} &
\includegraphics[width=\wwc,height=2cm]{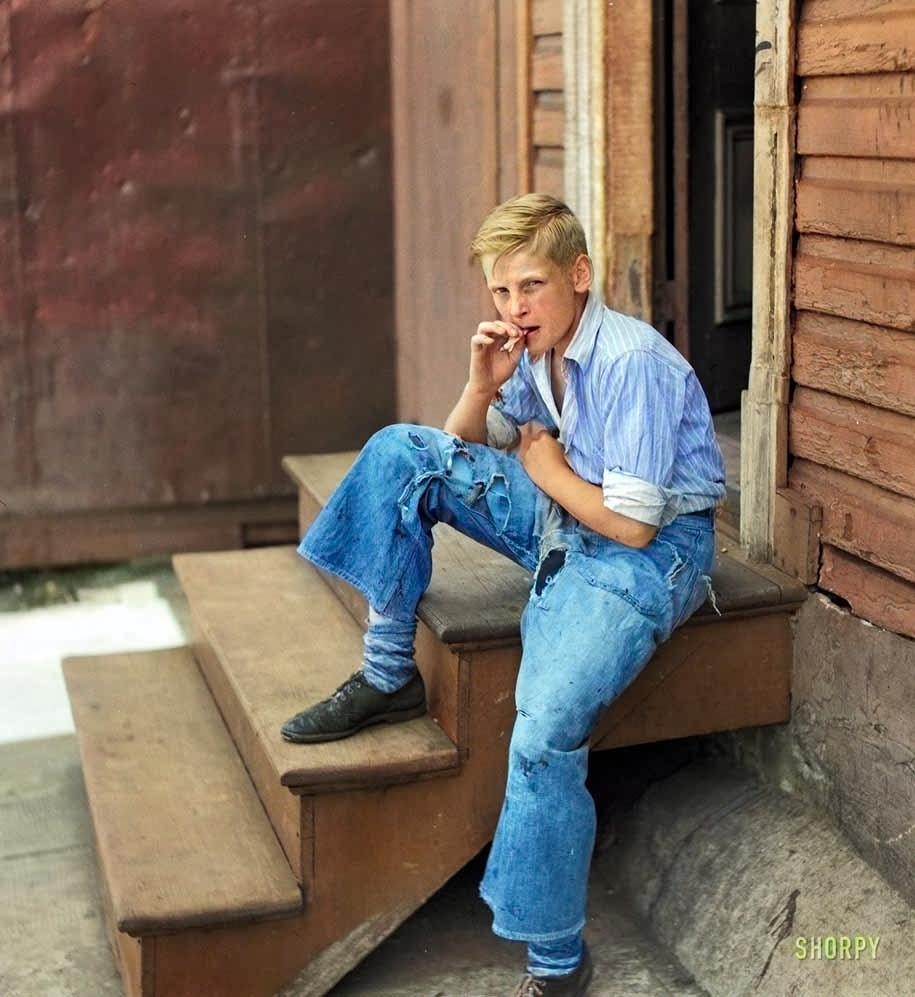} &
\includegraphics[width=\wwc,height=2cm]{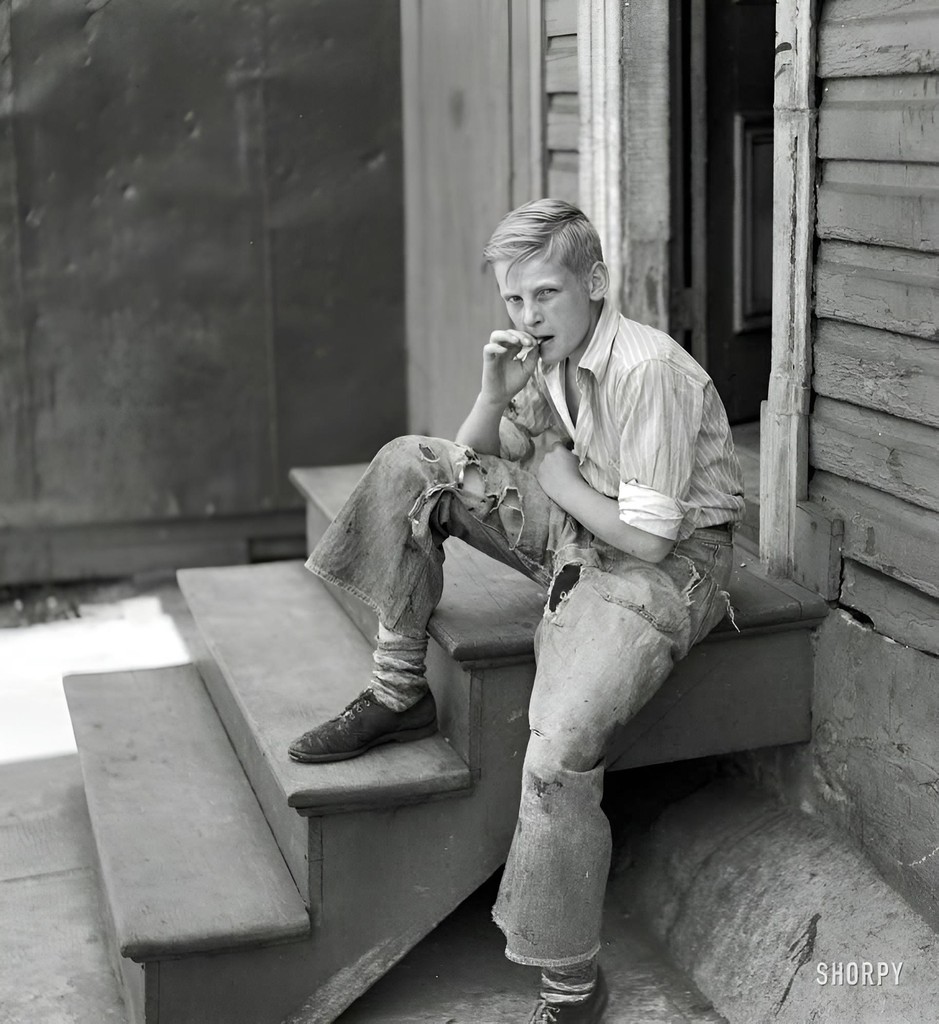} &
\includegraphics[width=\wwc,height=2cm]{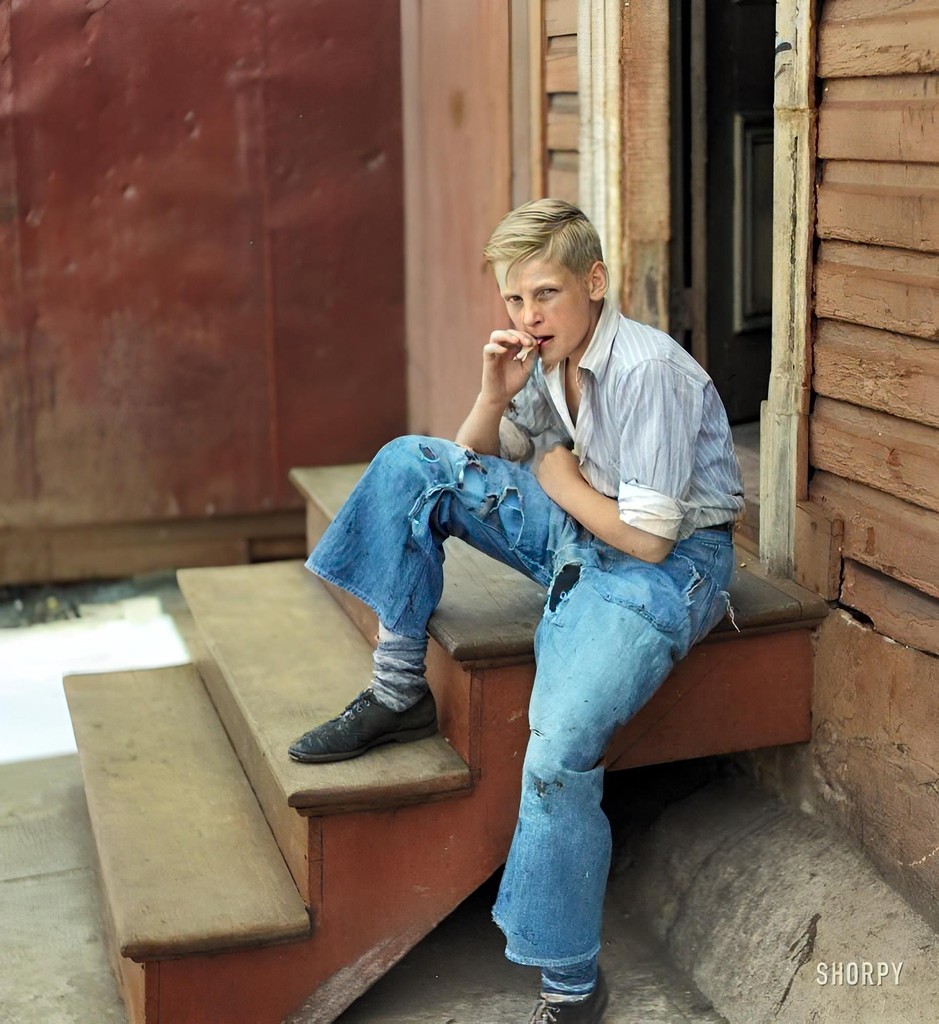}

\\ 

Input & Color & Restore  & Restore-Color

\end{tabular}

\caption{
\textbf{Colorization in Image Restoration Pipelines.} We use an image restoration model \cite{ESRGAN_wang2021gfpgan} followed by applying our image colorization. The colorization we achieve with this technique is better than when colorizing the non-restored images, as evidenced by the colorization on the child's pants on the sceond row, or the color bleeding on the wall in the last row. Image credits: ©Wolcott Marion, ©Office of War Information Photograph Collection, ©John Vachon.} 
\label{fig:interactive_restoration}

\end{figure}

Full descriptions and quantitative results of each approach showing that all four improve over the baseline grayscale captioning are provided in the supplemental material. 
For the first three strategies, adding a textual ``color direction'' in CLIP space (in a similar manner to StyleGan-NADA \cite{Gal2021StyleGANNADACD}) slightly improves the FID score, and rephrasing the grayscale caption either by hard-coding a list of words to discard or with the help of an LLM produces  better results than the ``color direction'' approach. 
Somewhat surprisingly, we got the best results by completely ignoring the image content and feeding it to the network with a constant null text, while using Classifier-Free Guidance (CFG)~\cite{Ho2022ClassifierFreeDG} with a constant negative prompt which we selected based on the most common phrases in the grayscale captions. 
We hypothesise that using a positive prompt tied to a specific image might lead the model to color the image in a less natural way than the null-text option;
however, a constant ``anti-grayscale'' prompt stirs the model away from the space of black-and-white images, and presumably, into the space of more prototypical color images. 

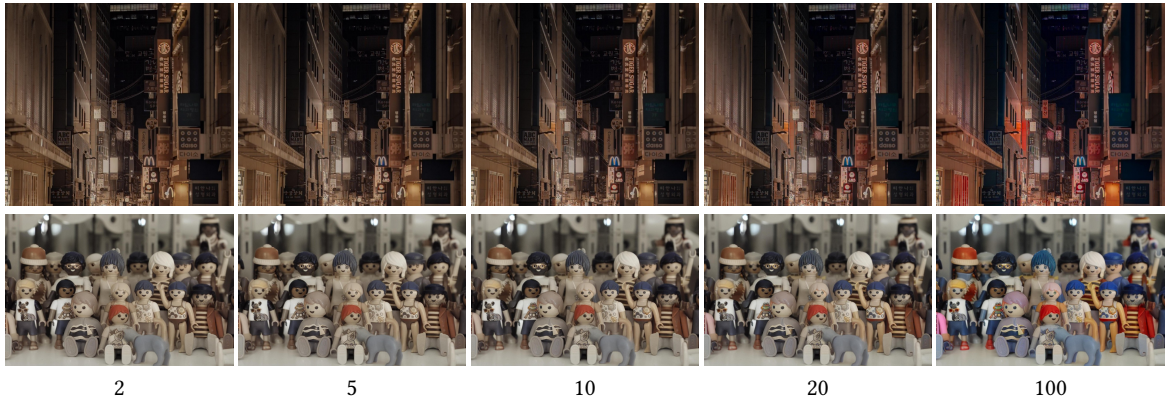
\begin{figure*}

\newlength{\wwp}
\setlength{\wwp}{0.170\linewidth}

\newcommand{\mytikzspy}[1]{
\begin{tikzpicture}[spy using outlines={rectangle, blue, magnification=3, size=1.2cm, connect spies}]
    \node{\includegraphics[width=\wwp]{#1}};
    \spy on (0.45, -0.75) in node [right] at (0.6, 0.62);
\end{tikzpicture}
}

\begin{tabular}{ccccc}
 
   \includegraphics[width=\wwp]{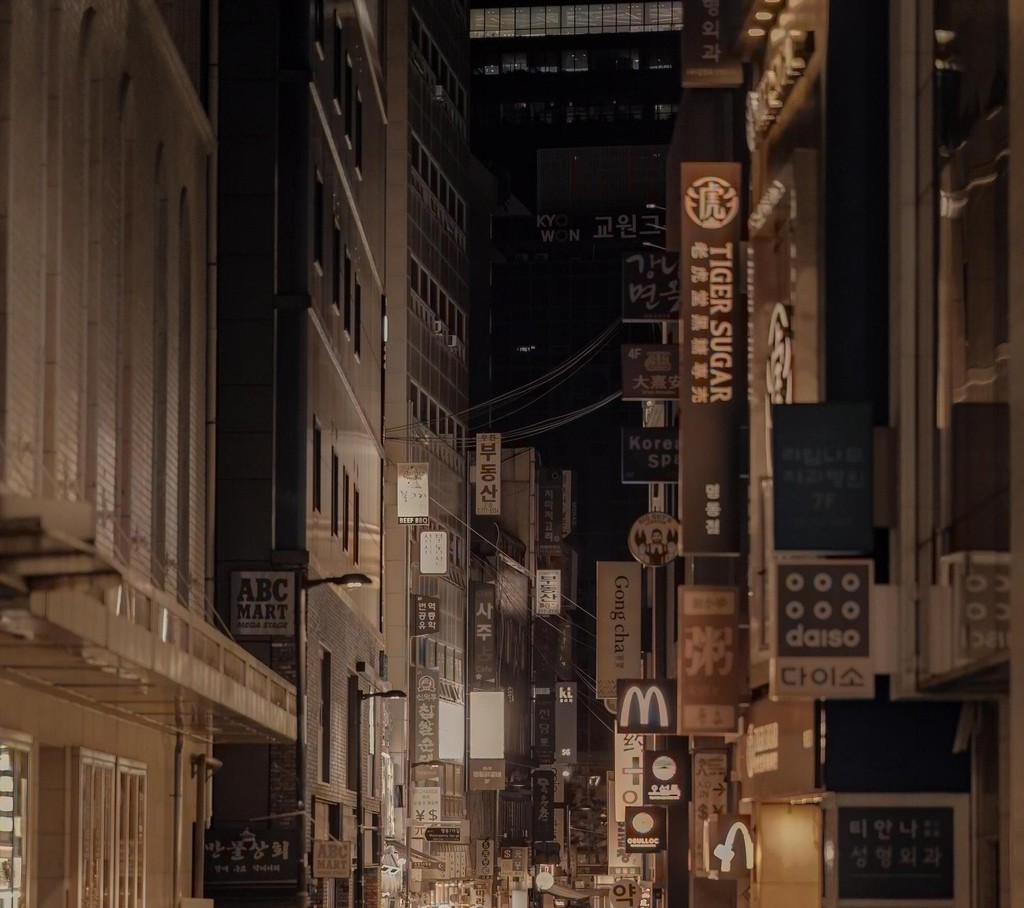} & \includegraphics[width=\wwp]{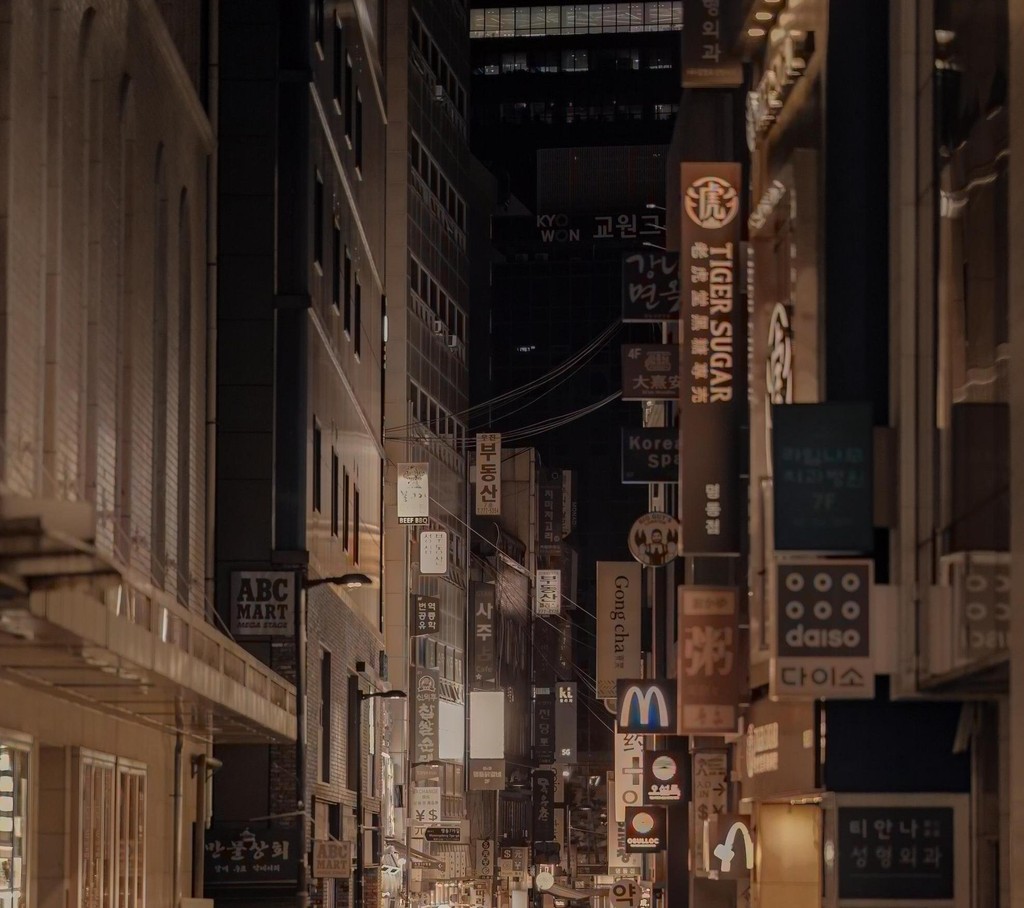} & \includegraphics[width=\wwp]{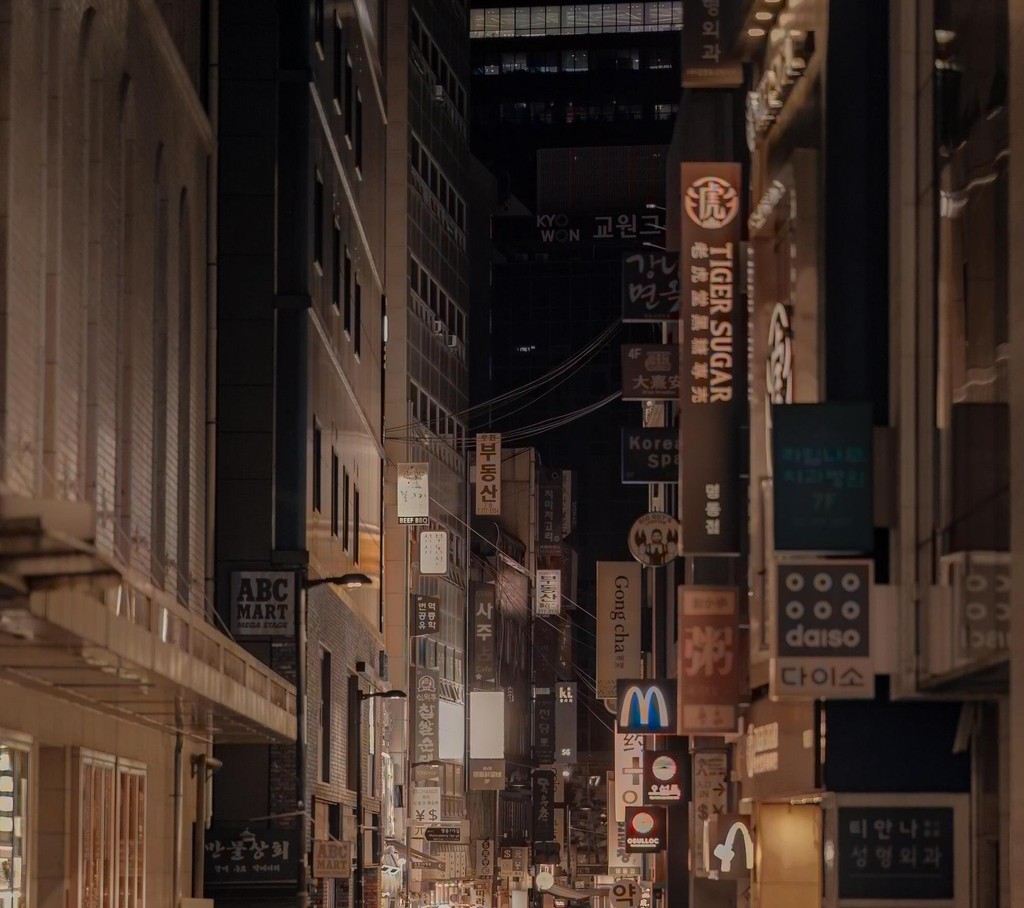} & \includegraphics[width=\wwp]{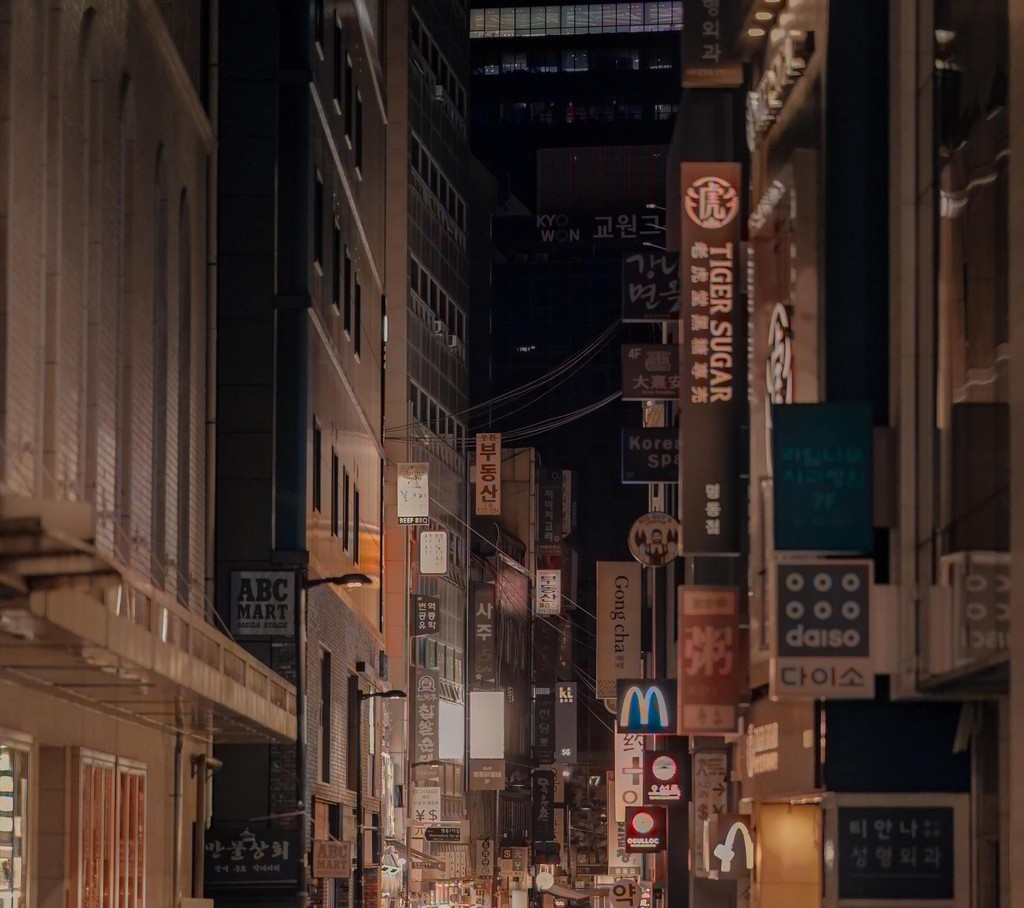} & \includegraphics[width=\wwp]{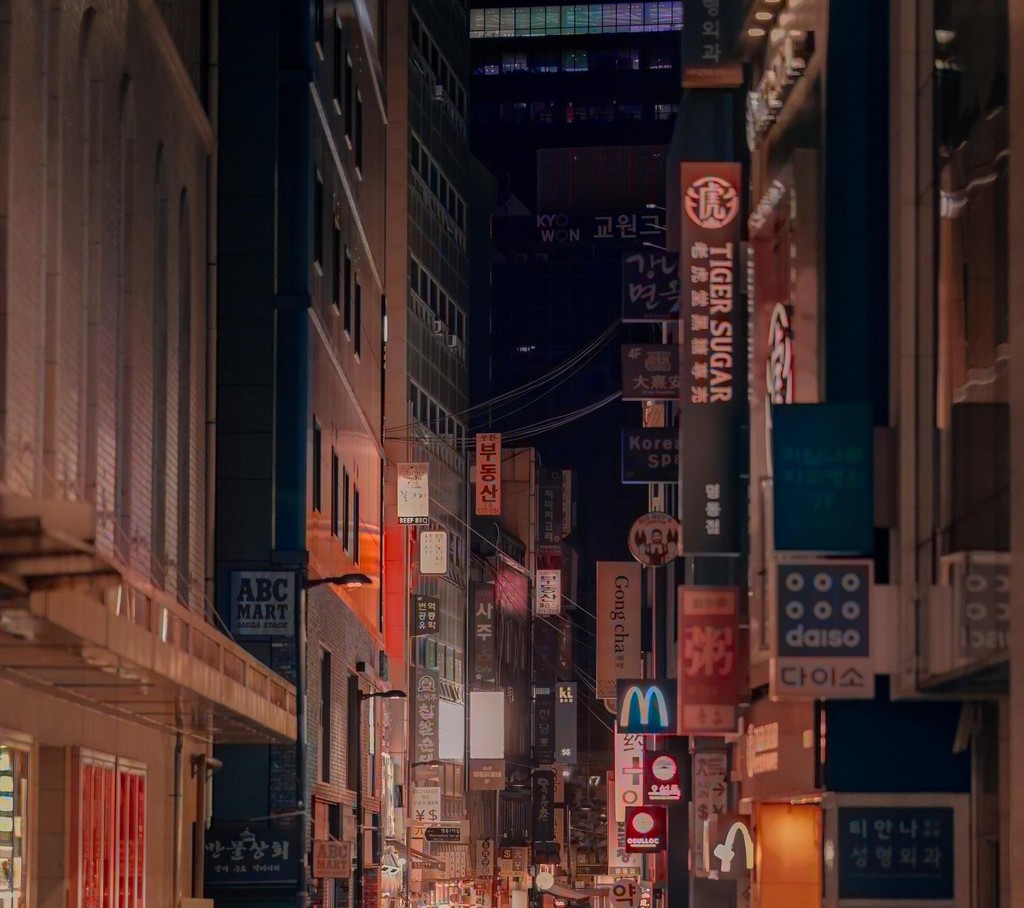} \\
 \includegraphics[width=\wwp]{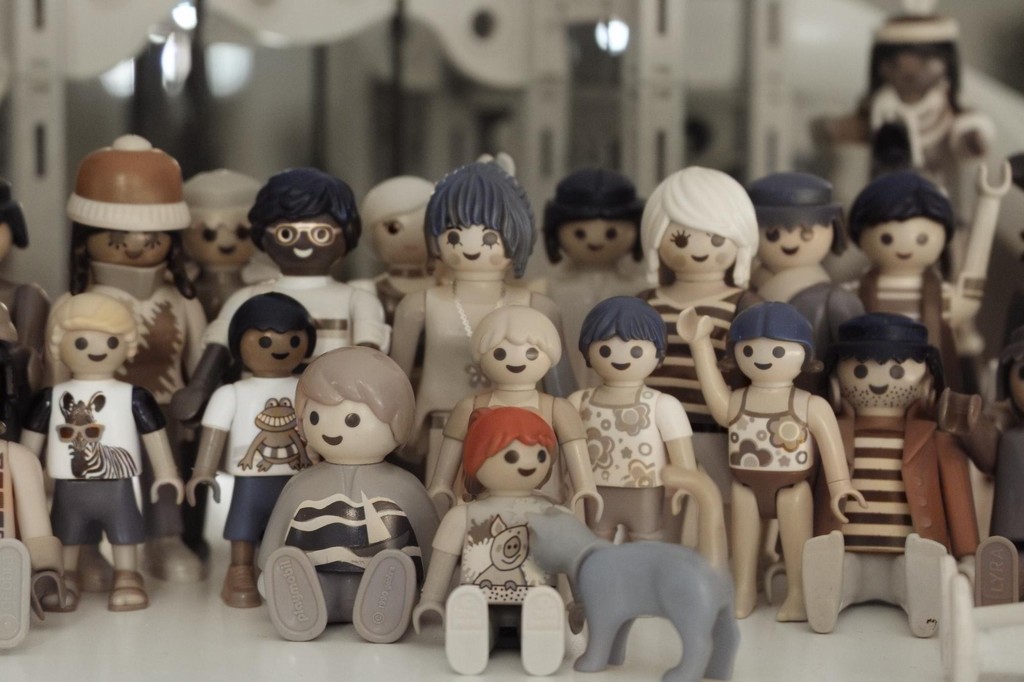} & \includegraphics[width=\wwp]{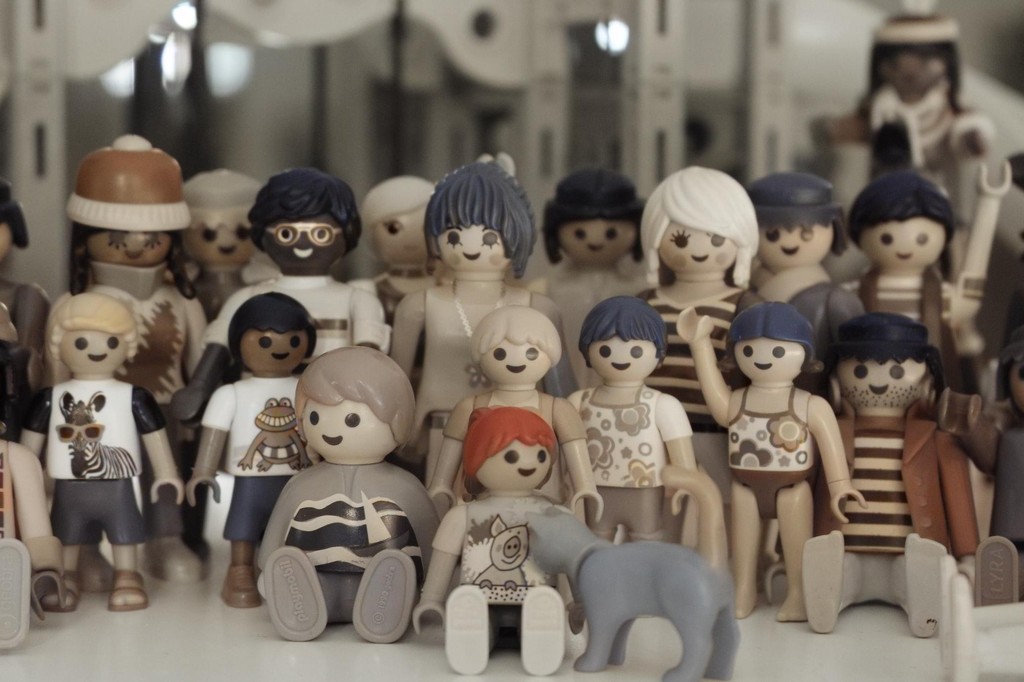} & \includegraphics[width=\wwp]{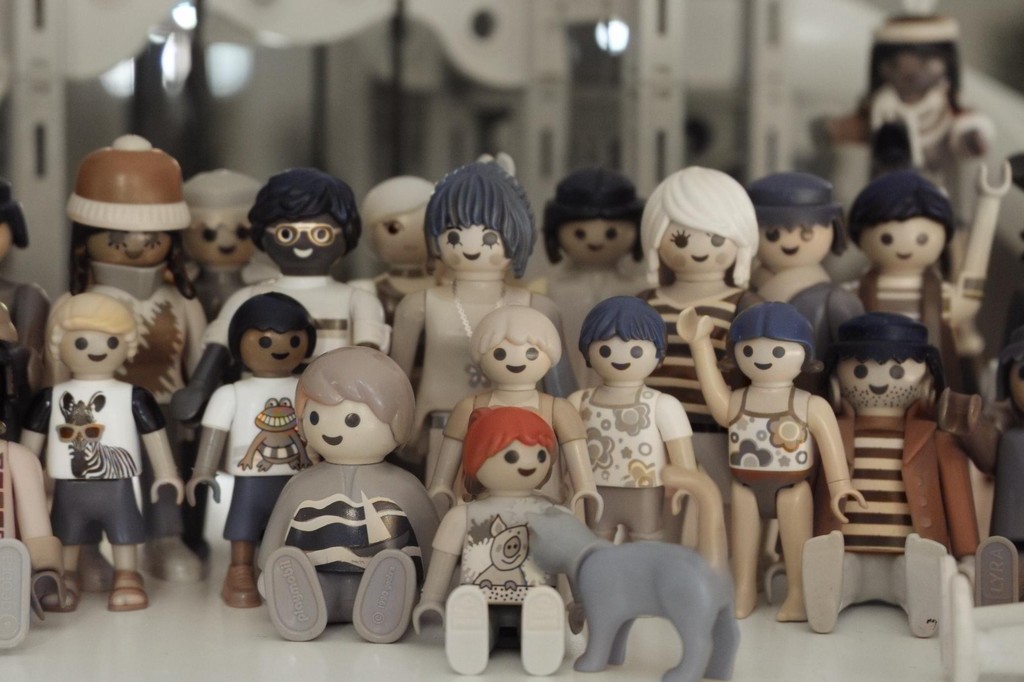} & \includegraphics[width=\wwp]{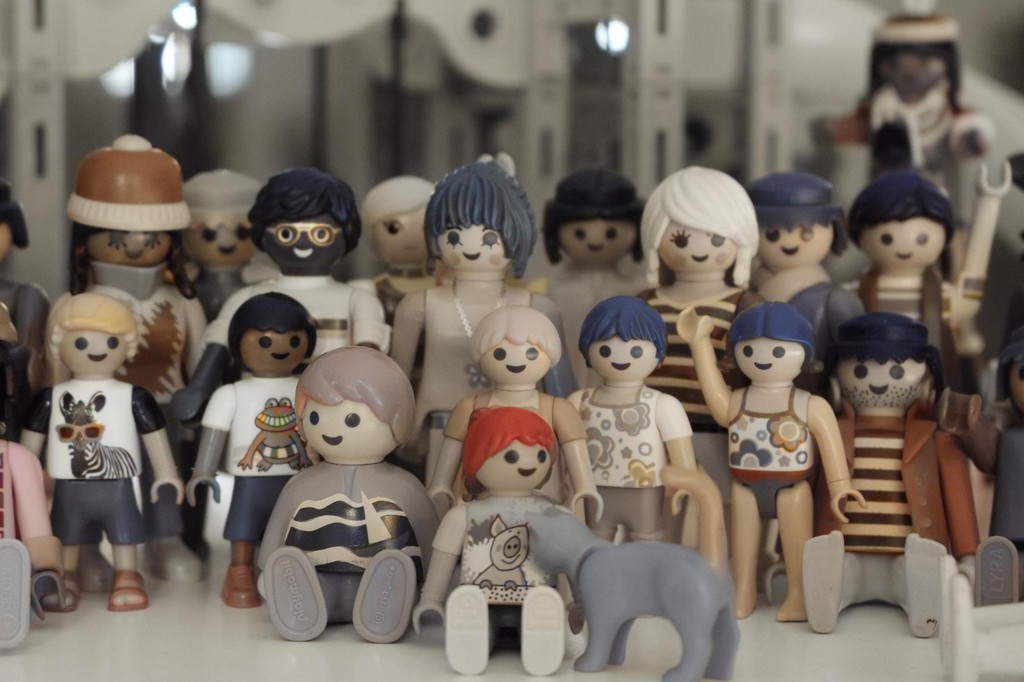} & \includegraphics[width=\wwp]{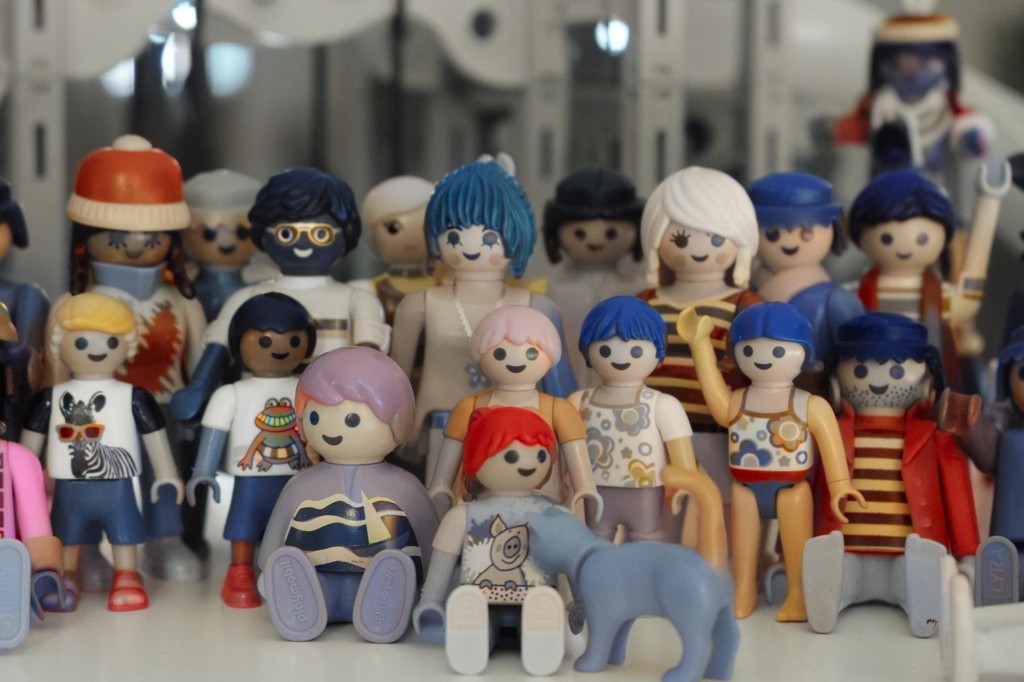} \\
 
 2 & 5 & 10 & 20 & 100  \\

\end{tabular}
\caption{ \textbf{Impact of Number of Steps on Colorization Results.} We showcase the impact of varying the number of steps from 2 to 100 (from the leftmost to the rightmost column) on the inference process. Fewer steps accelerate the inference time, whereas using larger number of steps results in bolder colors with higher variance and an expanded color palette. The first row highlights the enhancement of colors on the street light, while the second row demonstrates the introduction of additional colors to the collection of toys. Image credits: Unsplash ©zero take, Unsplash ©Teo Zac. }

\label{fig:stride_effect_fig}
\end{figure*}

\subsection{Results}
\label{exp:results}
\subsubsection{Evaluation Metrics.} 
\begin{figure*}
\setlength{\tabcolsep}{1pt}
\def\arraystretch{0.8}

\newlength{\ww}
\setlength{\ww}{0.121\linewidth}

\begin{tabular}{ccccccc}
  \centering

\includegraphics[width=\ww]{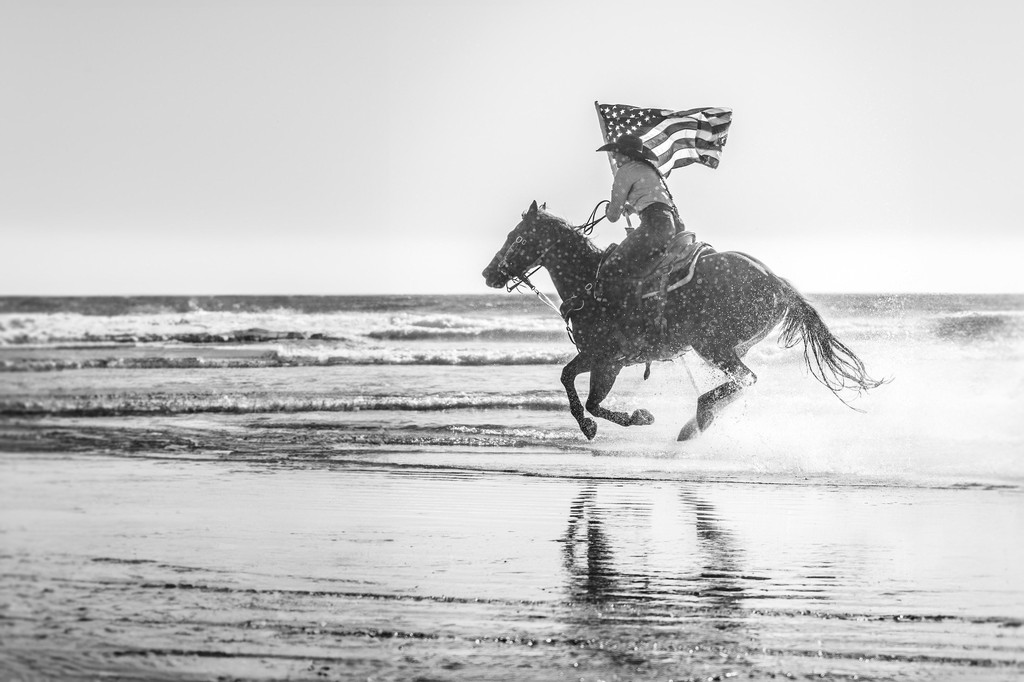} &
\includegraphics[width=\ww]{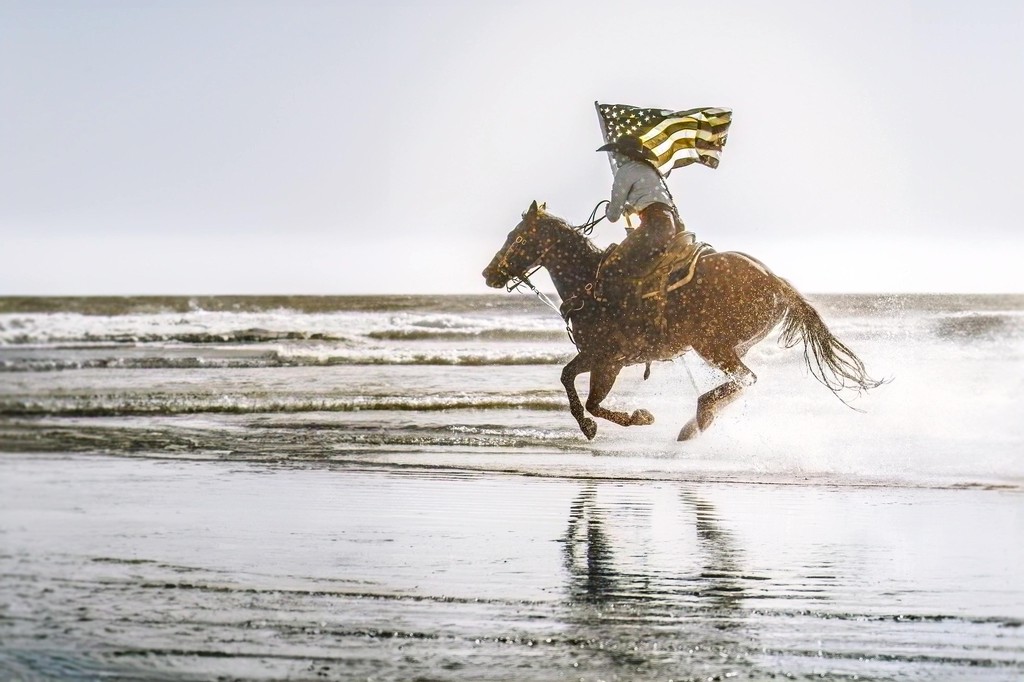} &
\includegraphics[width=\ww]{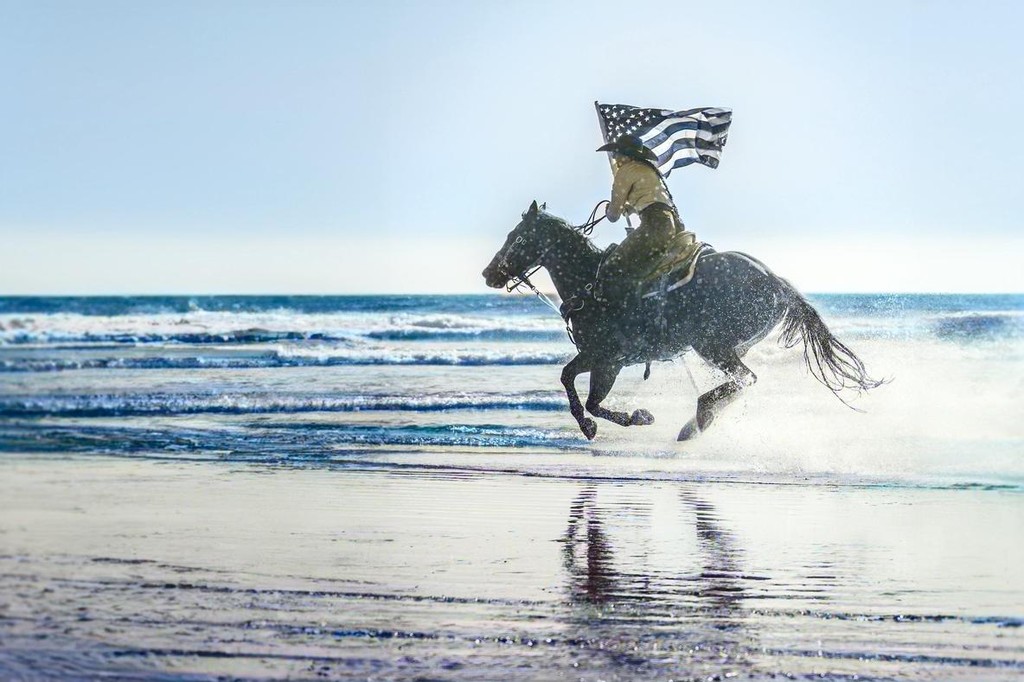} &
\includegraphics[width=\ww]{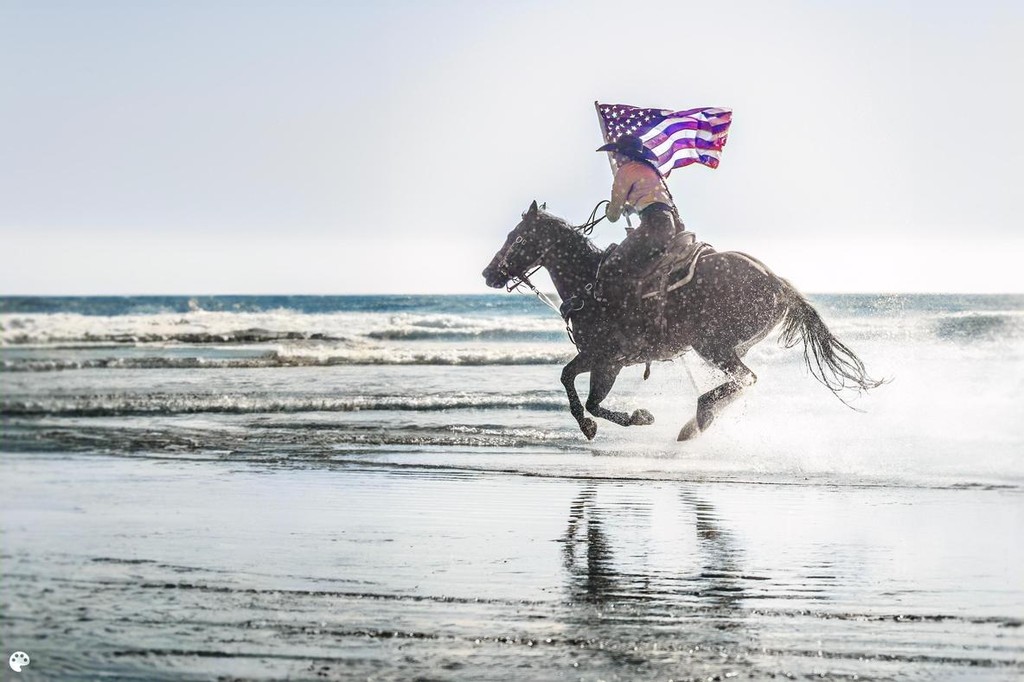} &
\includegraphics[width=\ww]{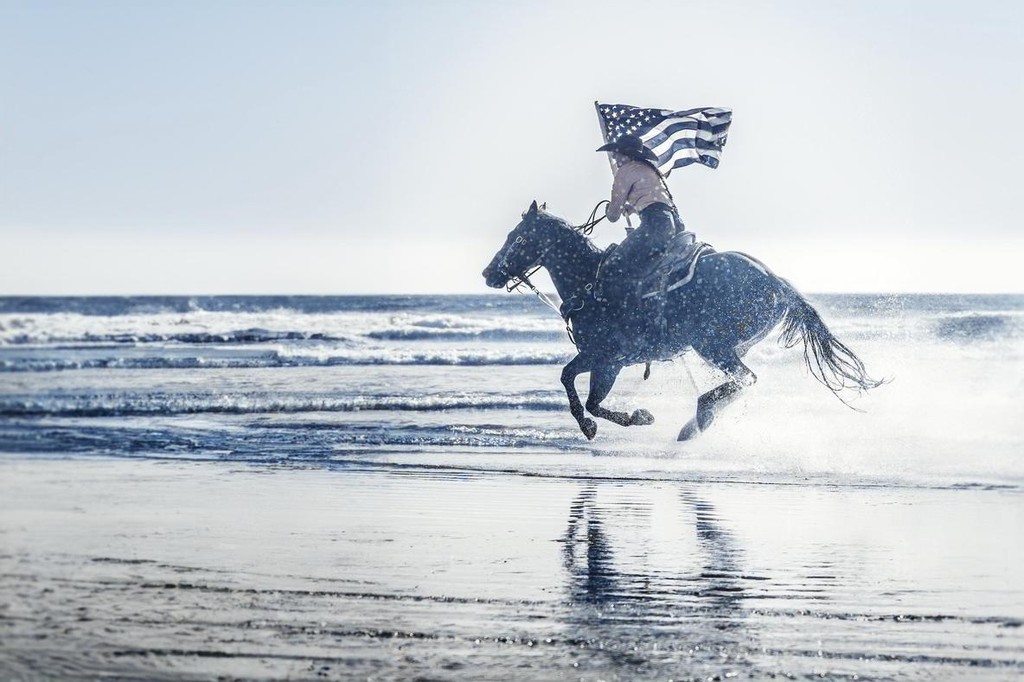} &
\includegraphics[width=\ww]{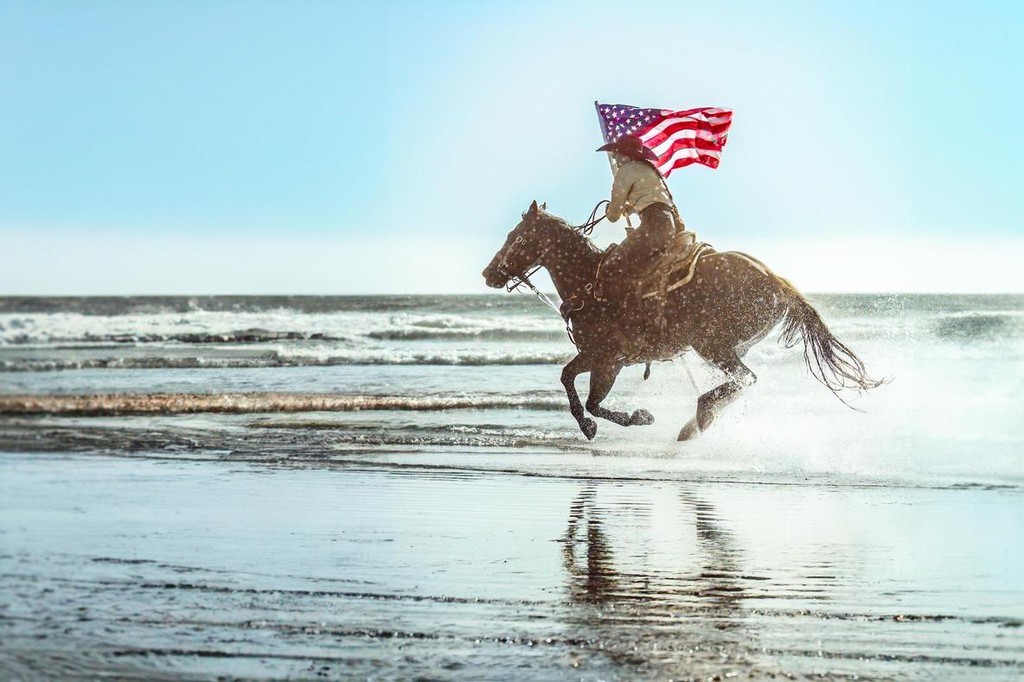} &
\includegraphics[width=\ww]{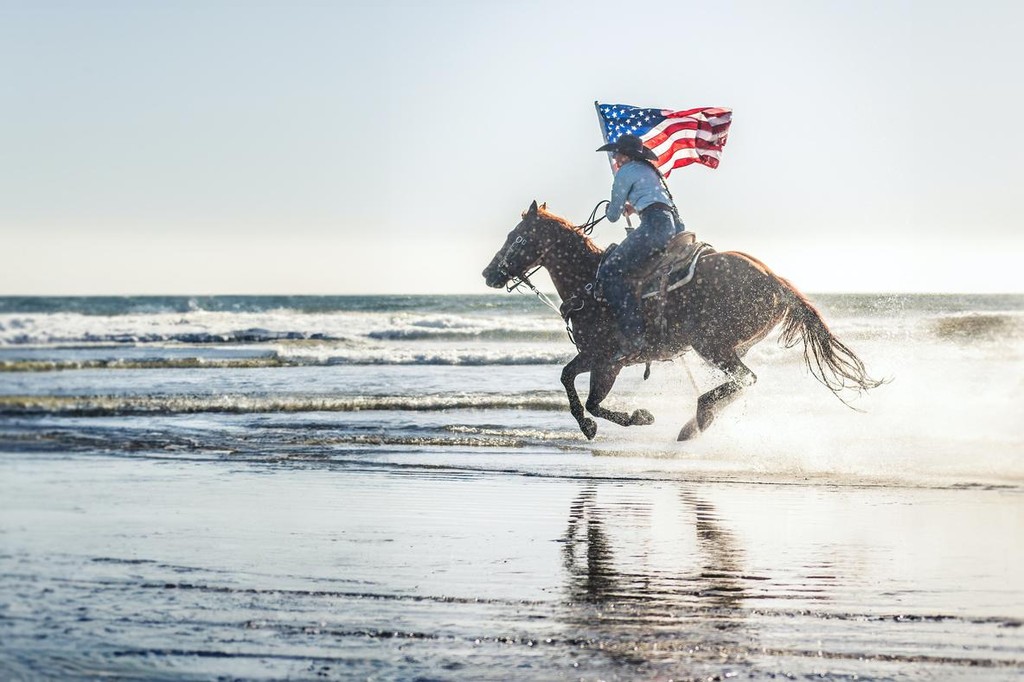}\\

\includegraphics[width=\ww]{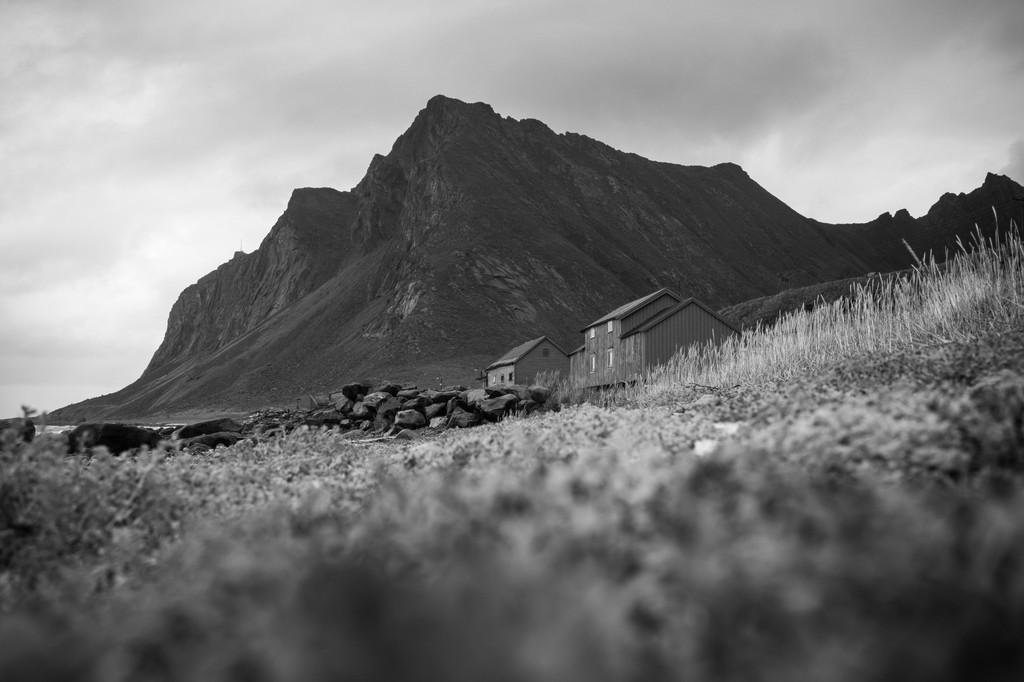} &
\includegraphics[width=\ww]{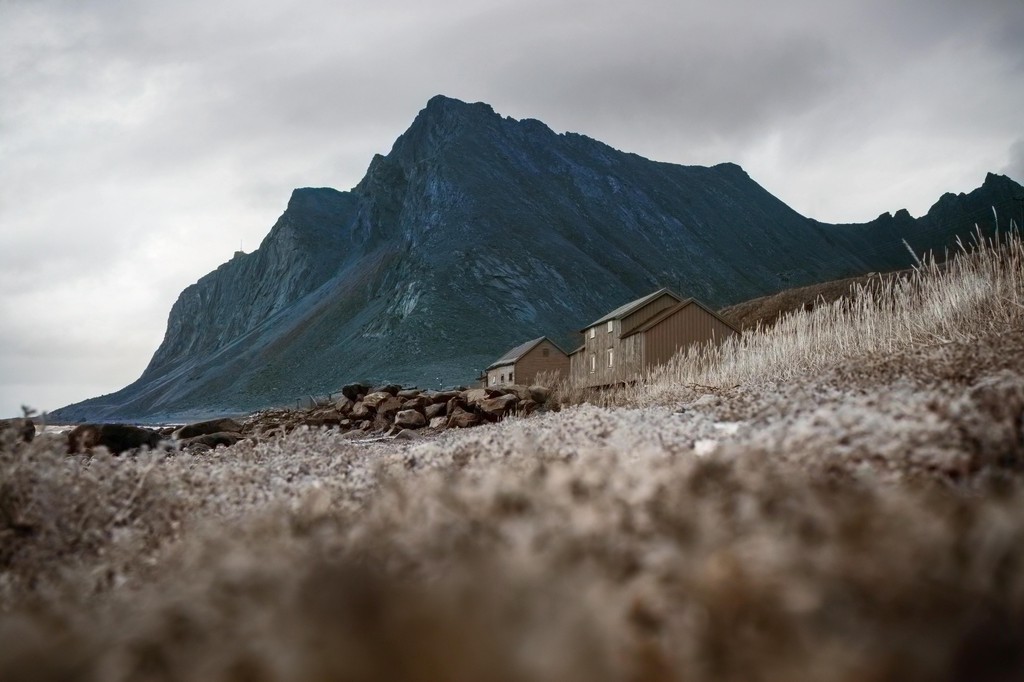} &
\includegraphics[width=\ww]{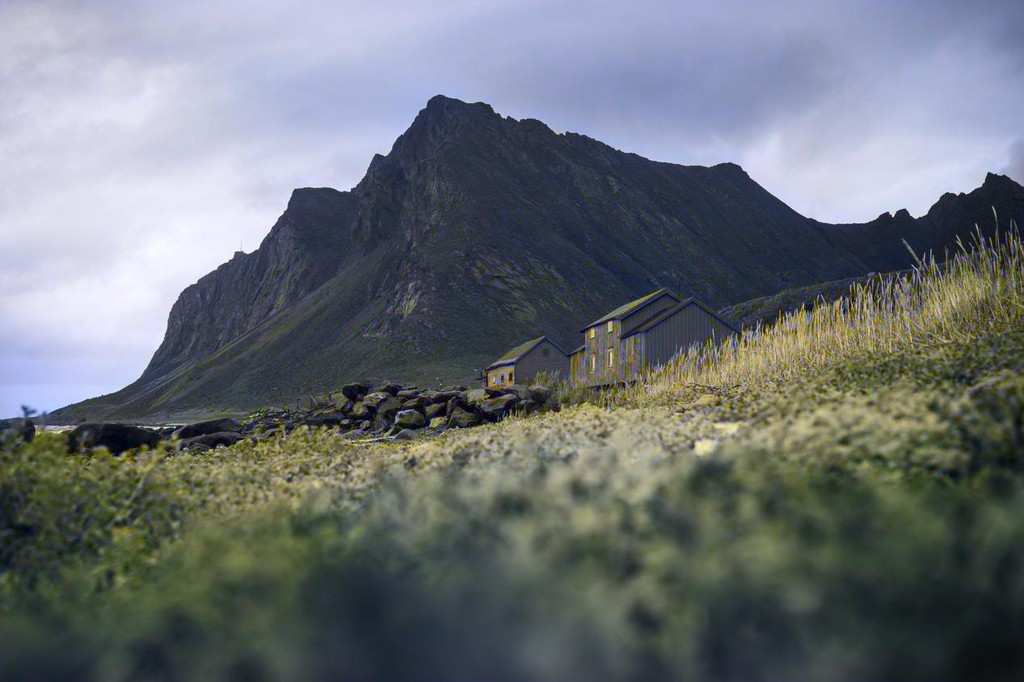} &
\includegraphics[width=\ww]{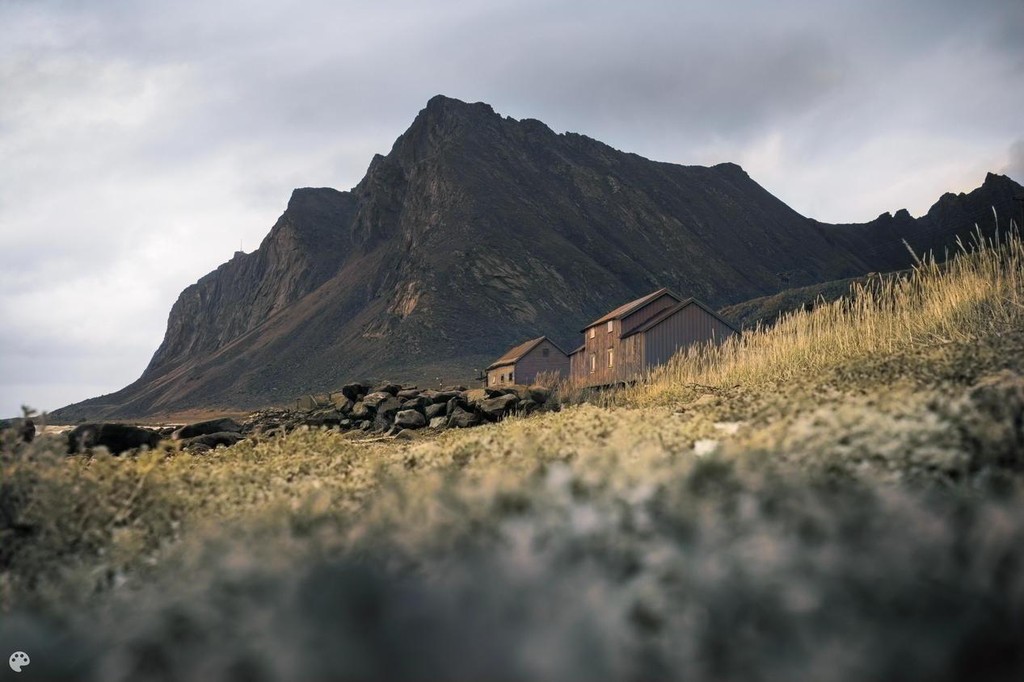} &
\includegraphics[width=\ww]{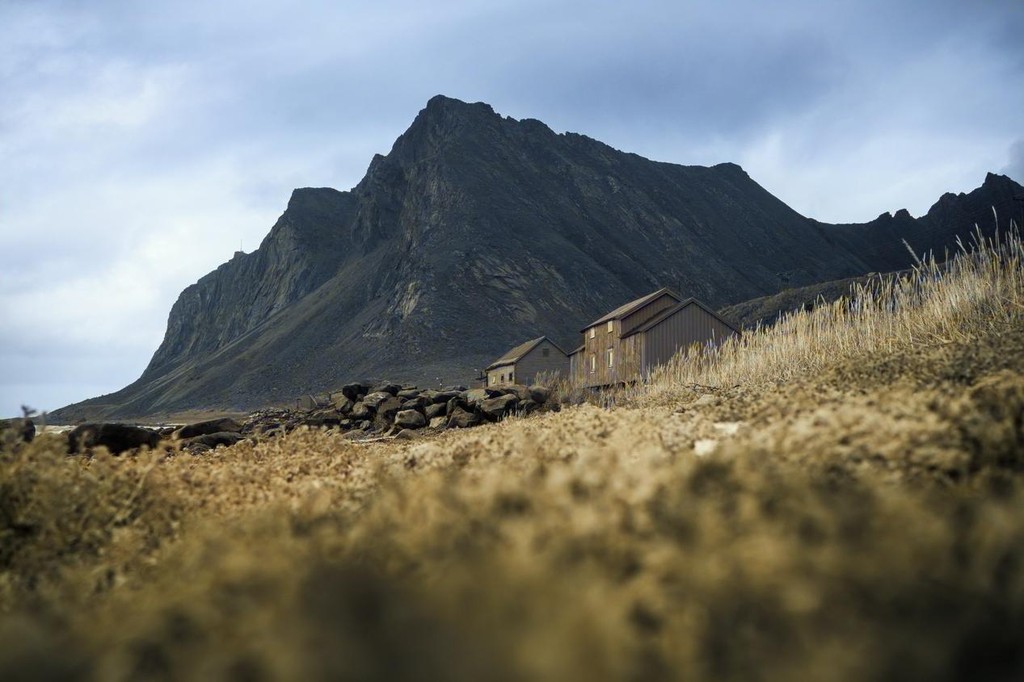} &
\includegraphics[width=\ww]{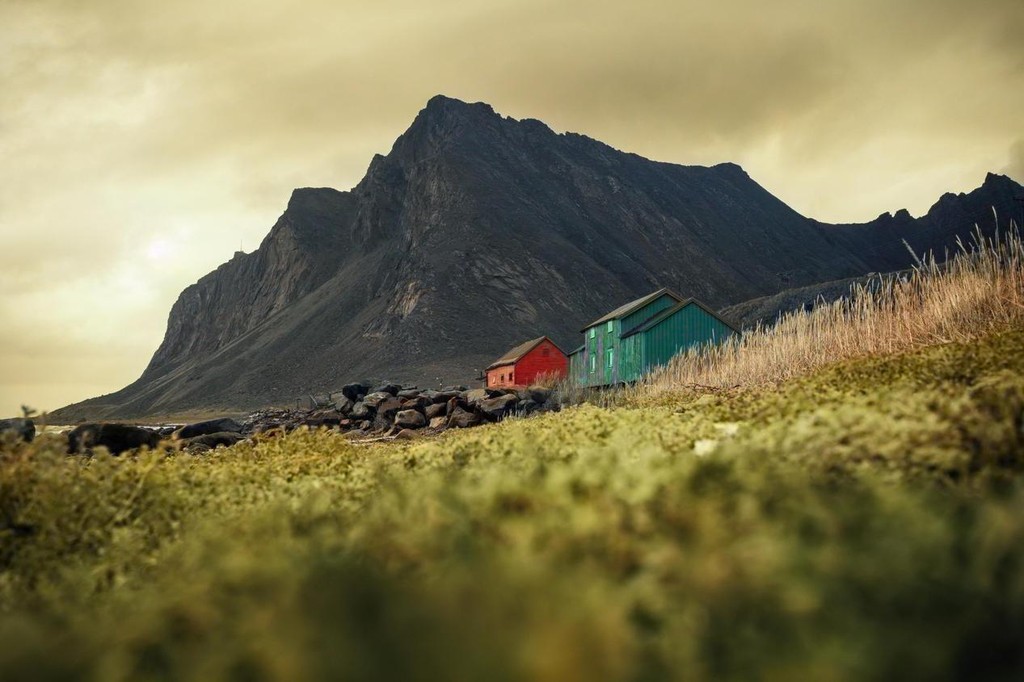} &
\includegraphics[width=\ww]{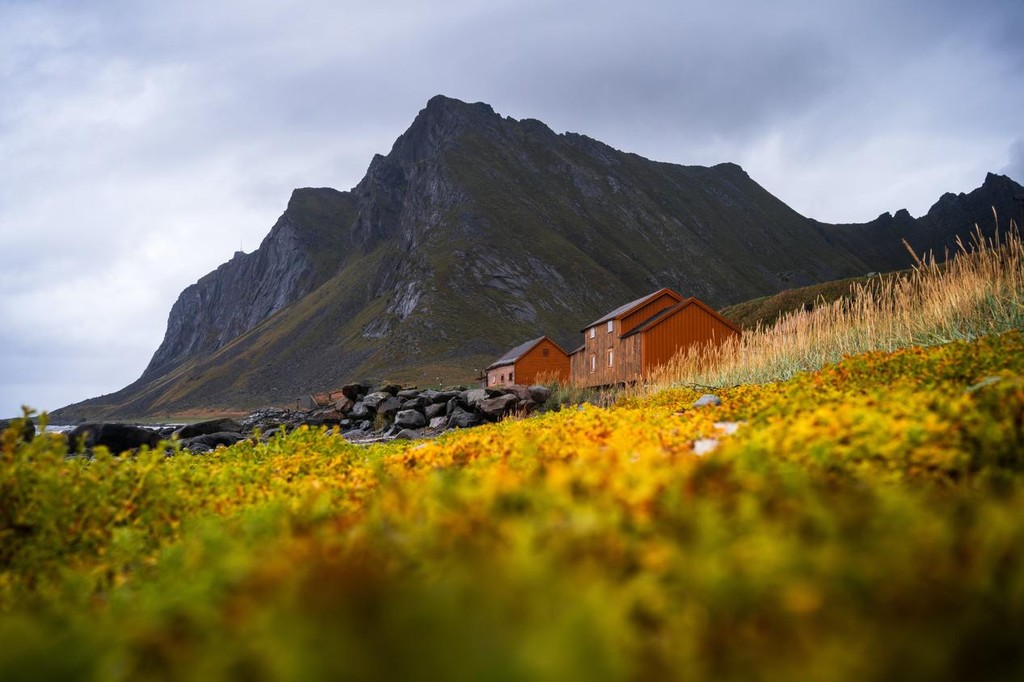}\\

\includegraphics[width=\ww]{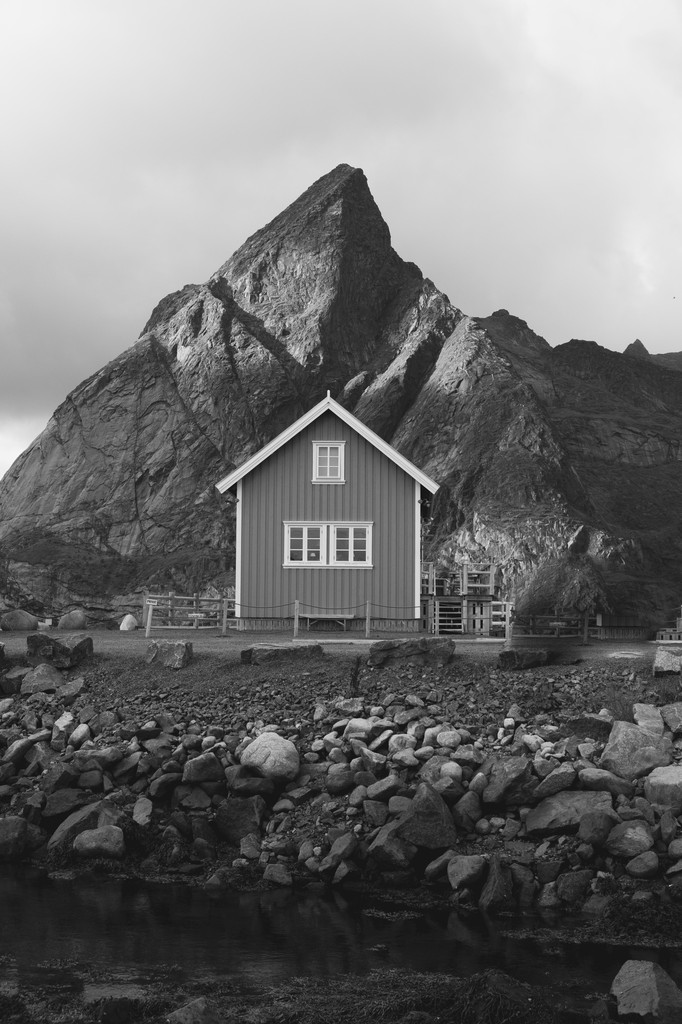} &
\includegraphics[width=\ww]{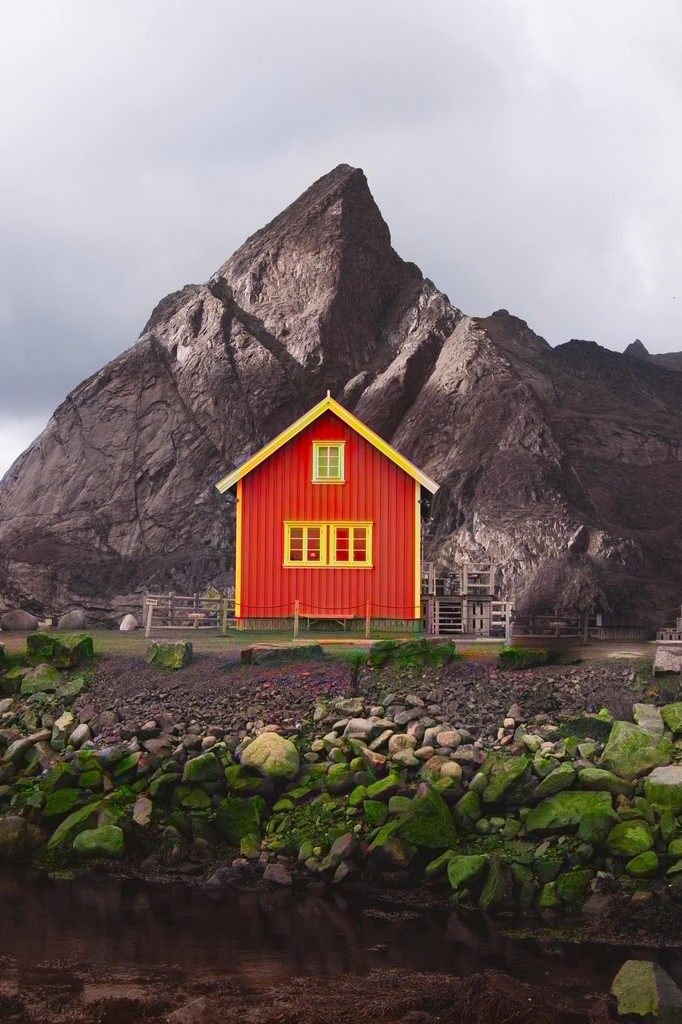} &
\includegraphics[width=\ww]{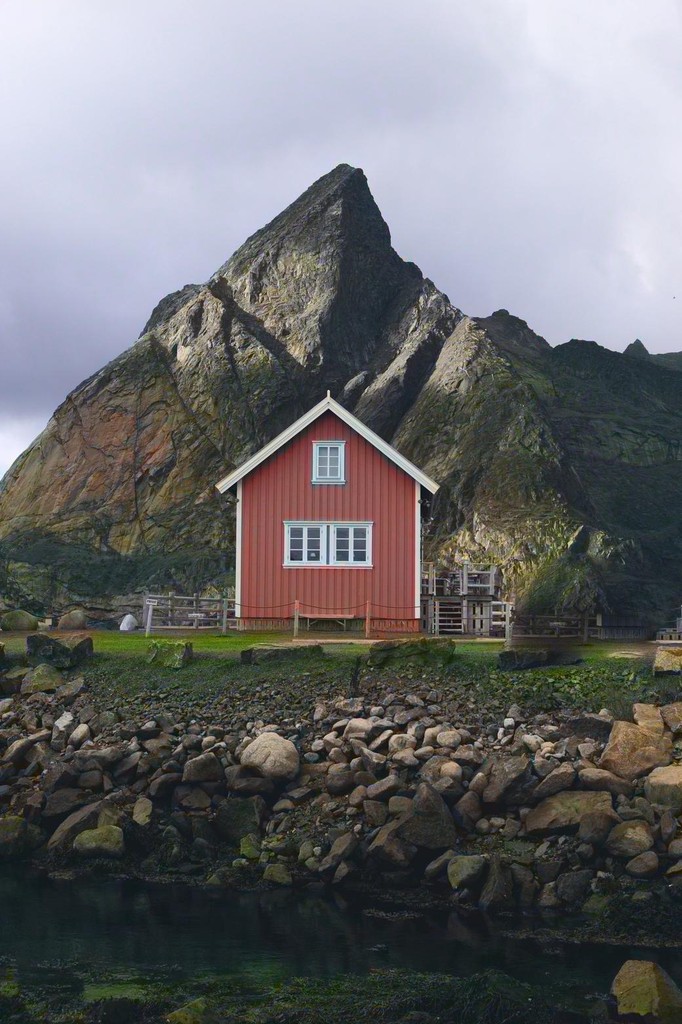} &
\includegraphics[width=\ww]{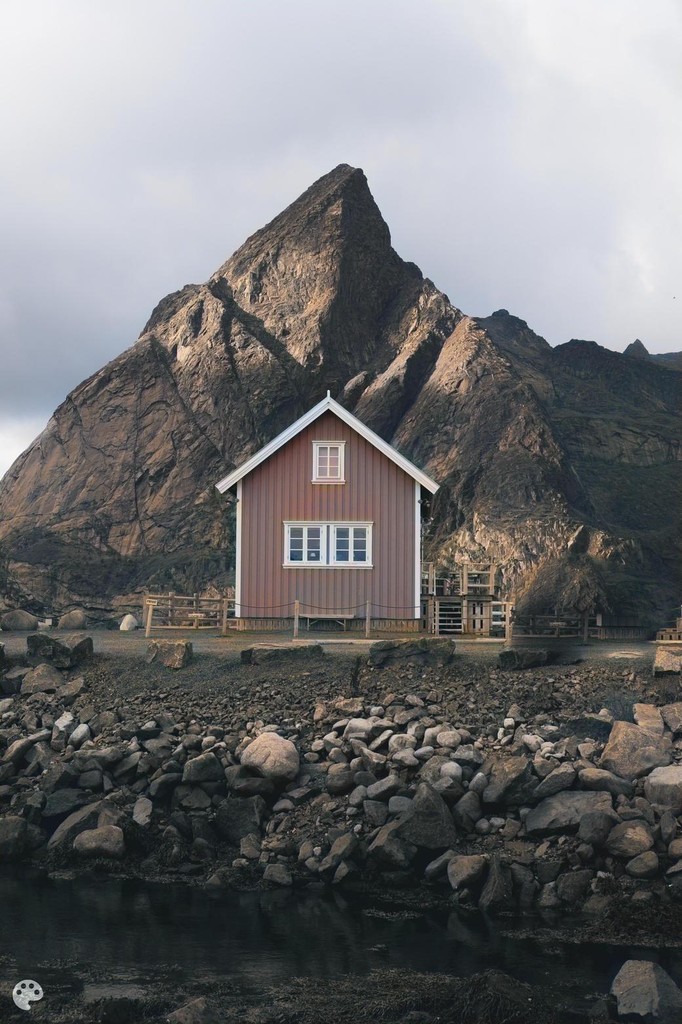} &
\includegraphics[width=\ww]{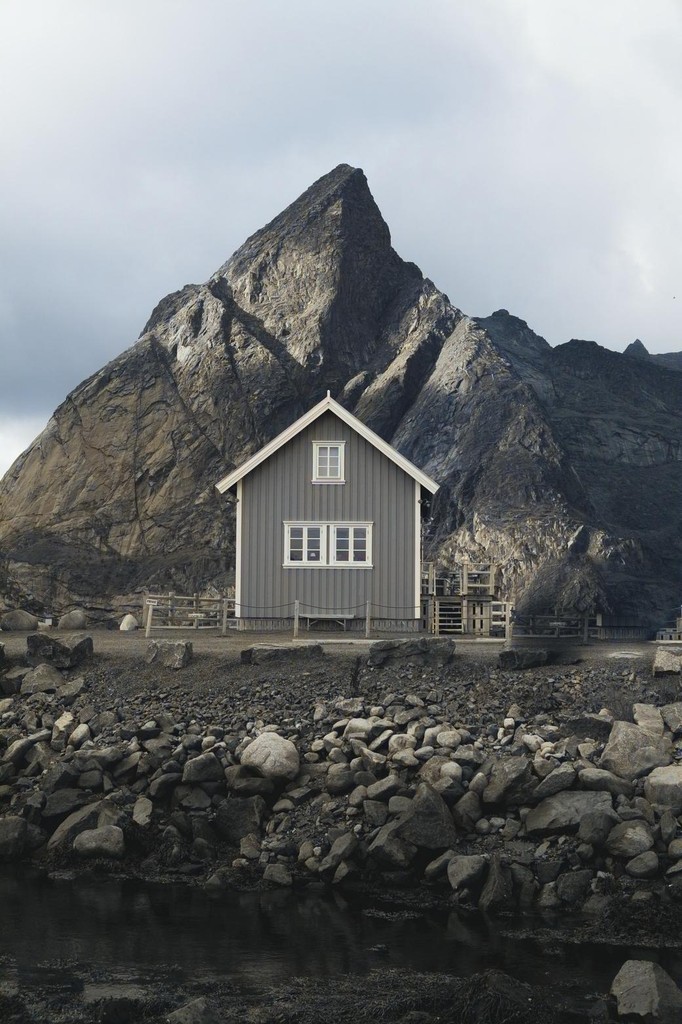} &
\includegraphics[width=\ww]{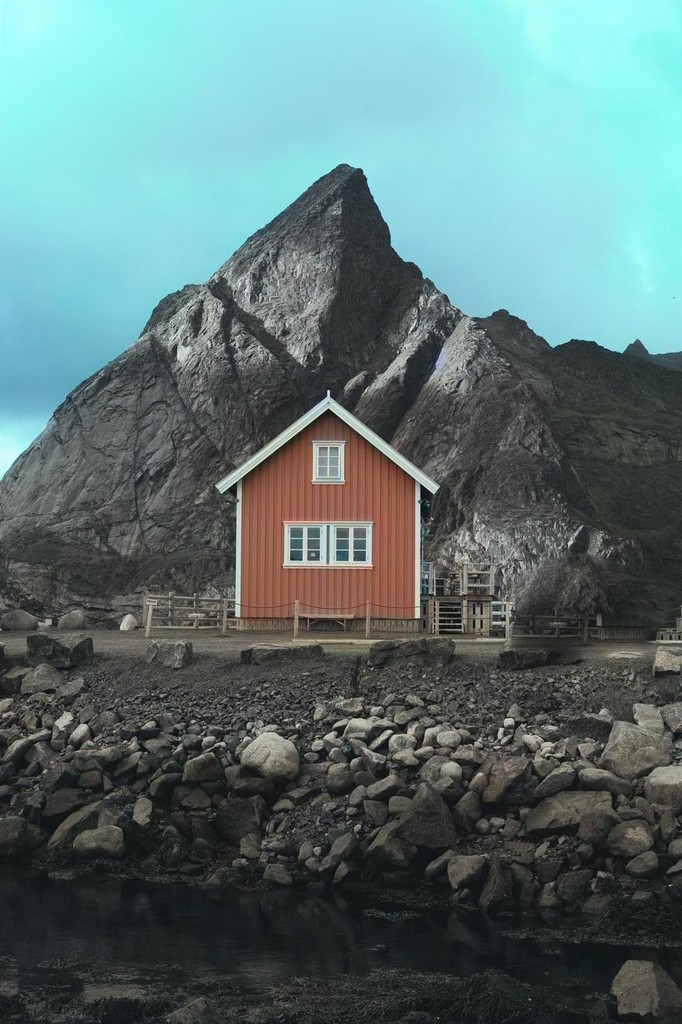} &
\includegraphics[width=\ww]{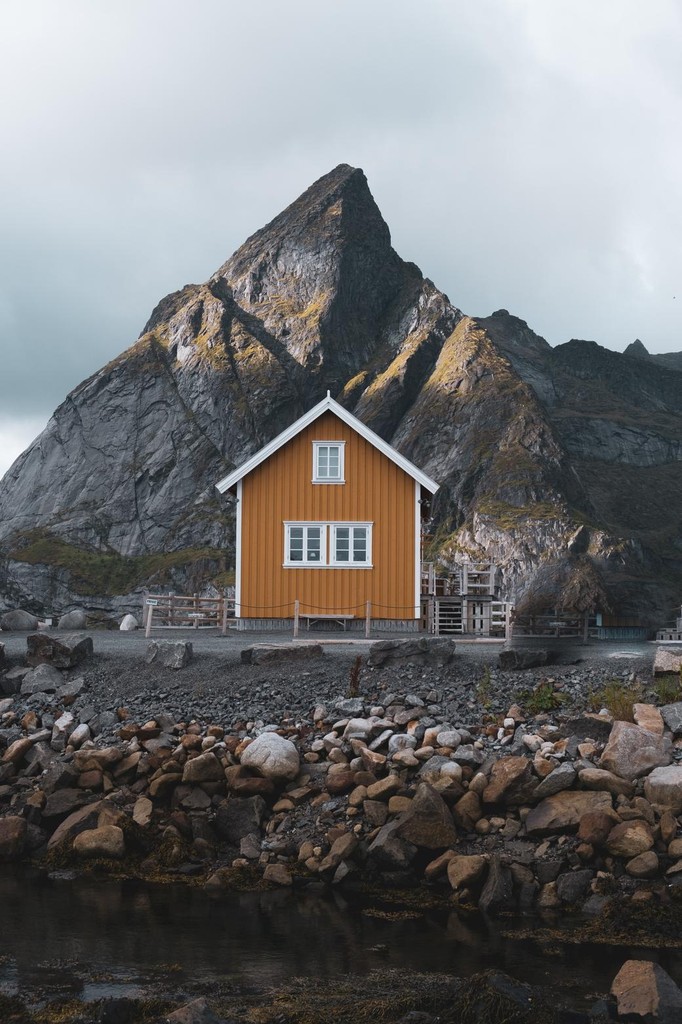}\\

\includegraphics[width=\ww]{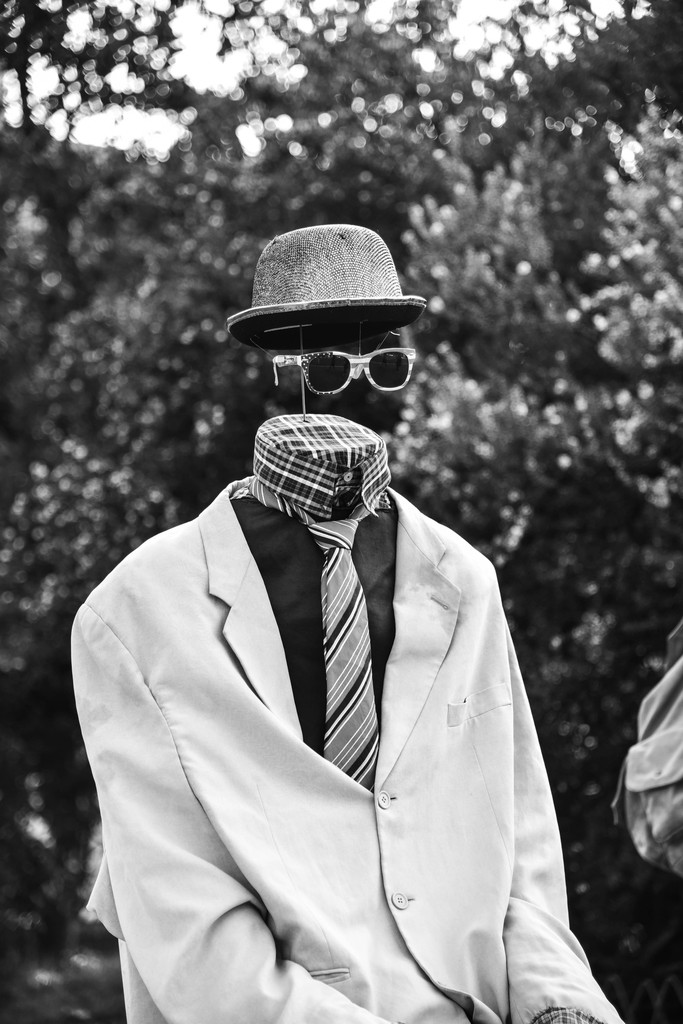} &
\includegraphics[width=\ww]{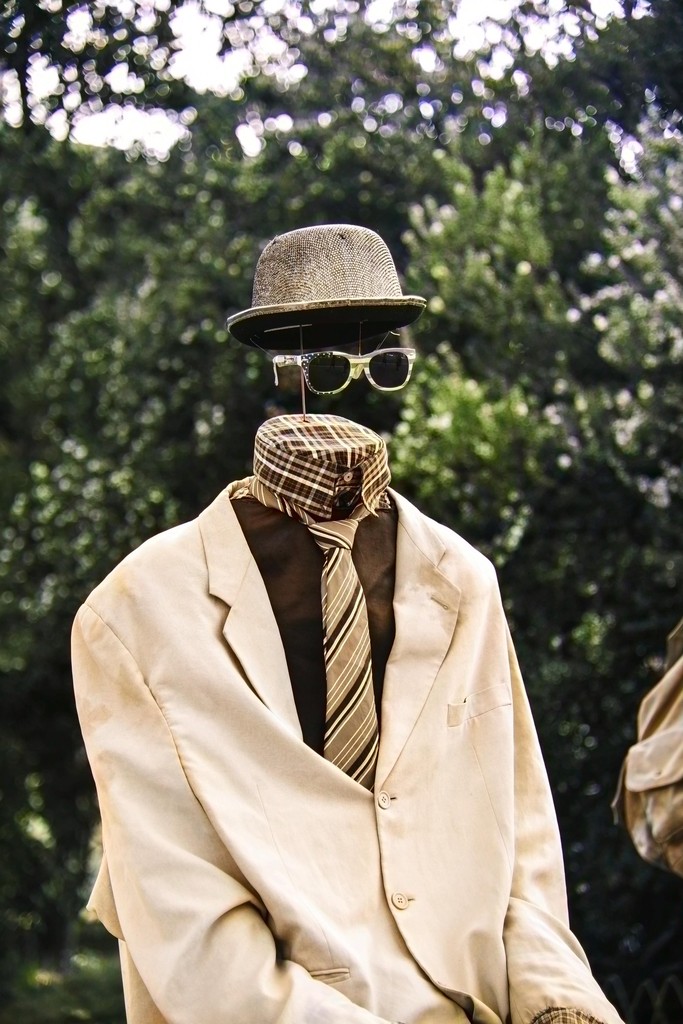} &
\includegraphics[width=\ww]{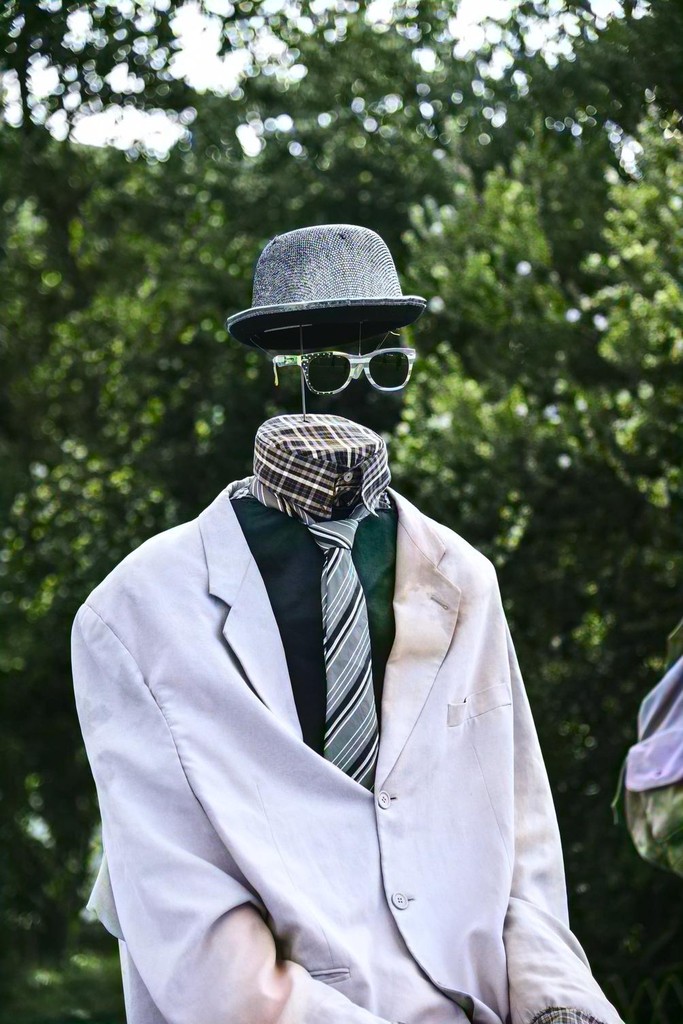} &
\includegraphics[width=\ww]{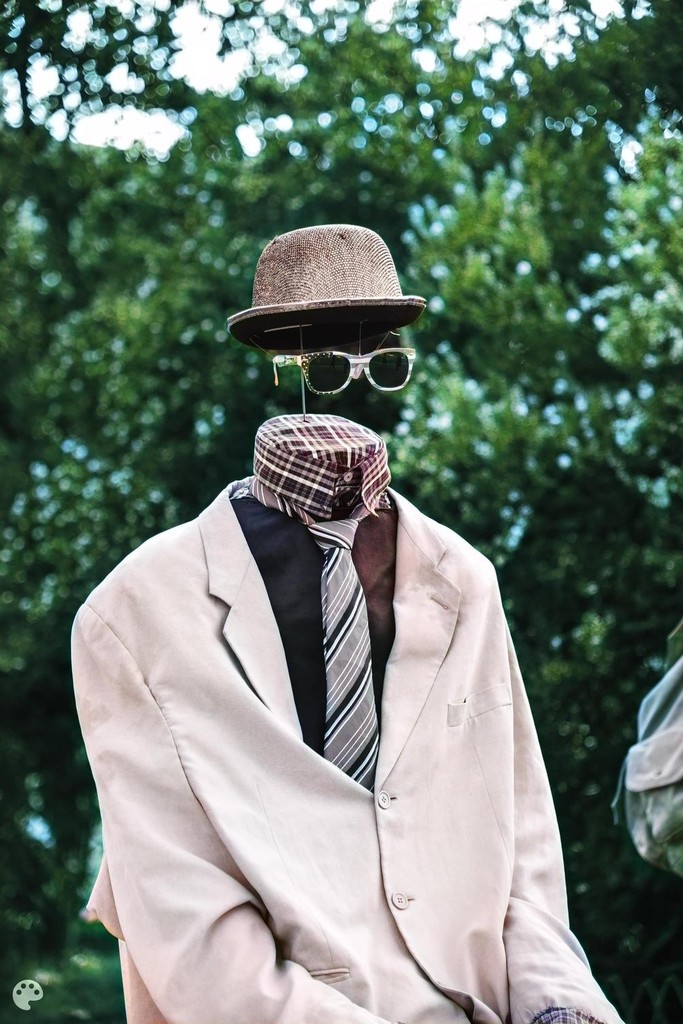} &
\includegraphics[width=\ww]{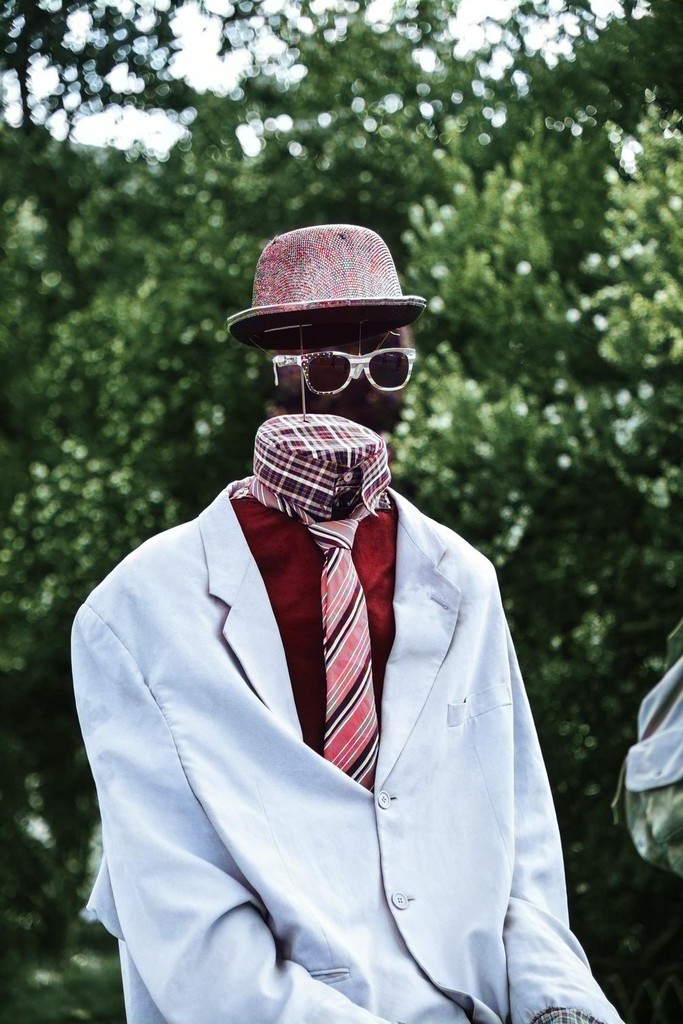} &
\includegraphics[width=\ww]{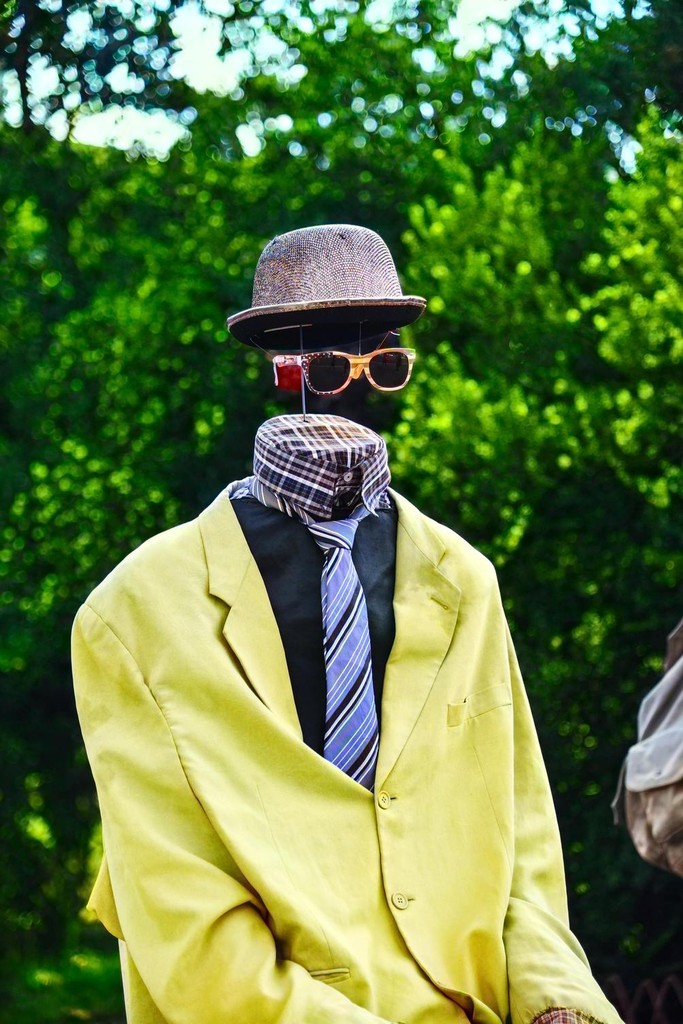} &
\includegraphics[width=\ww]{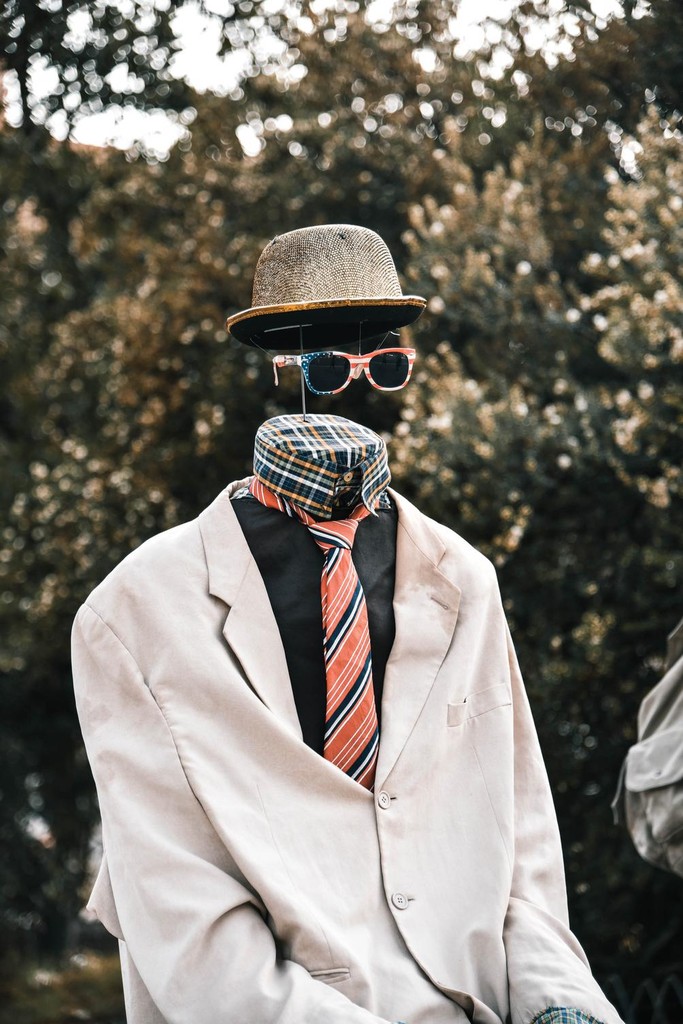}\\

 (a) Input & (b) UniColor &(c)  BigColor & (d) DeOldify &(e) Disco &(f) Ours & (h)  Ground truth \\ 

\end{tabular}

\caption{
\textbf{Visual Comparison on Challenging Examples.} Given an input grayscale image (a), we compare the outputs of previous colorization approaches (b-e) to ours with the default negative prompt (f)
Image credits (top to bottom): Unsplash ©Tim Mossholder, Unsplash ©David Becker, Unsplash ©Joshua Kettle, Unsplash ©Zane Lee.}
\label{fig:qualitative_comparison}
\end{figure*}

For quantitative measurements, we follow common practice and use a mix of reference-based and statistical metrics that measure different aspects of the colorization task. Our main metric for assessing perceptual realism is Fréchet Inception Distance (FID) \cite{FID}, as the primary aim is the generation of vibrant, plausible colors. Additionally, the Absolute $\Delta$-Colorfulness metric is utilized to compare the color vividness between original and re-colorized images. Other common metrics for image colorization include Colorfulness \cite{colorfulness_metric}, SSIM, PSNR and LPIPS which measure colorization \textit{factuality} (i.e., these metrics are not taking into account the multi-modality of the problem). We demonstrate in the SM that a numerical increase in Colorfulness doesn't necessarily reflect an improved colorization, and adjusting image saturation can easily affect this metric. In order to estimate the effect of the text prompts on the colorization process, we measure the CLIP similarity between auto generated prompts (based on the ground truth image) and the re-colorized image. We show that colorizations guided by auto generated text prompts score higher in this metric than using a constant fixed prompt. We show additional qualitative comparisons at \Cref{fig:text_qualitative_comparison,fig:text_effect,fig:qualitative_comparison} and in the SM. Note that unlike \Cref{exp:automatic_colorization} here we compare to auto generated prompts from the \textit{color} images.\\
\subsubsection{Scaling the color vector.} As described in \Cref{sec:color_ranker}, the color vector representation generated by the inference algorithm is determined up to an unknown scale, which we wish to optimize. We trained a regressor to rank the decoded images of the same color vector under different scale factors, aiming for a result that is both natural and preferred by humans, as defined by our manual labelling. We empirically found the best scale factor is mostly in the range $S=[.7,1.4]$ so we feed the regressor with images decoded by the following scaled color vectors: $\left[z_x'+s \cdot \Delta \vert s \in S]\right]$, where $s$ is sampled in steps of $0.1$. Our evaluation of this color-ranking approach, which utilizes multiple color scales and their ranking, demonstrates improved colorfulness metrics compared to those achieved using a constant color scale alone. As presented in \Cref{tab:quantiative_comparison}, our method achieves the lowest Absolute $\Delta$-Colorfulness distance from the ground truth, compared to baseline approaches. This is presumably due to our preference for more natural-looking color intensities during the manual labeling process. Additionally, we report in the SM the metrics with a variety of constant scales and automatic ranking approaches. We further present the tradeoff between FID and colorfulness difference with varying scale values.

\subsection{User-Study}
We conducted a user-study using the Survey Monkey platform\footnote{\url{https://www.surveymonkey.com}}, comparing to five top methods from the quantitative comparison. Our study involved testing on a diverse set of $79$ input images which were sourced randomly from various datasets, including the ImageNet validation set, images from Unsplash\footnote{\url{https://unsplash.com}}, and legacy photos, and included $402$ participants, each of which answered a total of $11$ questions. We evaluated our method on three metrics: (i) total number of votes for each method, (ii) most frequently selected method for each question, and (iii) Elo score (detailed in the SM). For each question, we randomly shuffled and displayed all $6$ colorized images and instructed the participants to select the image that appeared to be the most realistic. More information can be found in the SM. Our method outperformed all other methods in all image subsets, both individually and collectively as shown in \Cref{fig:user_study_main}.

\section{Additional Applications}
\subsection{Color Enhancement}
\begin{figure*}[h]
\setlength{\tabcolsep}{1pt}
\newlength{\wwr}
\setlength{\wwr}{0.20\linewidth}

\begin{tabular}{cccc}
  \centering
\includegraphics[width=\wwr]{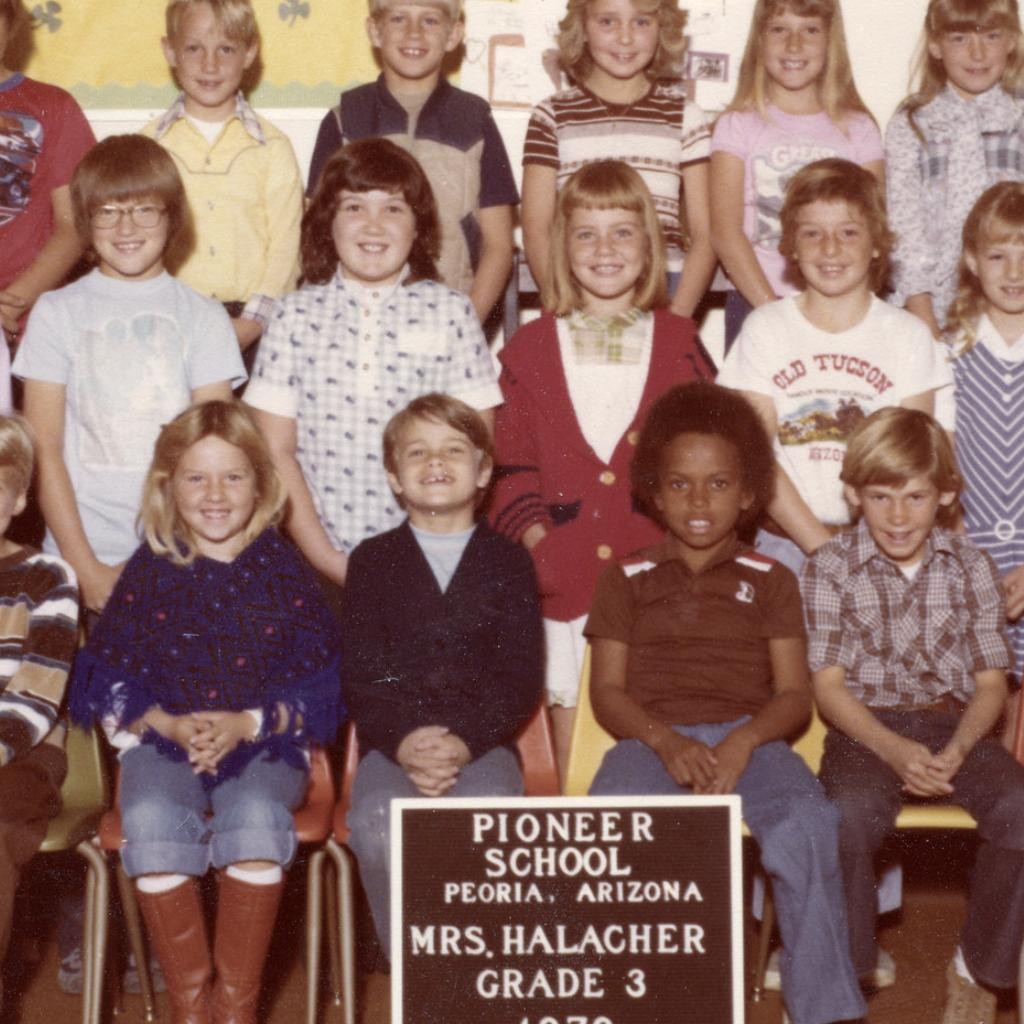} &
\includegraphics[width=0.2219\linewidth]{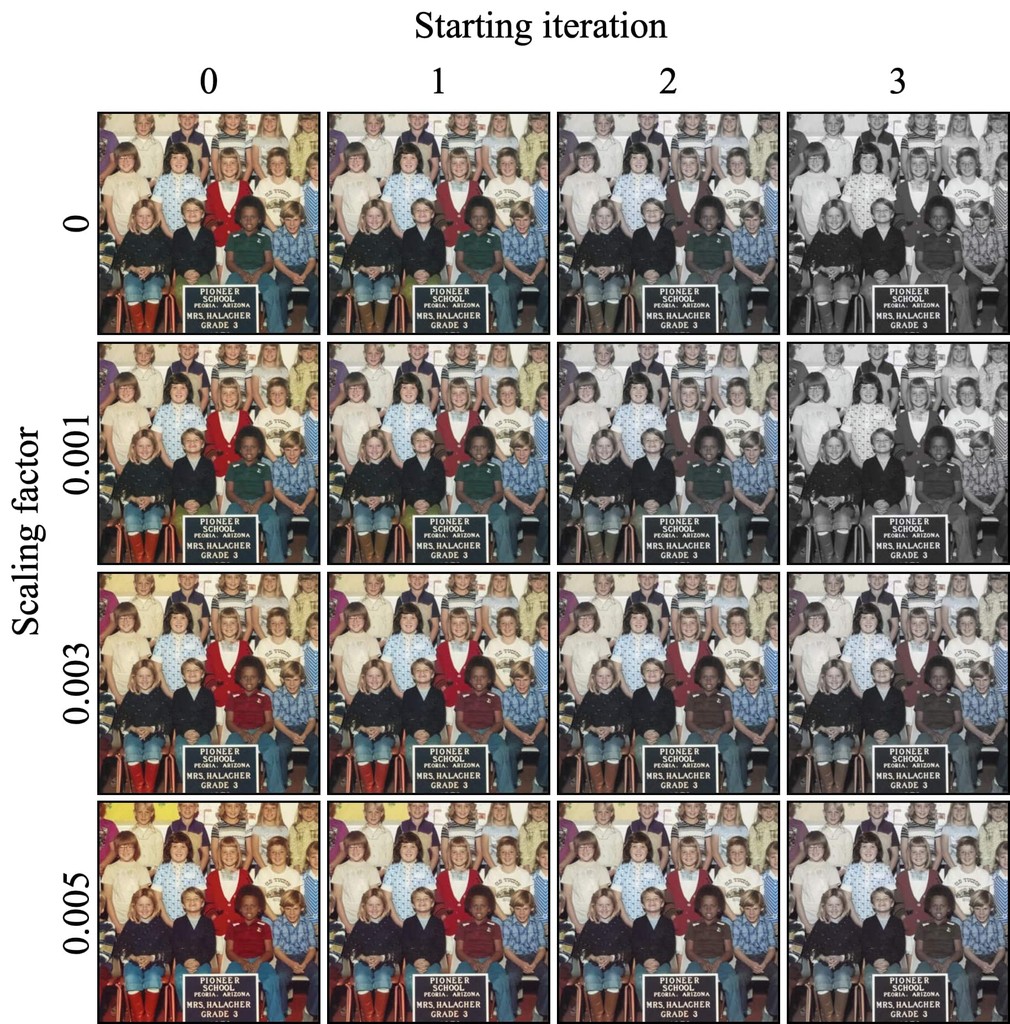} &
\includegraphics[width=\wwr]{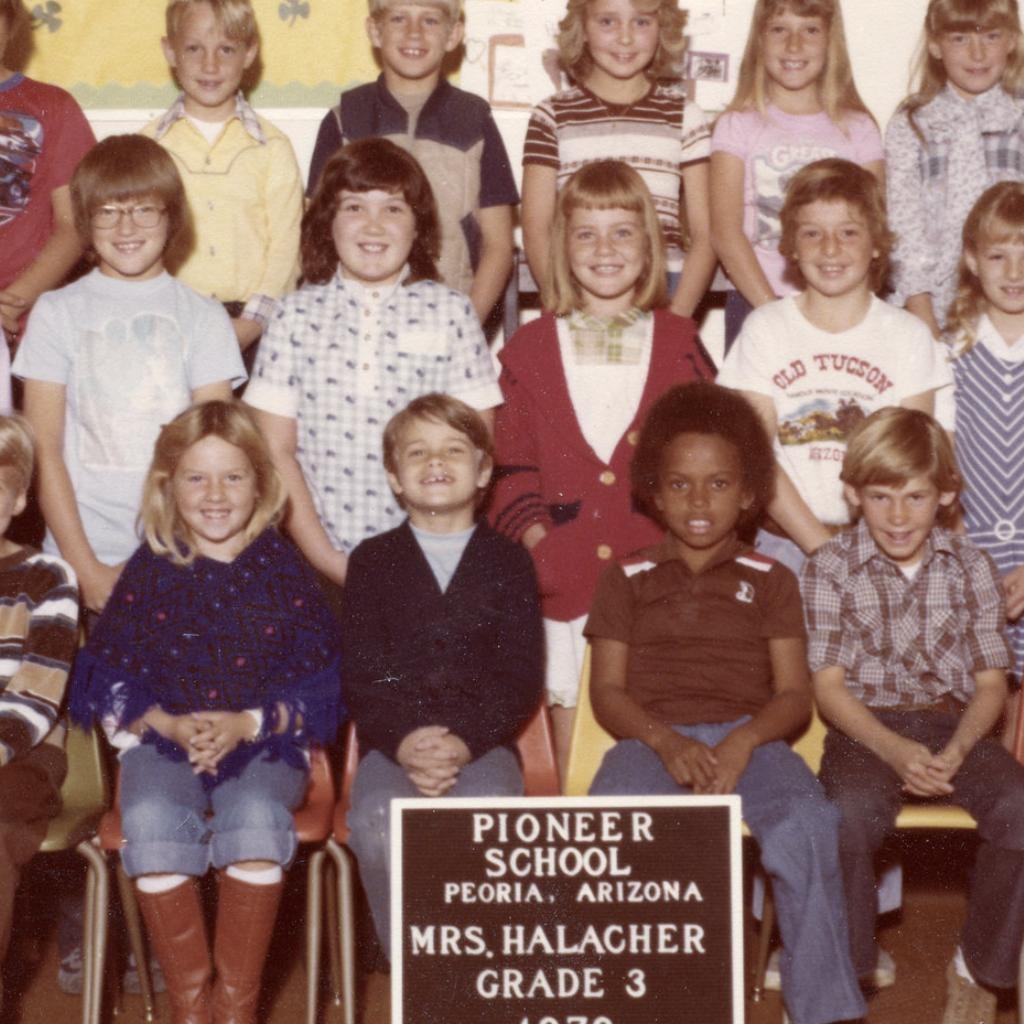} &
\includegraphics[width=\wwr]{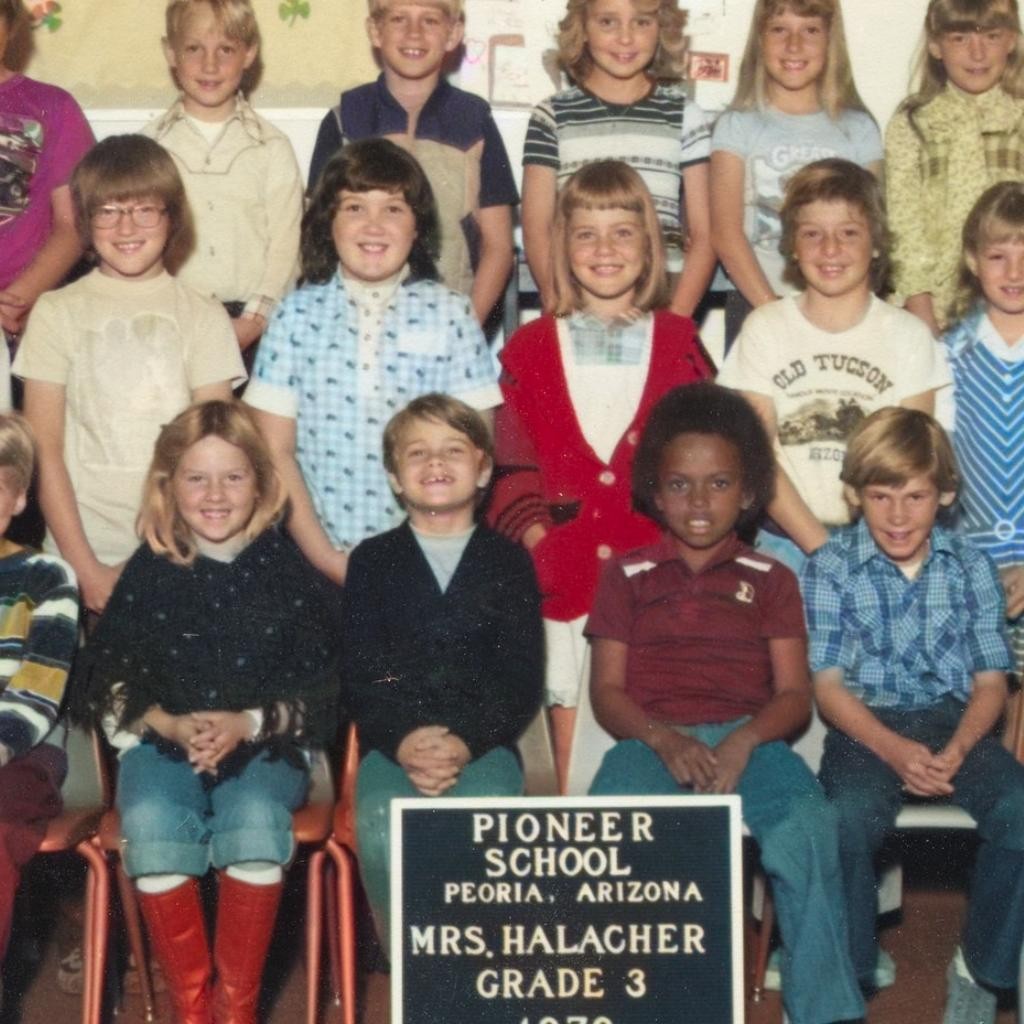}

\\
Input & Options grid & Adjusted Saturation & User Pick

\end{tabular}

\caption{
\textbf{Color Enhancement Workflow.} Starting with an old image with degraded colors (left), we colorize it starting with some of the colors of the original image instead of the grey image by multiplying the AB channels with a scaling factor (grid rows: [0, 0.001, 0.003, 0.005]). We also control the final result by changing the starting iteration of the diffusion process (grid columns: [0,1,2,3]). Input prompts are generated automatically using CLIP interrogator on the input RGB image. Notice that a minimal level of initial colors is required for the diffusion process to \emph{enhance} instead of \emph{invent} colors. A mere saturation change (``Adjusted Saturation'') is visibly inferior to our method. The two right images have the same mean S channel in HSV image representation. Image credits: Flickr ©<def>. } 
\label{fig:color_enhancement_v3}

\end{figure*}

Old photos often face color-related challenges such as fading, discoloration, and a loss of brightness due to factors like weather conditions and natural aging. These factors can disrupt the photo's color balance and cause it to look dull. Hence, effective restoration techniques are essential for maintaining the visual quality of these historical images.
Luckily, the iterative nature of our model makes it a natural candidate for enhancing existing color.

Our approach is to provide a starting point for the colorization process, one which will be enhanced using the diffusion procedure. 
Optionally, we also provide the user with the flexibility of choosing their preferred color intensity and fidelity for any given image by controlling the initial timestep and the level of color for the initial image. We showcase our approach in \Cref{fig:color_enhancement_v3}.
\subsection{Image Restoration}
Our method can be seamlessly integrated into existing restoration pipelines, as an interchangeable step. However, usually best results are achieved when it is performed as the last step. Our hypothesis is that at this stage the images are closer to the distribution the model was trained on. We showcase the effectiveness of our method
in \Cref{fig:interactive_restoration}, demonstrating its ability to produce vibrant and realistic colorizations.

\section{Limitations}
As in other learning models, our method encounters difficulties when processing out-of-distribution samples (such as those in \Cref{fig:interactive_restoration}). Operating in the VAE's latent space enables faster training, but it relies heavily on the VAE's encoding capabilities. If the VAE fails to generate in-distribution latent code, the network struggles to produce compelling colorization. 
To overcome these limitations, we can either utilize a more comprehensive training dataset that covers a wider array of natural images or apply diffusion within the pixel space instead of the latent space. Our model inherits limitations from its underlying diffusion model. For example, controlling the spatial output with text prompts is limited (See Figure 8 in the SM) and is an active area of research \cite{Controlnet_Zhang2023AddingCC}. Therefore, prompts determining counting, spatial layout, or complex scenes can often fail in producing the required results. 

\section{Conclusion}
We developed a text-guided colorization method motivated by the cold diffusion and latent diffusion techniques, capitalizing on their combined strengths of computational efficiency, expressiveness, and controllability to outperform existing colorization methods. 
We explored the diverse outputs achievable through different input captions, as well as through adaptive color scaling using our color ranker. Our text-guided colorization offers a complementary method for image restoration and color enhancement pipelines. By applying our method, we can rejuvenate historical images, enhancing visual quality for a wider audience.
\\

\clearpage
\balance
\bibliographystyle{ACM-Reference-Format}
\bibliography{bibliography}

\clearpage
\appendix
\newpage
\twocolumn[
\begin{@twocolumnfalse}
  \begin{center}
    {\Huge Diffusing Colors: Image Colorization with Text Guided Diffusion (Supplementary Material)}
  \end{center}
  \vspace{10mm} 
\end{@twocolumnfalse}
]

\section{Hyperparameters for Automatic Image Colorization}

\subsection{Scale and Guidance Scale}
We run on imagenet-val1k using grid search the best scale and guidance scale to be used, and evaluated with FID, as seen at \Cref{tab:grid_search_cfg}.
\begin{table*}[ht]
  \caption{We present a comparative analysis of different methods on the validation sets. We emphasize that the most relevant metrics are FID and $\Delta$-Colorfulness ($\Delta$-CLR), which takes into account the multi-modality of colorization, in contrast to PSNR,SSIM and LPIPS. Highest performing are highlighted in bold.  Baseline numbers are from \cite{DISCOXia2022DisentangledIC}. * denotes using the ground truth captions.} 

  \label{tab:quantiative_comparison_factual}

  \begin{tabular}{lcccccccc}
  
    \toprule
    \textbf{Method} & \multicolumn{4}{c}{ImageNet (10k)} & \multicolumn{4}{c}{COCO-Stuff (5k)} \\
    \cmidrule(lr){2-5} \cmidrule(lr){6-9}
    &  CLR ($\uparrow$) & PSNR ($\uparrow$) & SSIM ($\uparrow$ )& LPIPS ($\downarrow$) & CLR ($\uparrow$) & PSNR ($\uparrow$)& SSIM ($\uparrow$) & LPIPS ($\downarrow$)\\
    \midrule
    CIColor & 42.89  & 21.96 & 0.897 & 0.224 & 42.46   & 22.08 & 0.902 & 0.217 \\
    UGColor & 27.91   & \textbf{24.26} &  \textbf{0.919} & \textbf{0.174} &  28.64  & \textbf{24.34} & \textbf{0.924} & \textbf{0.165}  \\
    Deoldify &  23.41  & 23.34 &  0.907 & 0.188 & 23.62 & 23.49 & 0.914 & 0.181 \\
    InstColor  & 25.54  & 22.03 &  0.909 & 0.919 &  29.38   & 22.35 & 0.838 &  0.238 \\
    ChromaGAN & 29.34 & 22.85 &  0.876 &0.230 &  29.34 & 22.74 & 0.871 & 0.233 \\
    ColTran  & 38.64  & 21.81 &  0.892 &  0.218  & 38.95 & 22.11 &  0.898 & 0.210  \\
    Disco & \textbf{51.43}  & 20.72 & 0.862 & 0.229  & \textbf{52.85} & 20.46 & 0.851 &  0.236 \\
    BigColor  & 43.03  & 21.231 & 0.8625 & 0.2265  & 43.156 & 21.208 & 0.8712 & 0.218 \\
    Ours (Scale=0.8) &  33.26 & 23.133 &  0.9053 & 0.184 & 31.437 & 23.556 &  0.907 &  0.186\\
    Ours (Ranker) & 41.40 & 22.042 & 0.889 & 0.201  & 39.34   & 22.572 & 0.895 & 0.200\\
    \midrule
    Ours (Ranker)* &  43.24  & 22.869 & 0.896 & 0.178 & 40.284  & 23.416 & 0.900 & 0.183\\

    \bottomrule
  \end{tabular}
\end{table*}

\begin{table}[ht]
\setlength{\tabcolsep}{5pt}
\caption{Quantitative comparison on COCO-Stuff, using the ground truth captions. UniColor values taken from \cite{huang2022unicolor}.}

\begin{center}
\begin{tabular}{ccccc}
\toprule
\textbf{Method} & \textbf{CLIP Score} ($\uparrow$) & \textbf{FID} ($\downarrow$) \\
\midrule
UniColor & $24.50$ & $11.29$ \\
Ours & \textbf{$30.44$} & \textbf{$7.16$} \\
\bottomrule
\end{tabular}
\end{center}
\label{table:text_comparison_metrics}
\end{table}

\subsection{Captions generation for Image colorization}
We provide a detailed explanation of our prompt creation process for each of the methods described in the experiments section:
\begin{enumerate}
    \item \textit{Colorization Clip direction} - To implement this method, we utilized images from the imagenet cval1k dataset. First, we generated prompts for both the colorized and RGB images. Then, we computed the embeddings of these prompts and calculated the residual text latent vector for each image. To obtain a global delta tensor, we averaged the residual vectors across all images. During inference, we added this global residual to the textual embedding of the prompt.
    \item \textit{Remove grayscale hints} - For each image, we created the captions for both the colorized and grayscale versions using CLIP-interrogator. We automatically identified the phrases present in the grayscale caption but absent in the RGB caption. We then compiled a list of the top 100 most common phrases and automatically removed them, along with the phrases ``black and white'' and ``grayscale''. We applied this transformation to all of the grayscale captions. 
    \item \textit{LLM Rephrasing} - To implement this technique, we leveraged the OpenAI Completion API with the davinchi-2 model. We utilized the following prompt: ``You are a highly intelligent agent. Given an image caption of a grayscale image, rephrase it as a colorized RGB photo. Remove any keywords relevant to grayscale images (e.g., black and white). Maintain a similar style to the input caption. Input Caption: \{CAPTION\}. Output Caption:''. We set the parameters as follows: temperature=0.1, max\_tokens=500, frequency\_penalty=0, presence\_penalty=0.
    \item Optimal Negative Prompt - We selected common phrases from the grayscale captions (as mentioned in ``remove grayscale hints'') and manually evaluated a few of them with negative prompts. 
\end{enumerate}

\subsection{Number of steps and Running Time}
We ablate the number of steps (which is correlated to the running time) with the colorization results, as measured with FID. \Cref{tab:ablate_stride} depicting this tradeoff.

\begin{table}[ht]
\setlength{\tabcolsep}{4pt}
\begin{center}
\caption{This table presents an analysis of the effect of the number of steps on FID for ImageNet cval-1k, emphasizing the tradeoff between the running time and colorization quality. As illustrated in the main text, a larger number of steps generally leads to increased details and colorfulness in the output image. However, this enhancement comes at the cost of longer inference times. Interestingly, the best results were achieved with a number of steps of 50. We assume this is due to the original image's level of colorfulness, that can have different amounts of color details. Therefore, using 100 diffusion steps makes the image more colorful, while running less than 50 steps makes it less colorful than the original image. The timestamps were taken with a constant stride, correlated to the number of steps.}
\label{tab:ablate_stride}
\begin{tabular}{lccccccc}
\toprule

\# Steps & 1 & 2 & 5 & 10 & 20 & 50 & 100 \\
\midrule
FID & 24.95 & 24.89 & 24.80 & 24.52 & 24.20 & \textbf{24.01} & 24.28 \\
Runtime [s] & \textbf{0.05} & 0.08 & 0.19 & 0.37 & 0.73 & 1.80 & 3.64 \\
\bottomrule
\end{tabular}
\end{center}
\end{table}
\begin{table}[ht]
\setlength{\tabcolsep}{5pt}
\caption{Running time of our method, averaged over 5 runs on A-100 GPU.}

\begin{center}
\begin{tabular}{cc}
\toprule
\textbf{Operation} & \textbf{runtime [s] } \\
\midrule
VAE encode & $0.005$ \\
Inference (stride=1) & $3.64$ \\
Inference (stride=2) & $1.80$ \\
Inference (stride=5) & $0.73$ \\
Inference (stride=10) & $0.37$ \\
Inference (stride=20) & $0.19$ \\
Inference (stride=50) & $0.08$ \\
Inference (stride=100) & $0.05$ \\
VAE Decode & $0.04$ \\
Color Ranking & $5.51$ \\
\bottomrule
\end{tabular}
\end{center}
\label{table:runtime}
\end{table}
\begin{table}[ht]
\caption{\emph{Automatic Colorization Approaches.} We present experimental results on ImageNet cval-1K for various approaches to facilitate automatic colorization without user interaction. We employ the FID metric as the primary assessment tool for evaluating colorization performance.}
\setlength{\tabcolsep}{1pt}

\begin{center}

\begin{tabular}{lcccc}
\toprule

\textbf{Prompt} & \textbf{Default} & \makecell{\textbf{``Colorize''} \\ \textbf{Direction}} & \makecell{\textbf{Negative} \\ \textbf{ Prompt}} \\
\midrule
Input prompt & $38.83$ & $29.23$ & $38.12$ \\
Remove Grayscale Clues & $28.52$ & $27.19$ & $27.18$ \\
LLM Rephrasing & $28.08$ & $27.30$ & $26.93$  \\
Null Prompt & $25.58$  & $25.26$ & $\mathbf{24.52}$ \\
\bottomrule
\end{tabular}
\end{center}
\label{table:ablation_text_source}
\end{table}
\begin{figure}[b]

\begin{tabular}{c}
\includegraphics[width=0.5\textwidth]{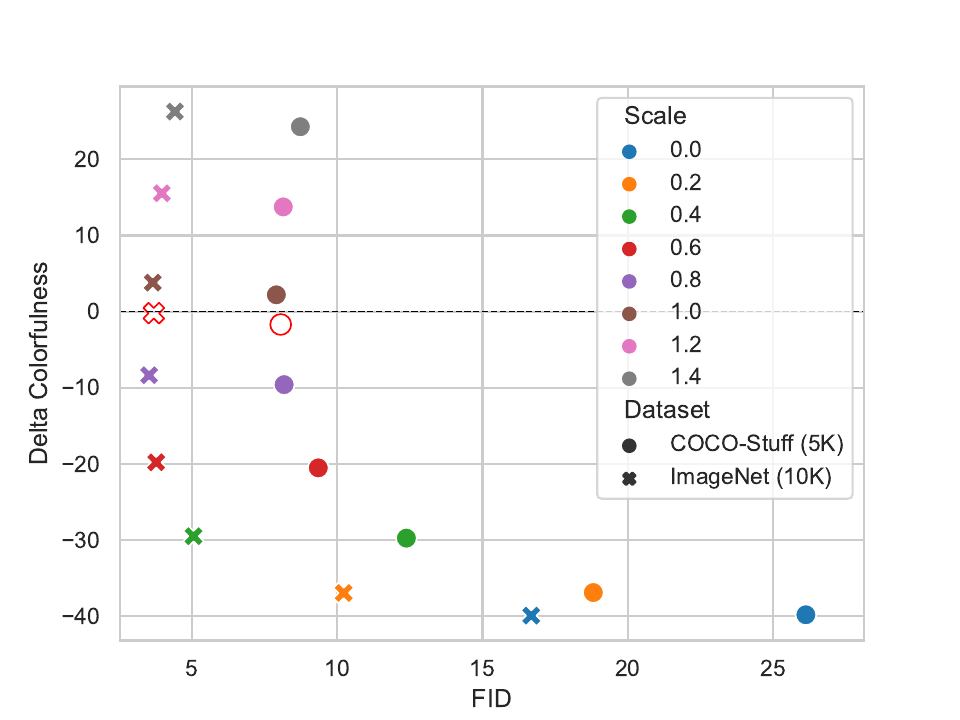} \\
\end{tabular}

\caption{
\textbf{Tradeoff between Colorfulness and FID.} The graph illustrates the relationship between delta colorfulness, which represents the cumulative colorfulness discrepancy between predicted colorizations and actual RGB images, and FID. Filled symbols depict metrics with a consistent color scale, colored according to that scale. By employing a color regressor, as denoted by hollow poins, we achieve an improved balance between perceptual realism and desired colorfulness. 
}
\label{fig:color_scale_vs_ranker_delta_colorfulness}
\end{figure}
\begin{figure*}
\setlength{\tabcolsep}{1pt}
\newlength{\www}
\setlength{\www}{0.10\linewidth}

\begin{tabular}{ccccccccc}
  \centering
\includegraphics[width=\www]{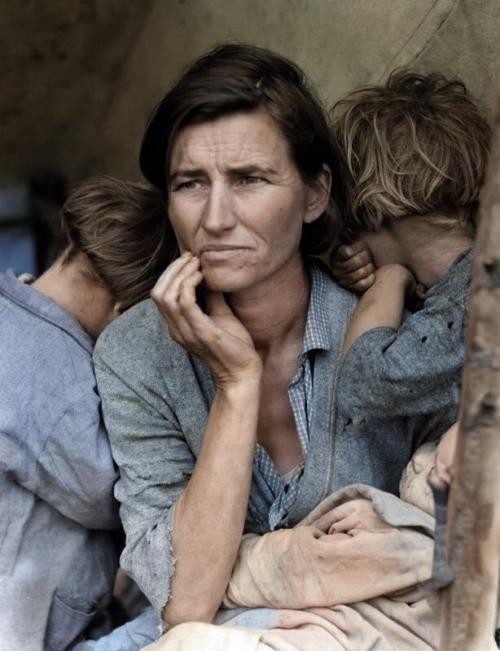} &
\includegraphics[width=\www]{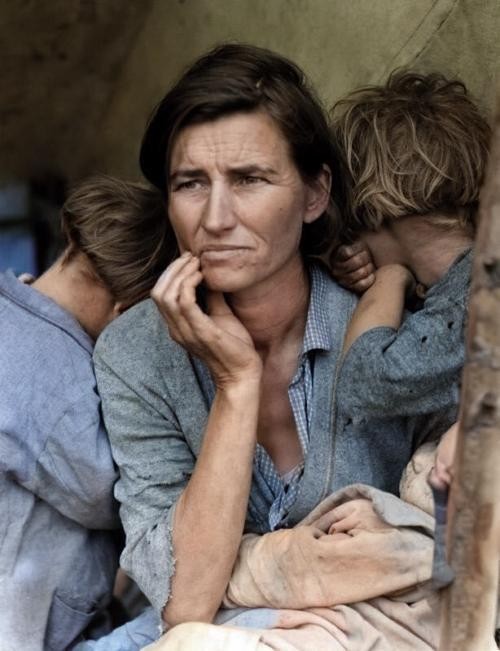} &
\includegraphics[width=\www]{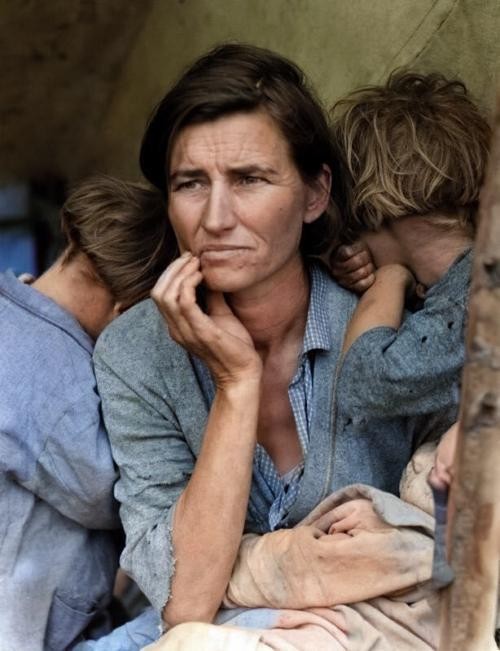} &
\includegraphics[width=\www]{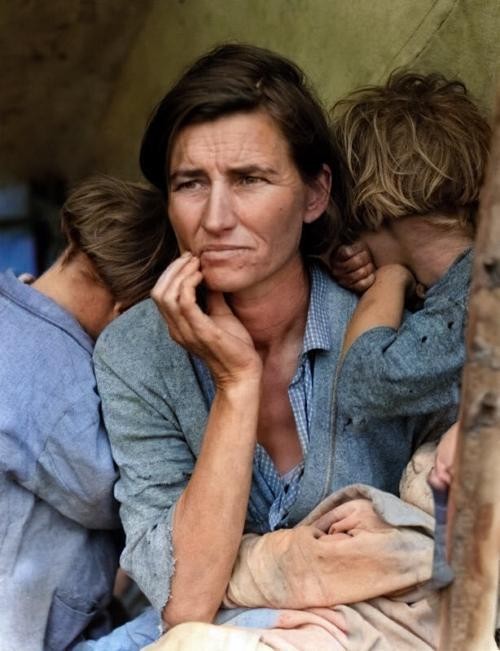} &
\includegraphics[width=\www]{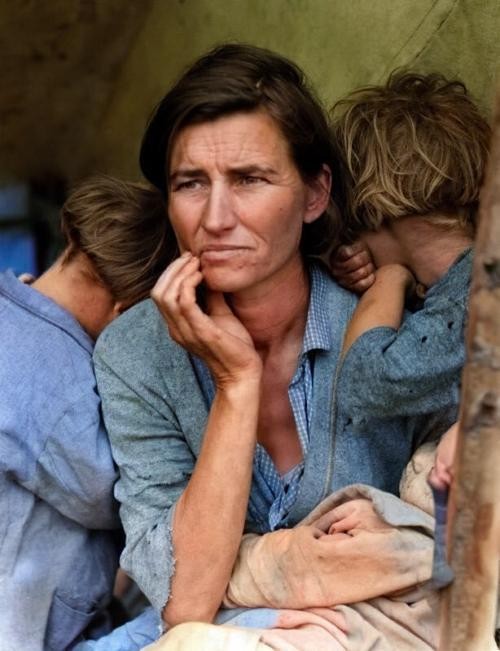} &
\includegraphics[width=\www,cfbox=blue]{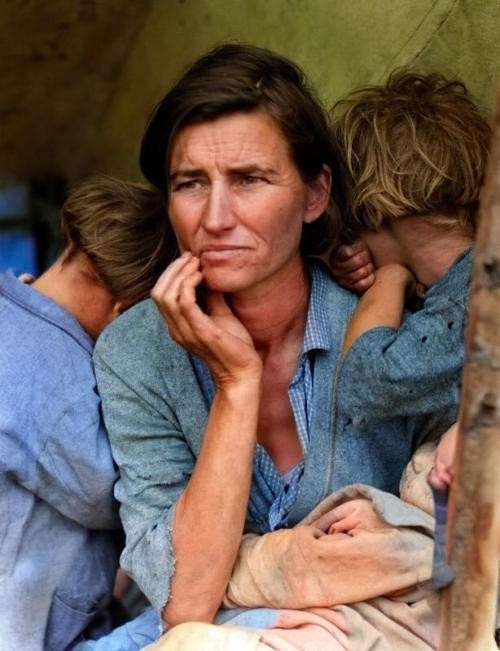} 
&
\includegraphics[width=\www]{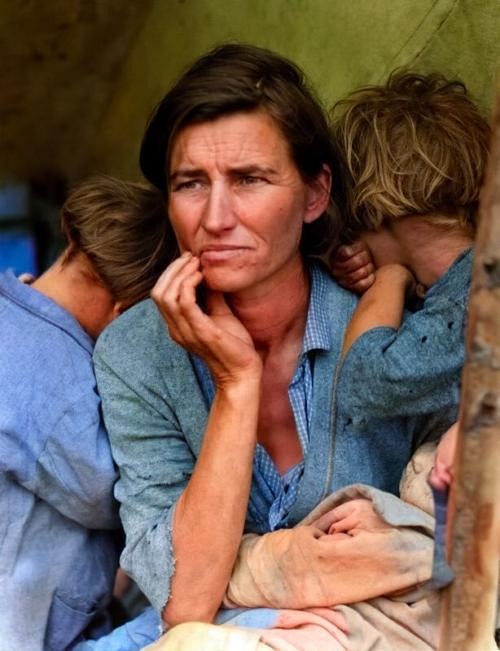} &
\includegraphics[width=\www]{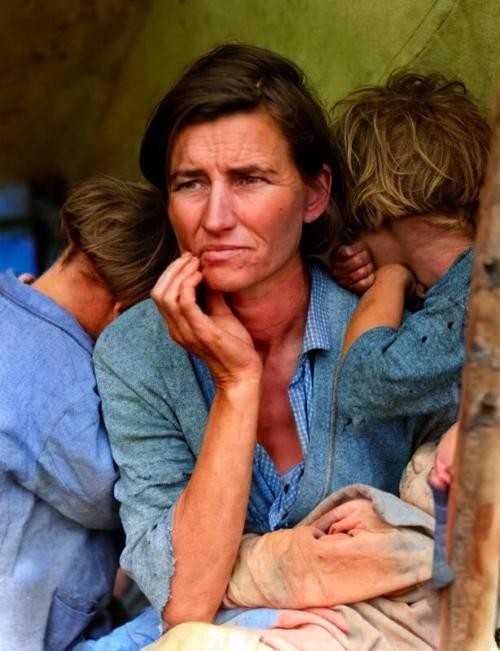} &
\includegraphics[width=\www]{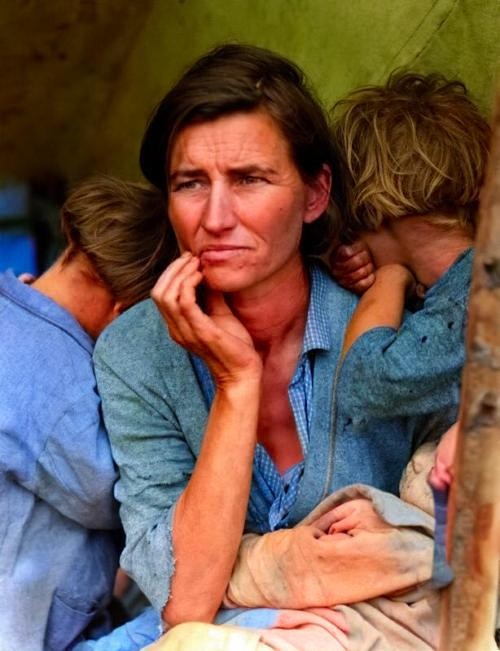}  \\

\includegraphics[width=\www]{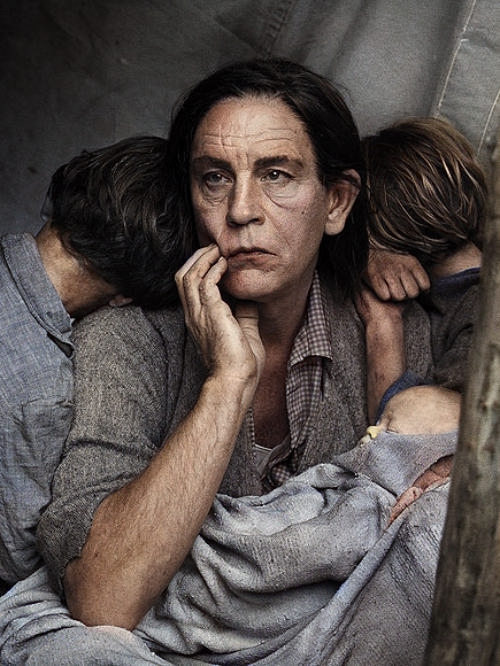} &
\includegraphics[width=\www]{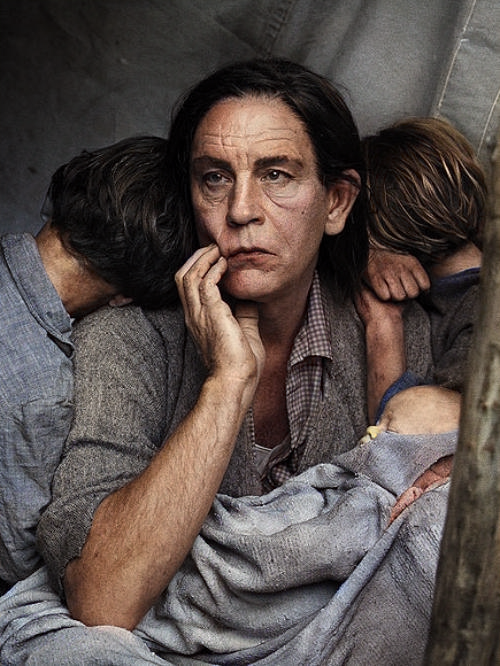} &
\includegraphics[width=\www]{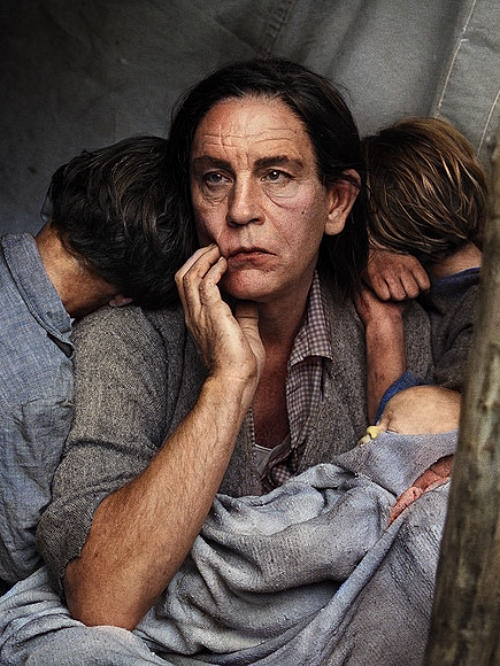} &
\includegraphics[width=\www]{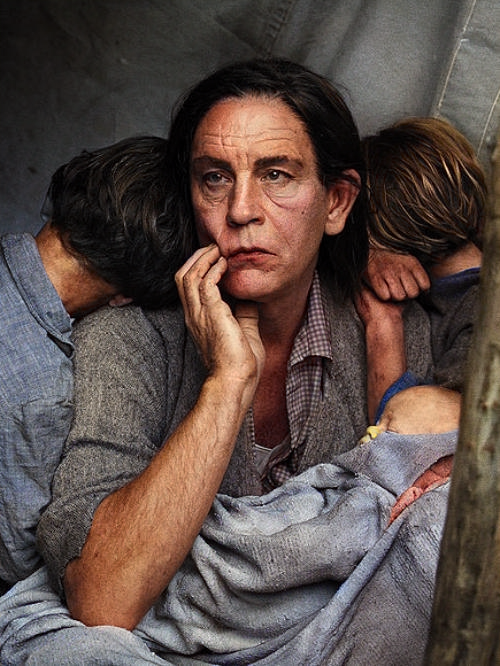} &
\includegraphics[width=\www,cfbox=blue]{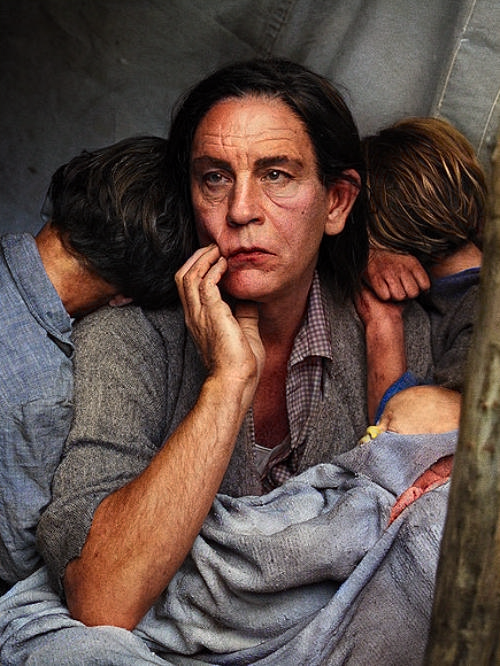} &
\includegraphics[width=\www]{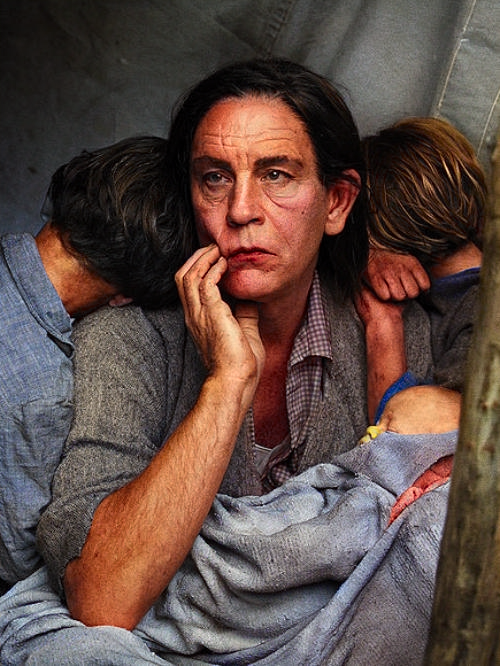} 
&
\includegraphics[width=\www]{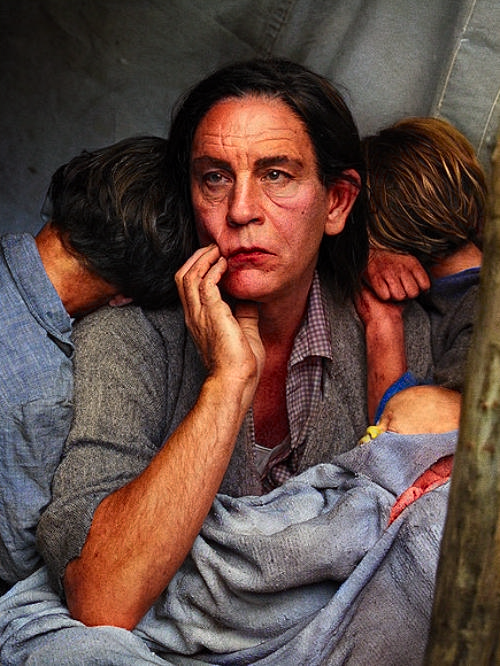} &
\includegraphics[width=\www]{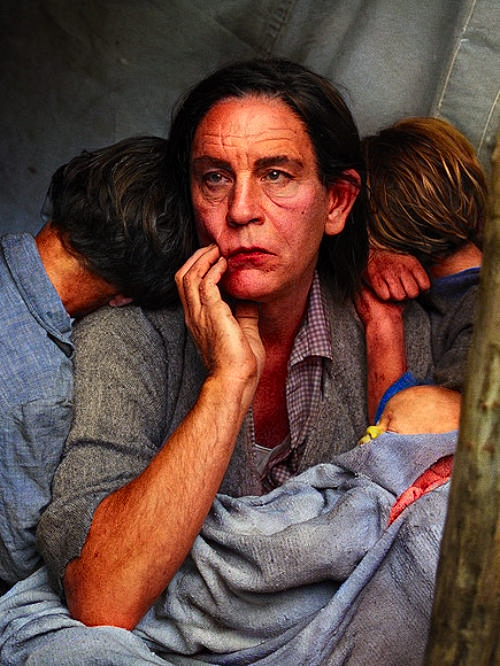} &
\includegraphics[width=\www]{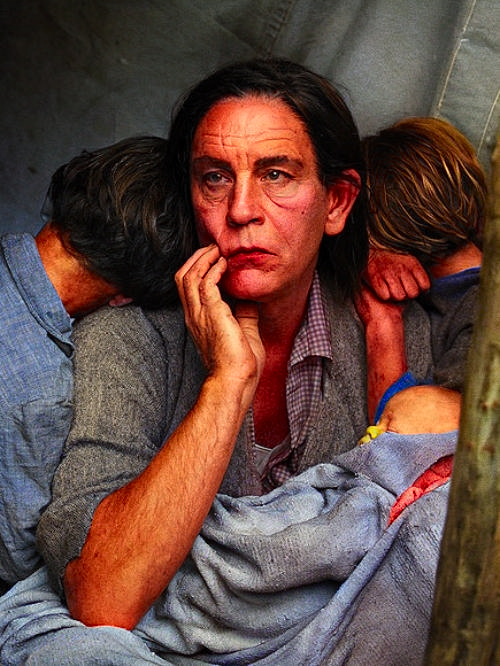}  \\

0.6 & 0.7 & 0.8 & 0.9 & 1.0 & 1.1 & 1.2 & 1.3 & 1.4 

\end{tabular}

\caption{
\textbf{Color Ranker.}
We visualized the effect of the latent color scaling of  scales in range 0.6 to 1.4. Enclosed by a blue frame is the most preferred scale, as predicted by the color ranker. Image credits: `Migrant Mother' by Dorothea Lange.} 
\label{fig:color_ranker_viz}

\end{figure*}

\section{Implementation Details}
We describe in detail the training and inference algorithms, in \Cref{alg:training,alg:inference}.
Similarly to other image colorization works, we trained our model on ImageNet and evaluated on both ImageNet and Coco-Stuff using the same model. During training, the images are resized to $(512, 512)$ and normalized to have mean and standard deviation of $0.5$ along each channel. We used color and blur augmentations randomly for 10\% of the images. We trained with learning rate 1e-5 with the AdamW optimizer, setting $(\beta_1, \beta_2)= (0.9, 0.999)$ and weight decay to $0.01$ for $420K$ steps, where the time steps are sampled uniformly from the range $\left[0,100\right]$. We set the batch size to $16$. The architecture (including the pre-trained VAE, U-Net and text encoder) is the same as that of Stable Diffusion v1.5, with weights initialized from \footnote{ \url{https://huggingface.co/runwayml/stable-diffusion-v1-5}} \cite{StableDiffusionRombach2021HighResolutionIS}, where we optimized only the U-Net. Captions were randomly removed from $10\%$ of the samples. We apply augmentation to the input gray scale image, by randomly changing the contrast and brightness for $10\%$ of the images in a factor sampled uniformly in the range $\left(0.8, 1.2\right)$ and $\left(0.9, 1.1\right)$, respectively. Similar to BigColor~\cite{BIG_COLOR_Kim2022BigColorCU}, we added color augmentations by scaling the chromaticity of the real output image, to yield more colorful images, scaling the chroma channels by a factor sampled in $\left(1.0, 1.2\right)$ for $10\%$ of the samples. We trained with mixed precision of FP16. We discarded images with mean saturation value lower than $0.1$ to remove black and white images from our training set. During evaluation, all images are resized to $256\times256$. We used the following scales for the color ranker $[0.7, 0.8, 0.9, 1.0, 1.1, 1.2, 1.3, 1.4]$. All experiments were done on A100 GPUs with PyTorch 1.13. 
\begin{algorithm}[tbp]

\KwIn{Pre-Trained $\text{VAE}=(E(x), D(z))$, 

UNet, dataset of image-caption pairs $\left(x_i, c_i\right)$, number of steps $n$}
\KwOut{A trained UNet for latent color regression}
\For{$i \gets 1$ \KwTo $n$}{
Sample $t \sim \mathbb{U}\left[0,1\right]$, image-caption pair $\left(x, c\right)$ \\
$z_{x}=E(x),\thinspace z_{x'}=E(\textit{Grayscale}(x))$ \\ %
$z_{t}=(1-t)\cdot z_{x'} + t \cdot z_x$\\%\Delta$ \\
$Loss = L_{2}(z_{x}, \textit{UNet} \left(z_x^t, t, c\right) + z_x^t)$
}

\caption{Training a latent cold-diffusion model}
\label{alg:training}
\end{algorithm}
\begin{algorithm}[tbp]
\KwIn{$VAE=(E(x), D(z))$, UNet, a gray-scale image $x'$, number of steps $T$, positive and negative prompts $c, \tilde{c}$, guidance scale $s$, color scale $s'$}
\KwOut{Colorized image $x$}
$z_x^t = z_{x'}=E(x')$ \\
\For{$t=T, T-1, ..., 1$}{
    \texttt{$Pred_{c}, Pred_{\tilde{c}}=UNet(z_x^t, t, c), UNet(z_x^t, t, \tilde{c})$} \\
    $\hat{z_x}=z_x^t + Pred_{\tilde{c}} + s \cdot (Pred_{c} - Pred_{\tilde{c}})$ \\
    $z_x^t = z_x^t + \frac{1}{T}\cdot (\hat{z_x} - z_{x'})$ \\
}
$z_{x}={z_{x'}}+s' \cdot \hat{z_x}-z_{x'}$ \\ 
\Return $D(z_{x})$
\caption{Inference}
\label{alg:inference}
\end{algorithm}

\begin{figure*}
\setlength{\tabcolsep}{1pt}

\newlength{\qwe}
\setlength{\qwe}{0.220\linewidth}

\newlength{\qwh}
\setlength{\qwh}{0.1\textheight}

\begin{tabular}{cccc}
  \centering

\includegraphics[width=\qwe,height=\qwh,frame]{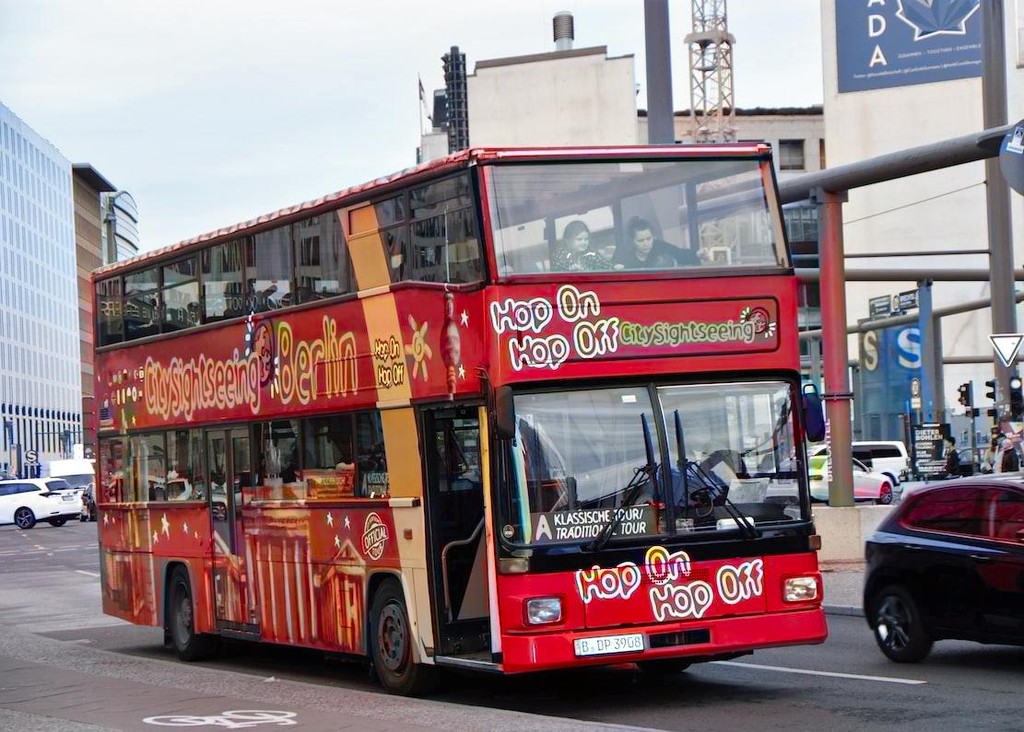}&
\includegraphics[width=\qwe,height=\qwh,frame]{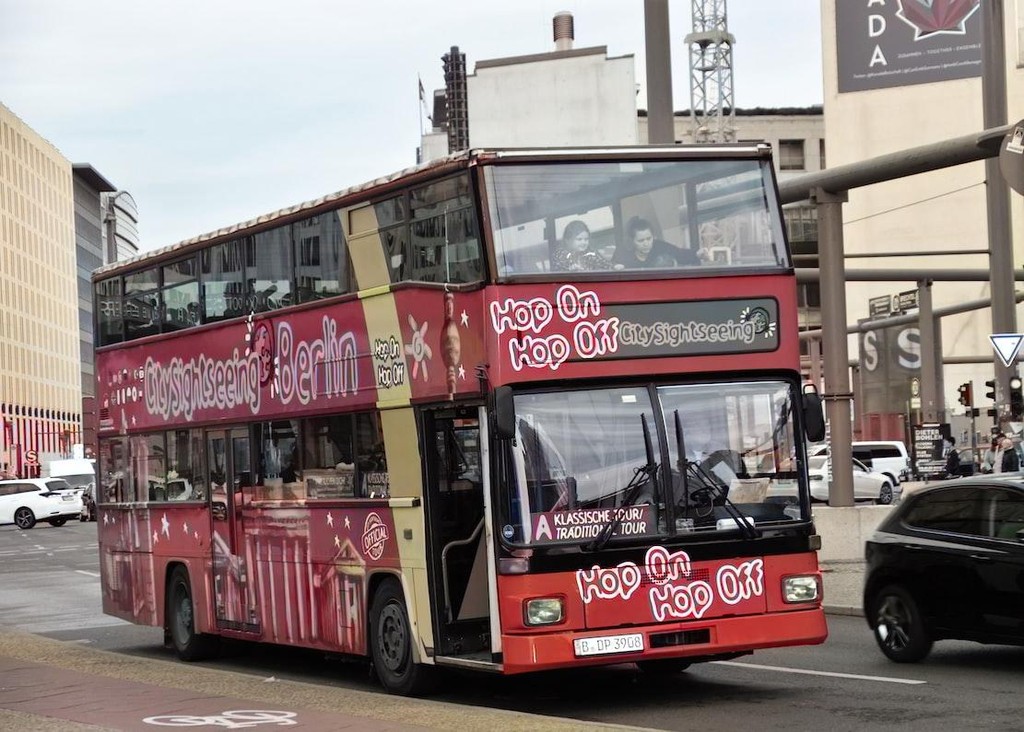} &
\includegraphics[width=\qwe,height=\qwh,frame]{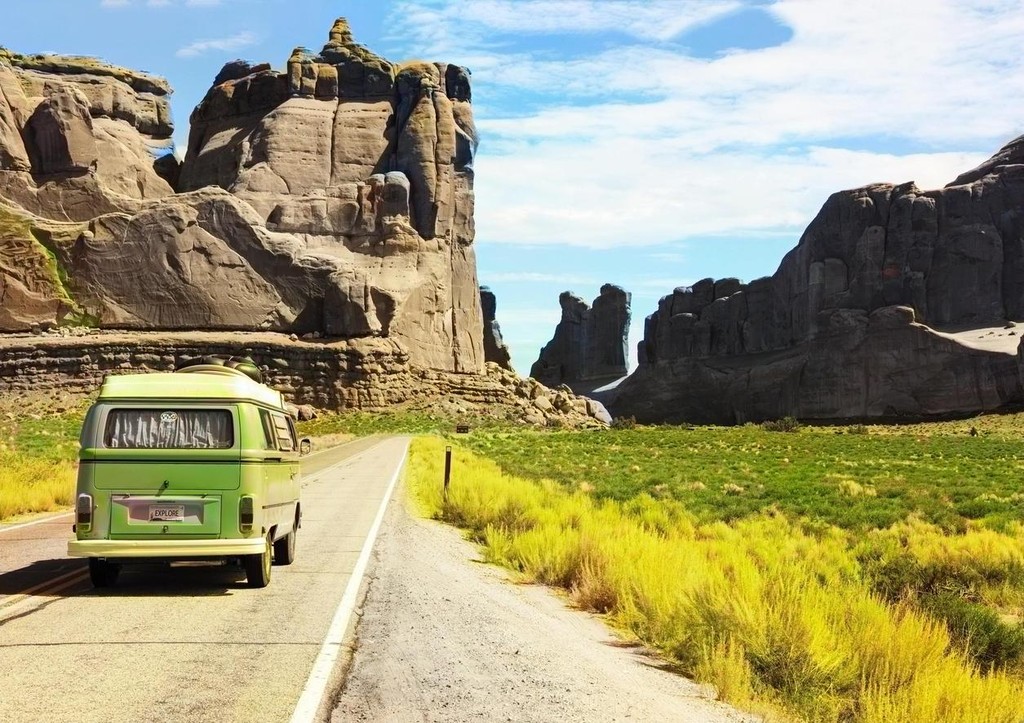} &
\includegraphics[width=\qwe,height=\qwh,frame]{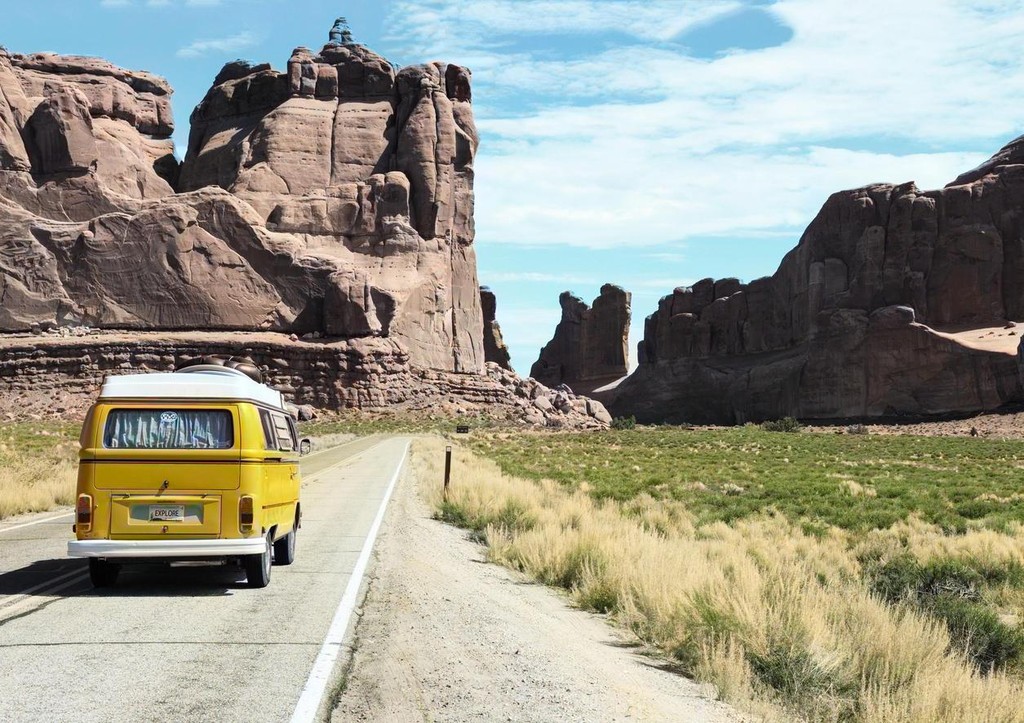} 
\\
\multicolumn{2}{c}{\textcolor{red}{Red} bus} & \multicolumn{2}{c}{\textcolor{yellow}{Yellow} Volkswagen van on road} \\

\includegraphics[width=\qwe,height=\qwh,frame]{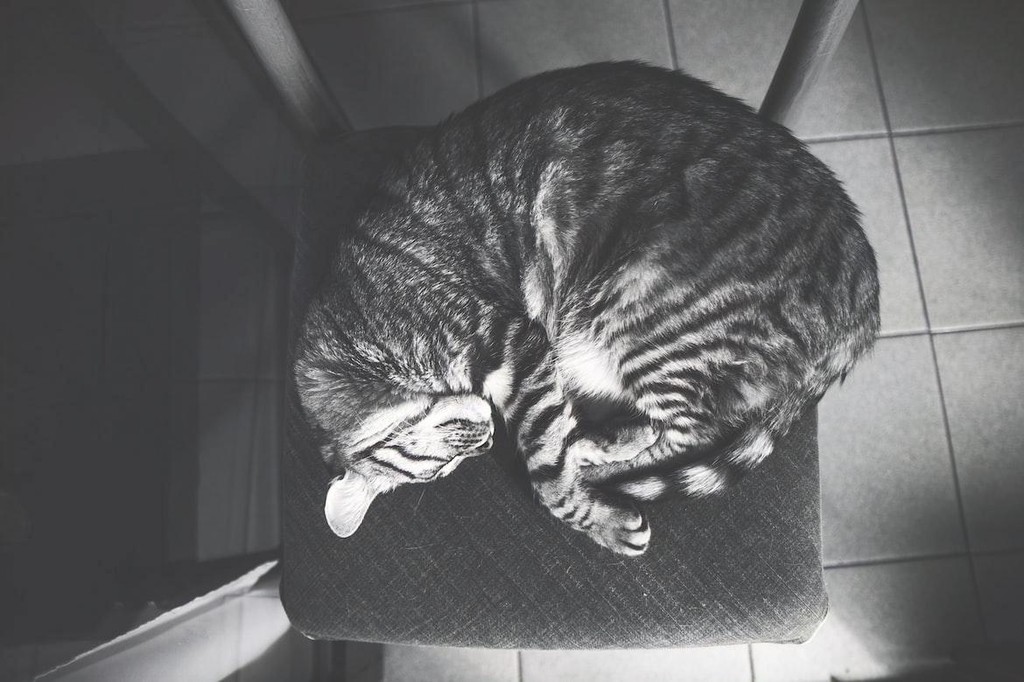}&
\includegraphics[width=\qwe,height=\qwh,frame]{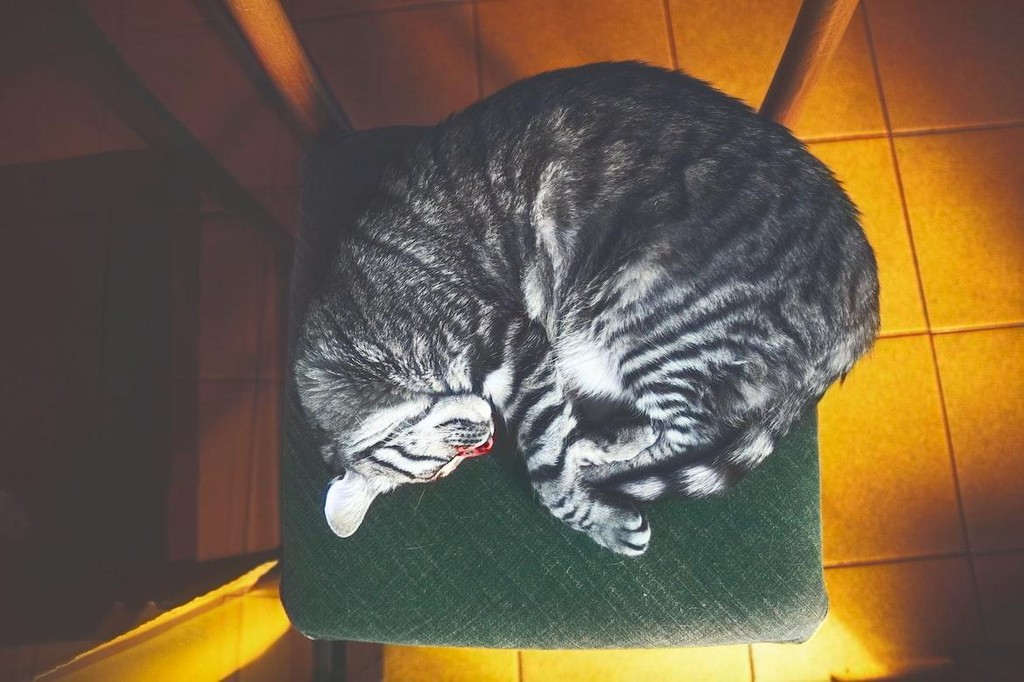} &
\includegraphics[width=\qwe,height=\qwh,frame]{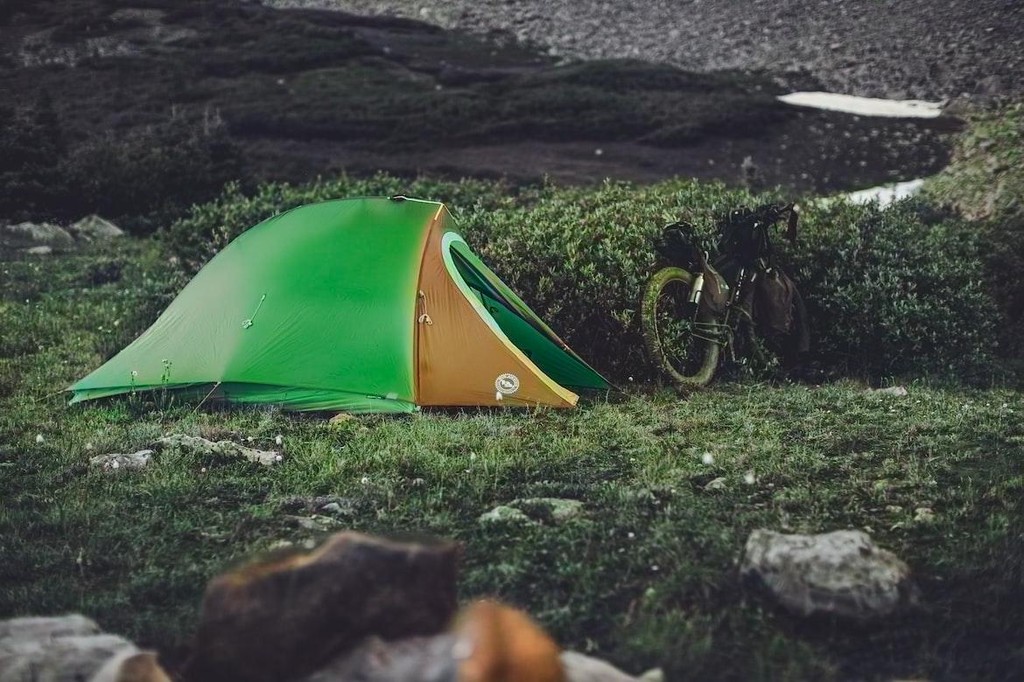} &
\includegraphics[width=\qwe,height=\qwh,frame]{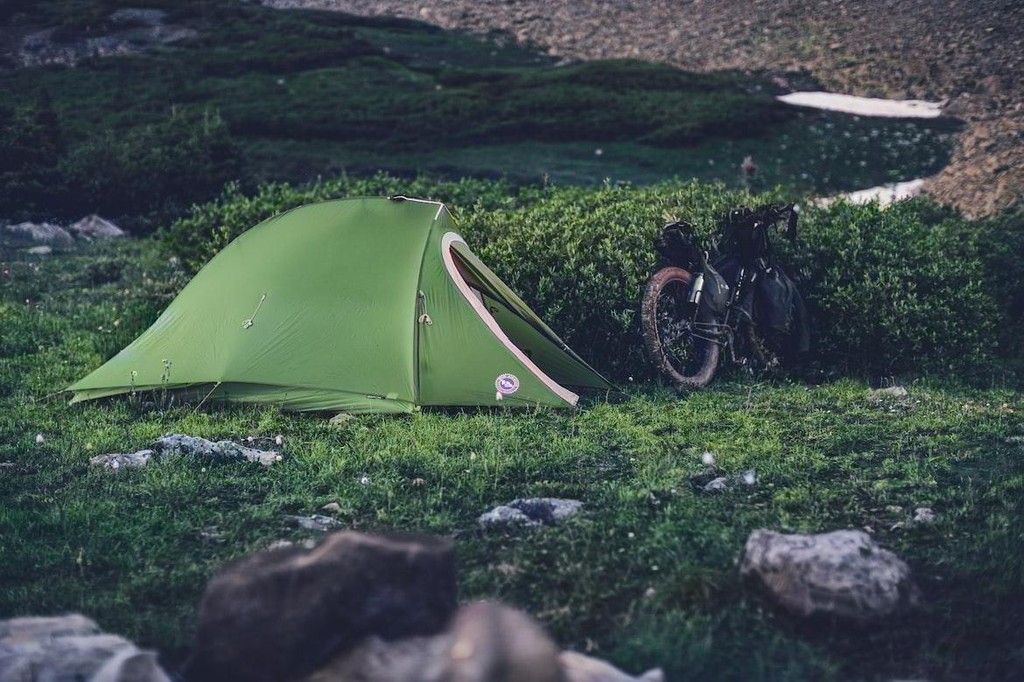} 
\\
\multicolumn{2}{c}{\textcolor{gray}{Gray} cat has curled up into a ball} & \multicolumn{2}{c}{A parked motorcycle next to a \textcolor{green}{green} tent} \\

\includegraphics[width=\qwe,height=\qwh,frame]{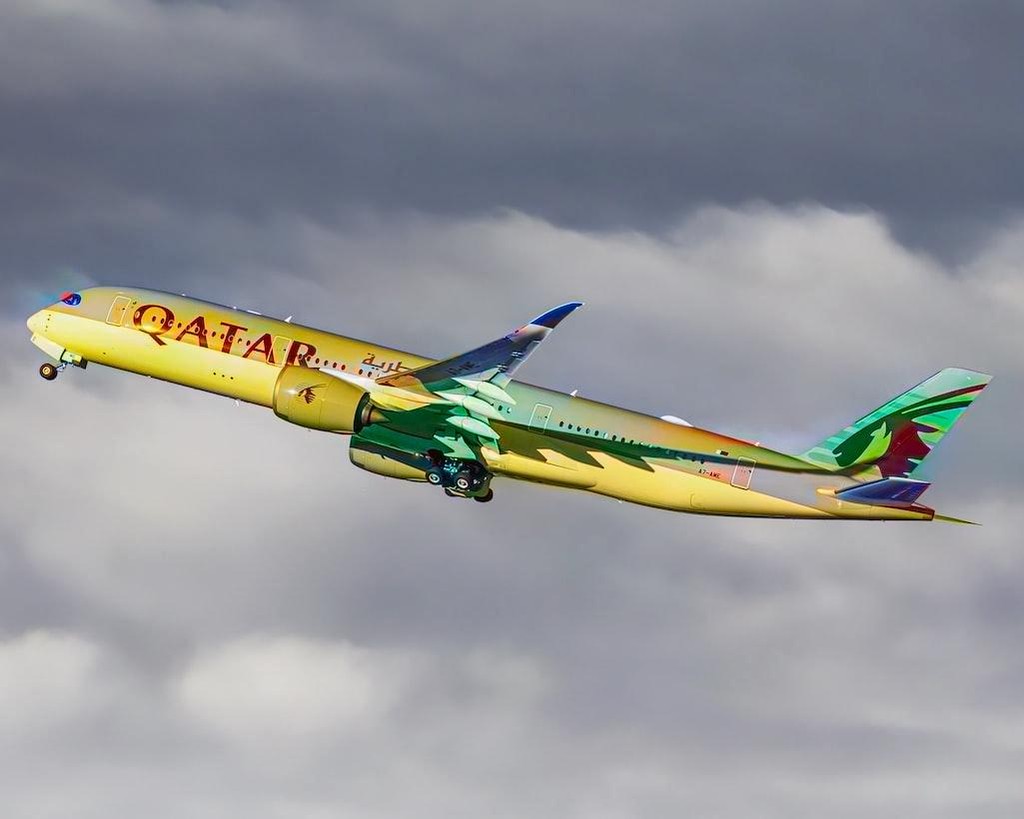}&
\includegraphics[width=\qwe,height=\qwh,frame]{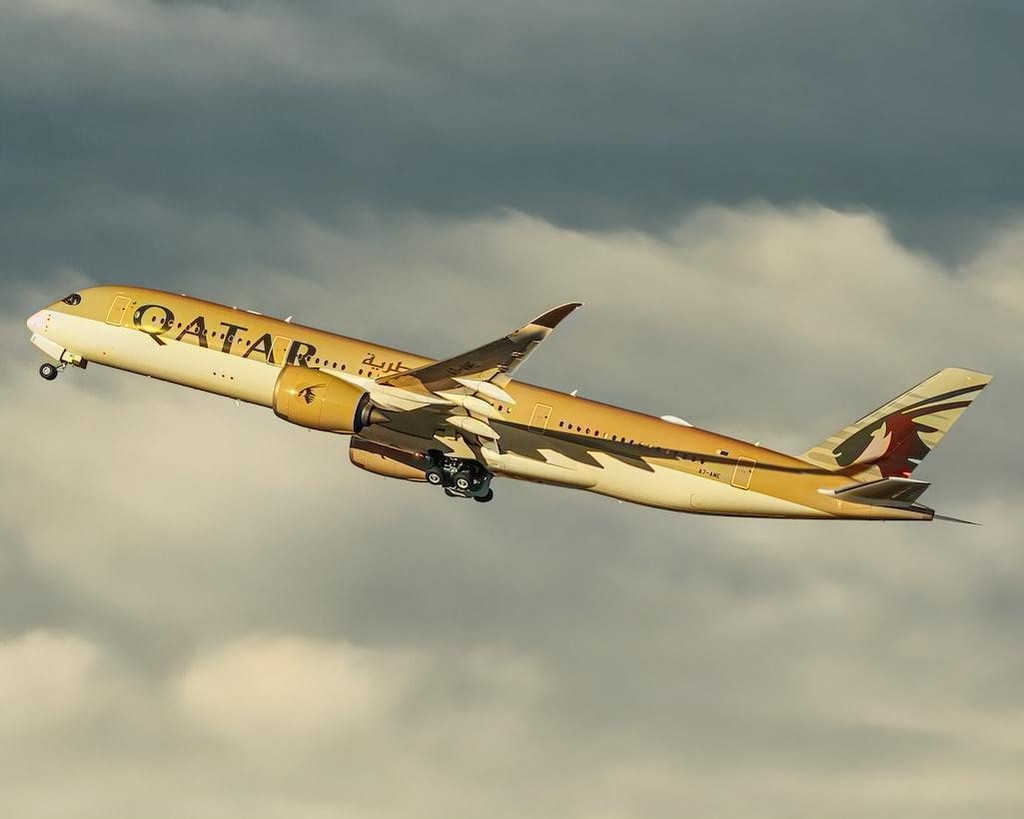} &
\includegraphics[width=\qwe,height=\qwh,frame]{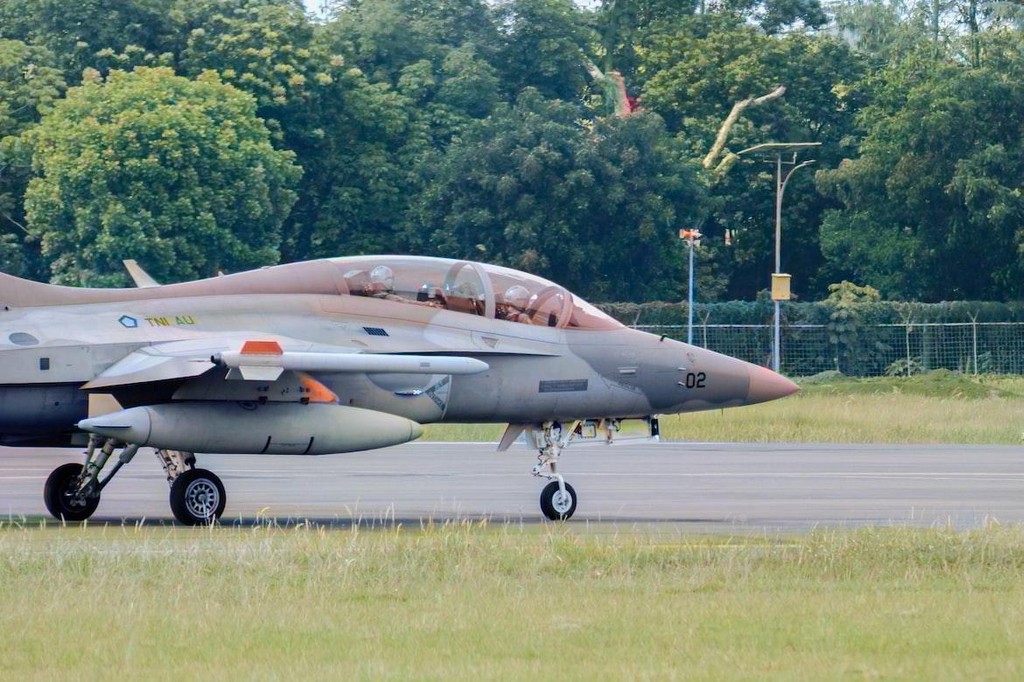} &
\includegraphics[width=\qwe,height=\qwh,frame]{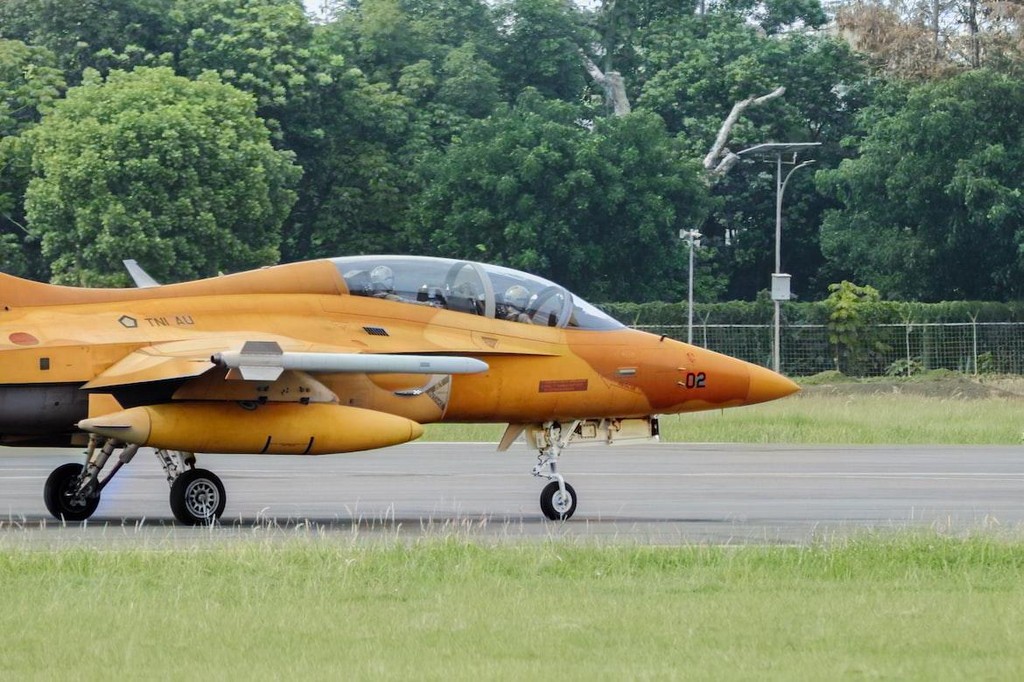}
\\
\multicolumn{2}{c}{A large \textcolor{yellow}{golden}
 airplane on the runway} & \multicolumn{2}{c}{A large \textcolor{yellow}{golden} airplane on the runway} \\

\includegraphics[width=\qwe,height=0.25\textheight,frame]{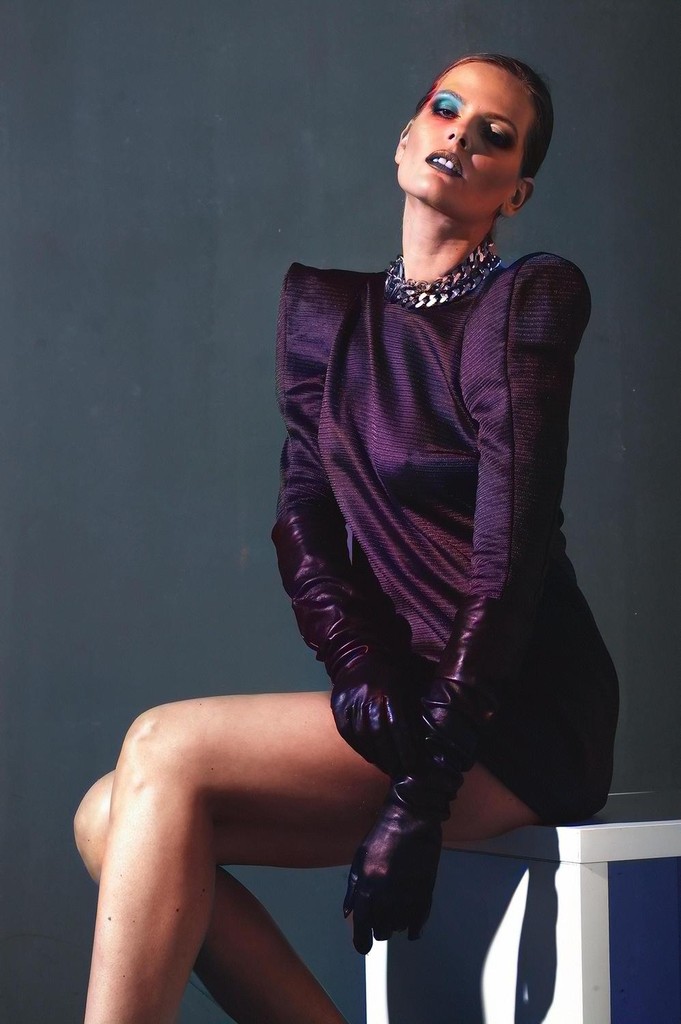}&
\includegraphics[width=\qwe,height=0.25\textheight,frame]{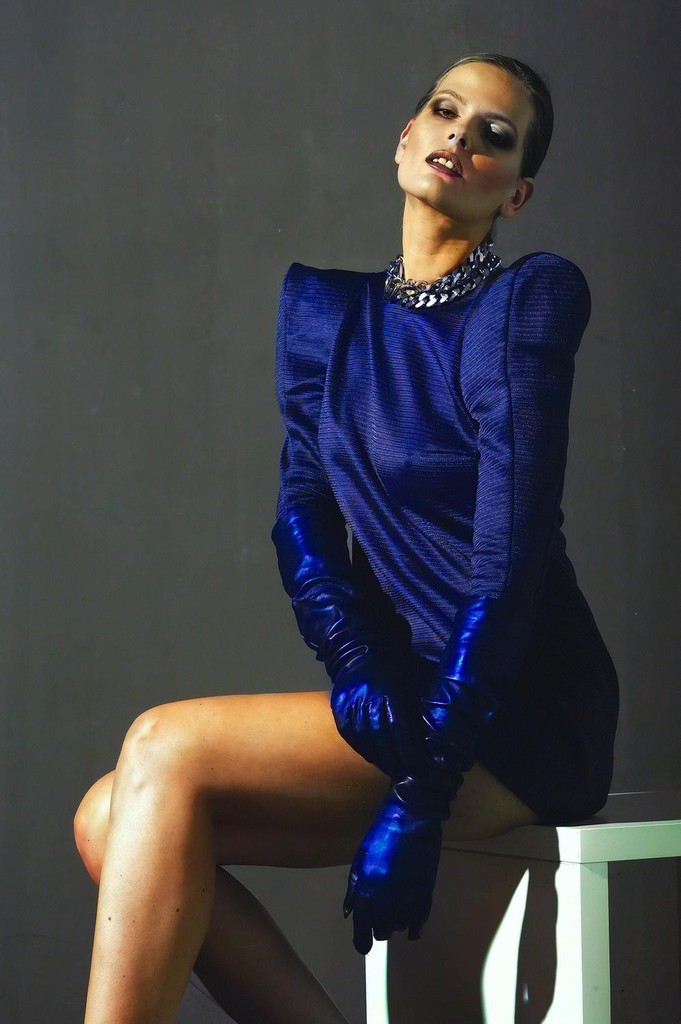} &
\includegraphics[width=\qwe,height=0.25\textheight,frame]{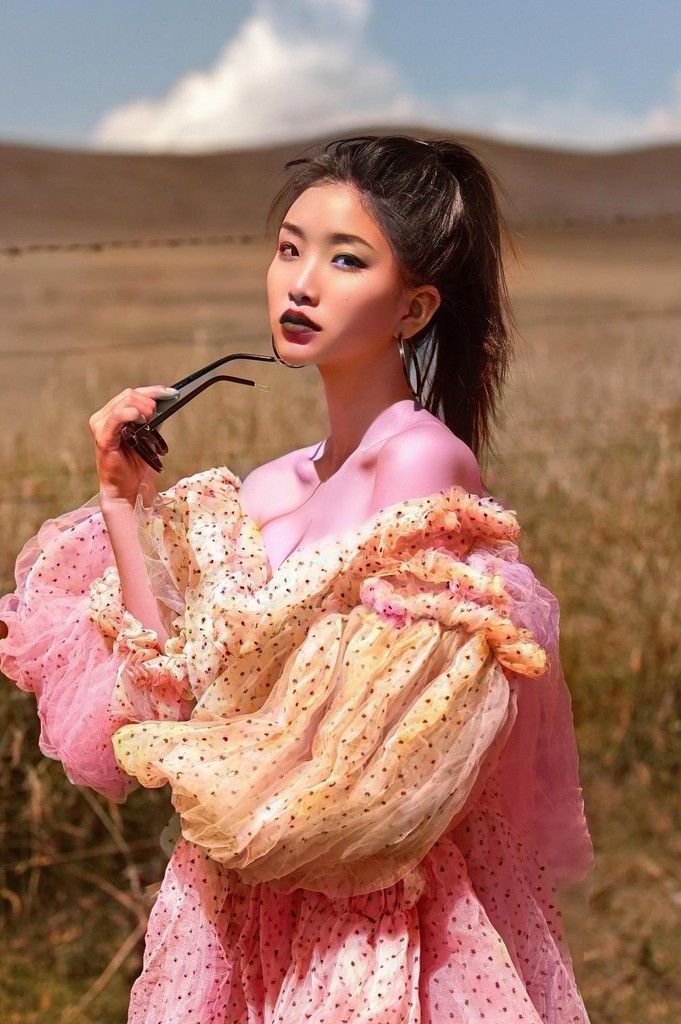} &
\includegraphics[width=\qwe,height=0.25\textheight,frame]{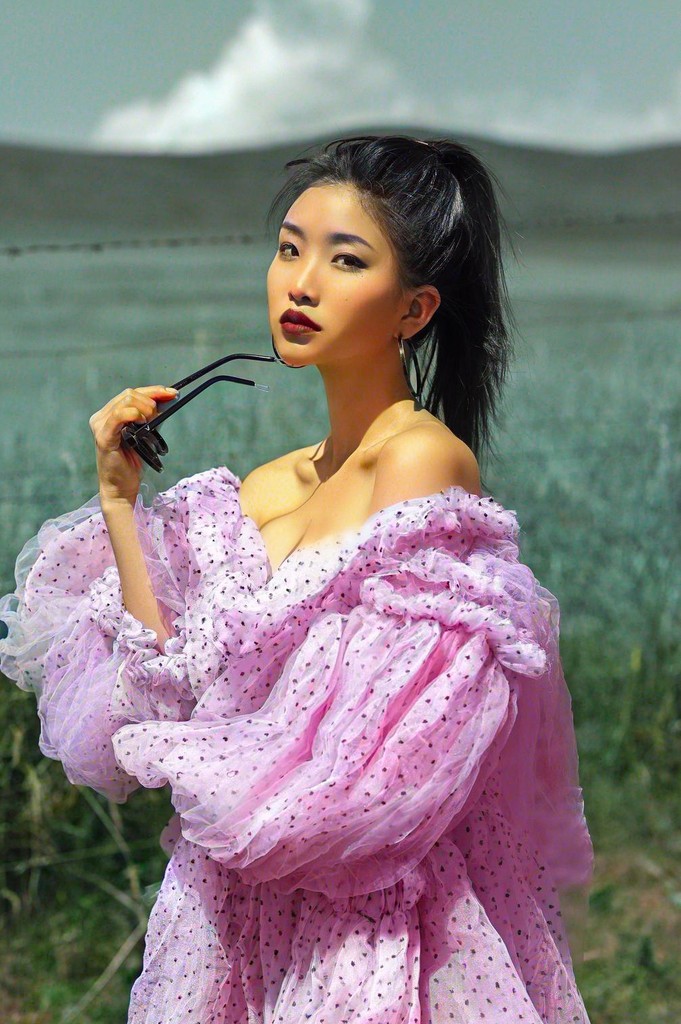} 
\\
\multicolumn{2}{c}{Woman in \textcolor{blue}{blue}
 long sleeved dress sitting on white wooden chair.
} & \multicolumn{2}{c}{Woman in \textcolor{pink}{pink} floral dress holding black sunglasses.} \\

\includegraphics[width=\qwe,height=\qwh,frame]{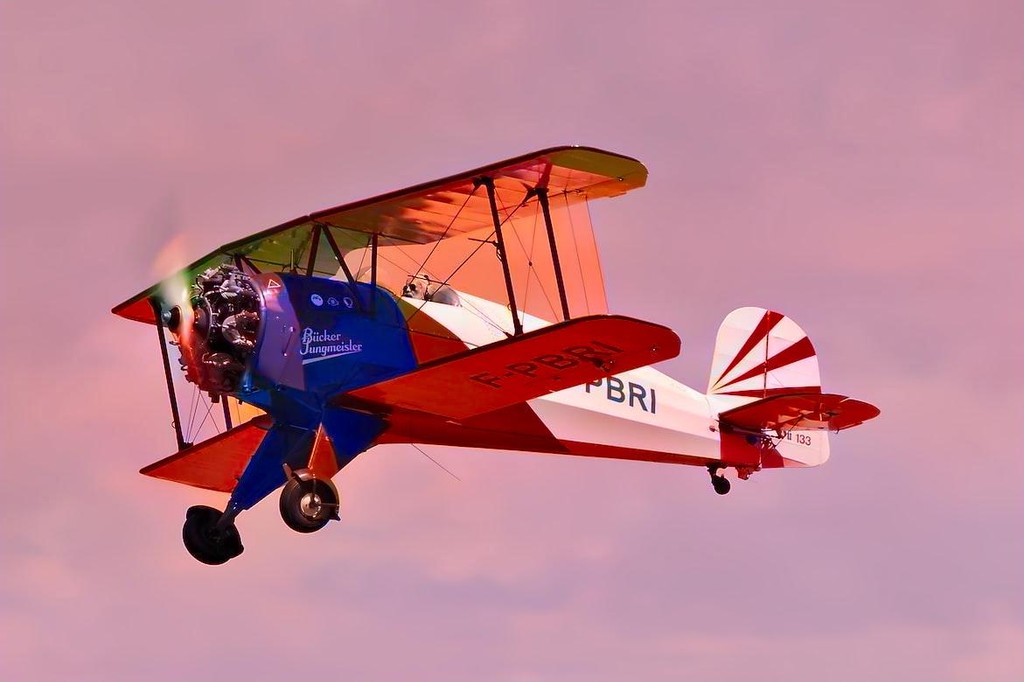}&
\includegraphics[width=\qwe,height=\qwh,frame]{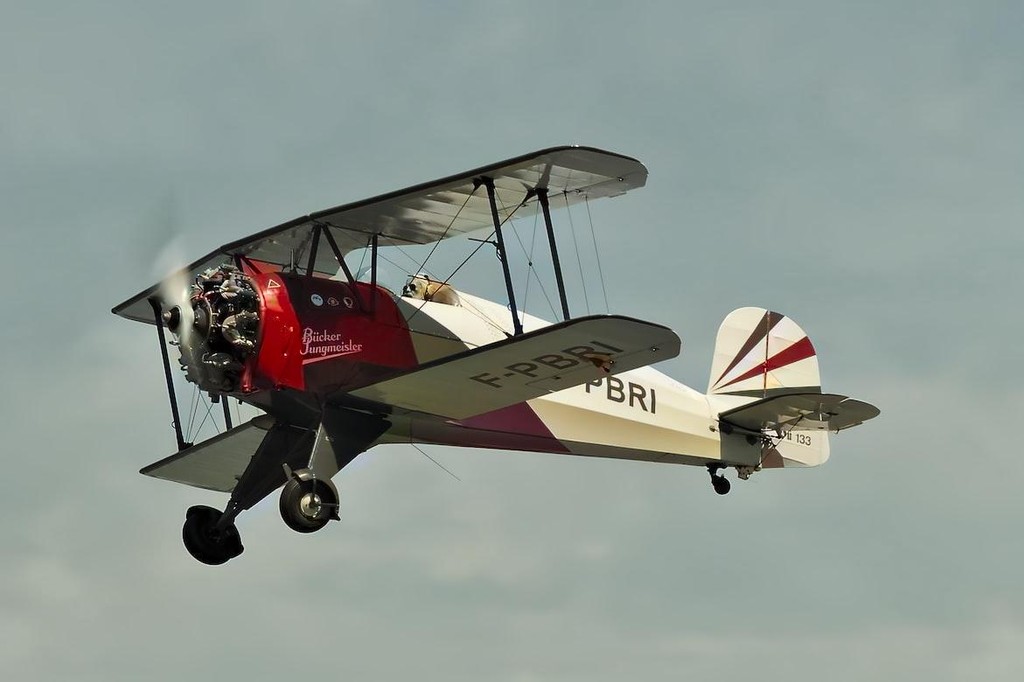} &
\includegraphics[width=\qwe,height=\qwh,frame]{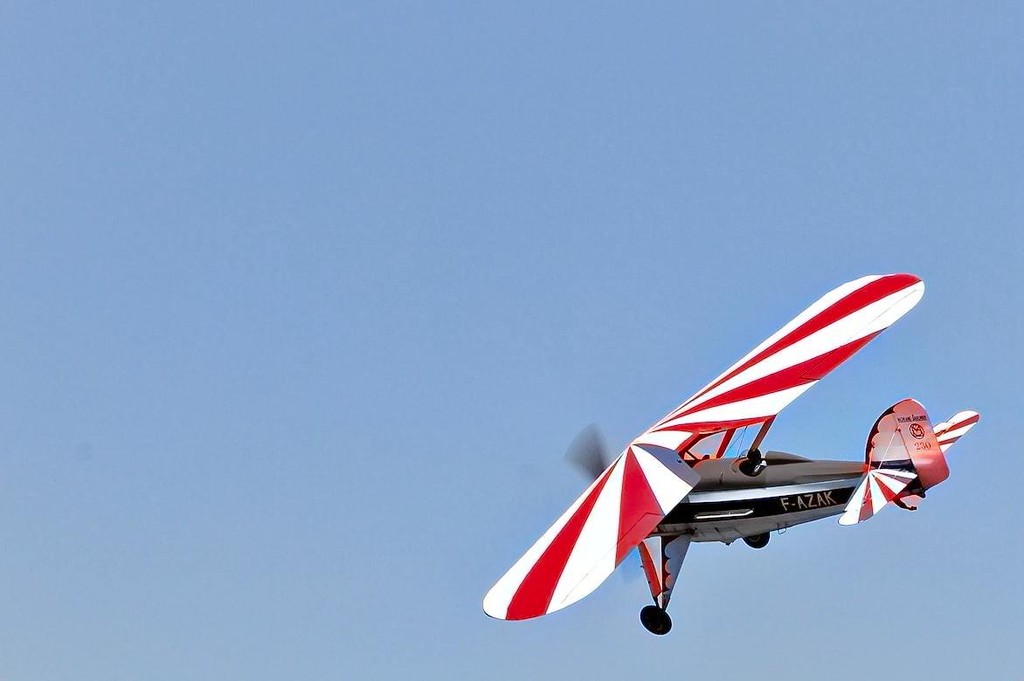} &
\includegraphics[width=\qwe,height=\qwh,frame]{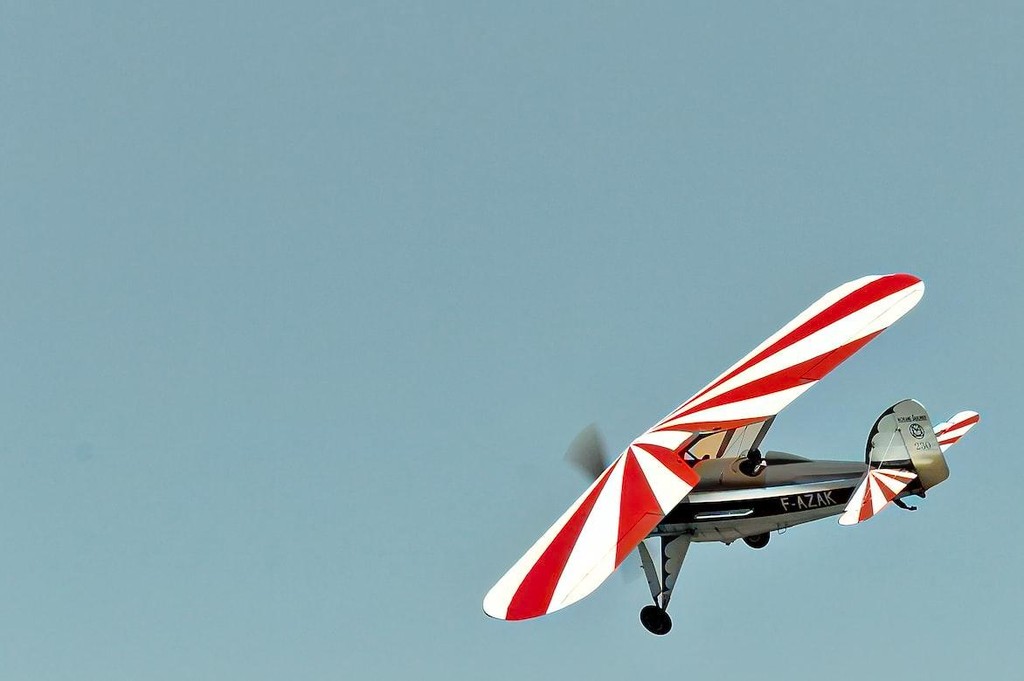} 
\\
\multicolumn{2}{c}{White and \textcolor{red}{red}
 biplane flying through the air
} & \multicolumn{2}{c}{White and \textcolor{red}{red}
 biplane flying through the air} \\

\includegraphics[width=\qwe,height=\qwh,frame]{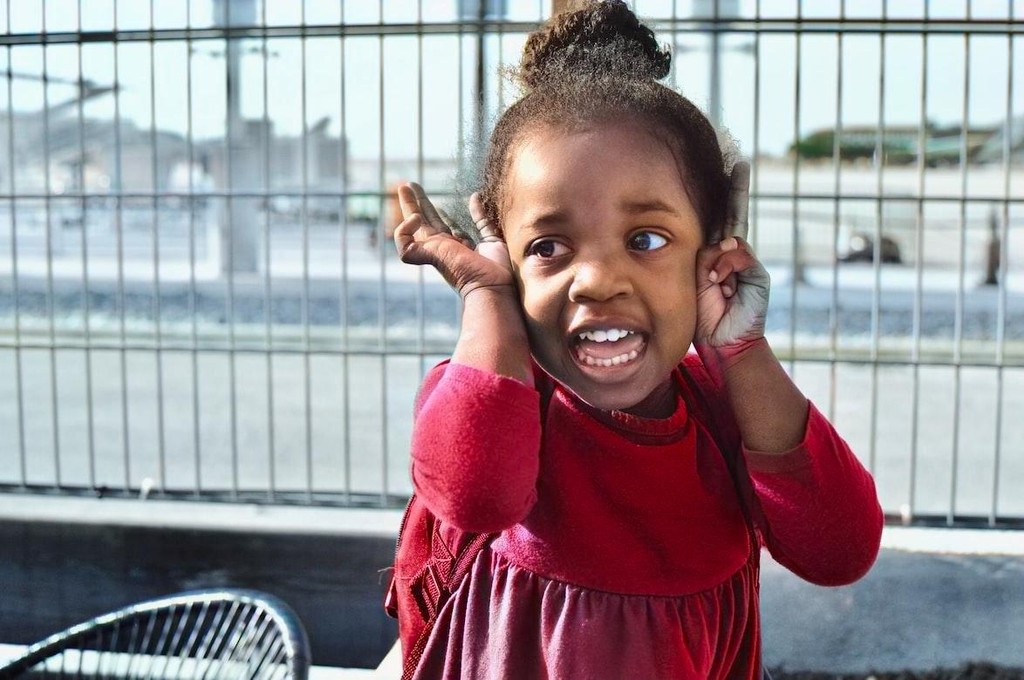}&
\includegraphics[width=\qwe,height=\qwh,frame]{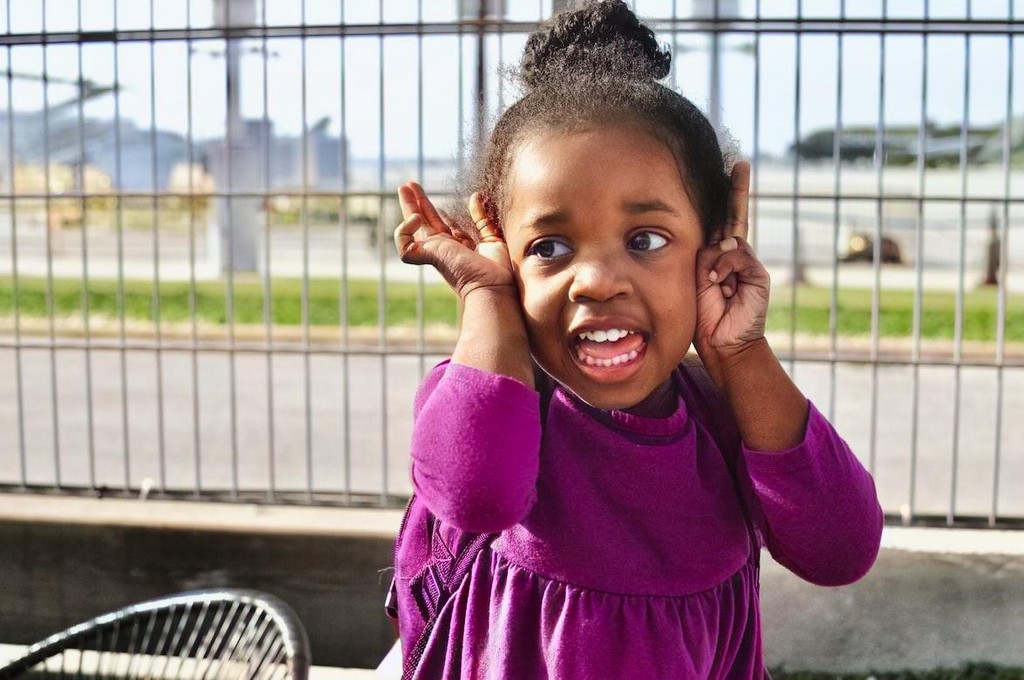} &
\includegraphics[width=\qwe,height=\qwh,frame]{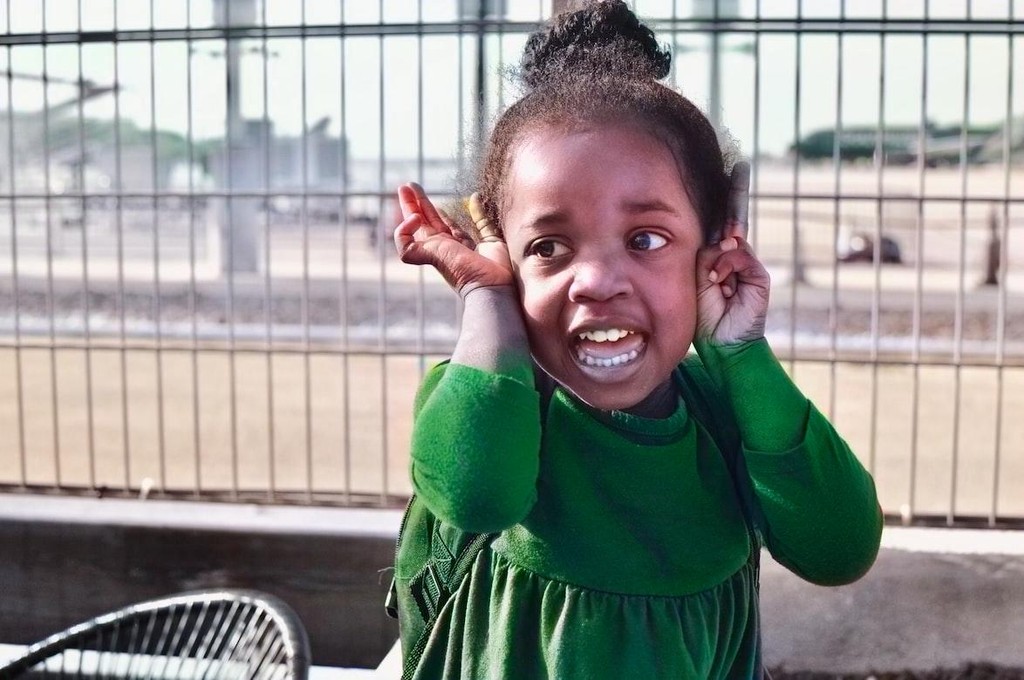} &
\includegraphics[width=\qwe,height=\qwh,frame]{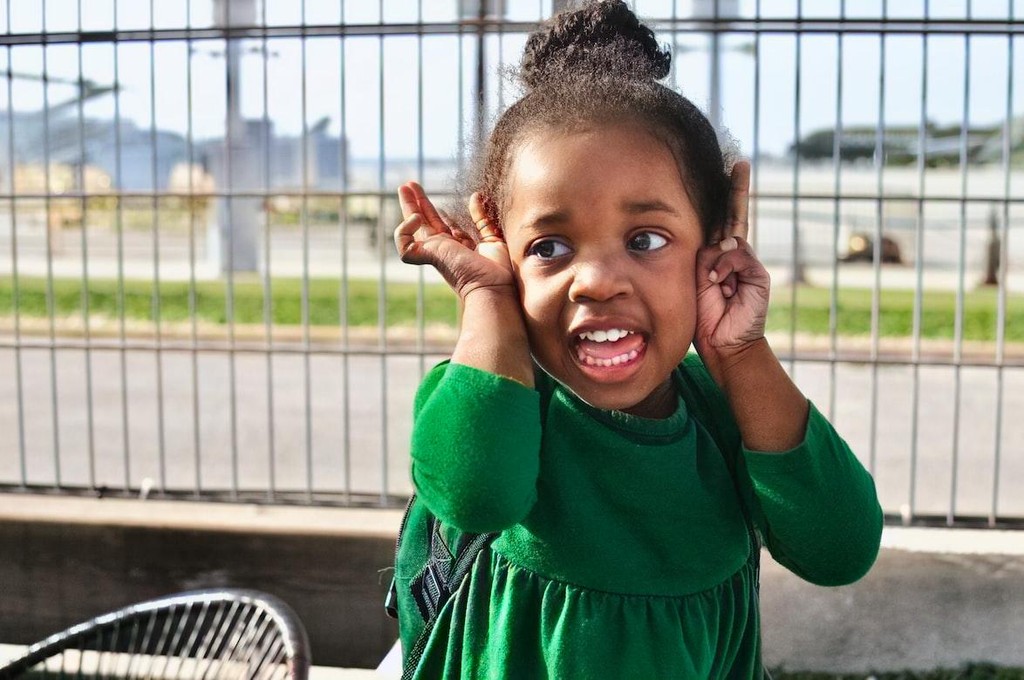} 
\\
\multicolumn{2}{c}{Toddler wearing a \textcolor{pink}{pink}
 shirt} & \multicolumn{2}{c}{Toddler wearing a \textcolor{green}{green}
 shirt} \\

\textbf{UniColor} & \textbf{Ours} & \textbf{UniColor} & \textbf{Ours} \\

\end{tabular}

\caption{
\textbf{Visual Comparison of Text Guided Colorization.} We demonstrate additional text guided colorization of our method alongside the UniColor method. Image credits from unsplash: ©Waldemar, ©Gianandrea Villa, ©Arkin Si, ©Curology, ©Thiago Gonçalves, ©Daniel Eledut, ©IIONA VIRGIN (left column) ©Dino Reichmuth, ©Christian Chen, ©Fasyah Halim, ©Dmytro Pidhrushnyi, ©Daniel Eledut, ©IIONA VIRGIN(left column). }

\label{fig:unicolor_text_guided_comparisons}
\end{figure*}

\section{User Study}
\begin{figure}
\centering
\includegraphics[width=\linewidth]{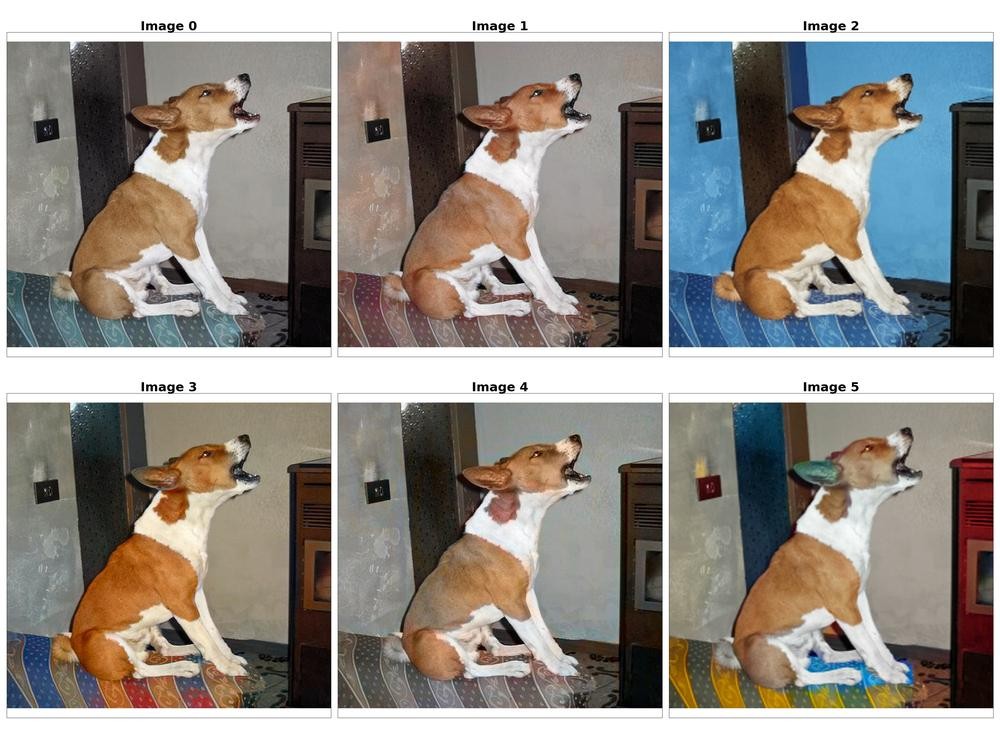}

\caption{\textbf{Survey Question Format Example.} Each query incorporates six images that are arranged in a random sequence. We maintain this random sequence to ensure we can trace back to the selected image and link it to its originating method. The participant was tasked with selecting the image they found most realistically colored from a set of six options.}

\label{fig:survey_question}
\end{figure}
In our empirical investigation, we engaged a total of 402 participants, drawn from the participant pool provided by the Surveymonkey crowdsourcing platform. A significant proportion of these participants were from the United States. Each participant was subjected to an evaluation that began with a trap question and was followed by $10$ questions chosen randomly from our expansive pool of 79 potential questions.\\
Our question pool was constructed utilizing three distinct datasets. We generated $40$ questions from the \textit{ImageNet}, $25$ from \textit{Unsplash} dataset, and $14$ from a \textit{Legacy} dataset.\\
In this trap question, we provided the participants with four images - three in color and one in grayscale. The participants were tasked to select the image that they felt was colored most realistically. Responses that failed to correctly answer this trap question were consequently omitted from further analysis.\\
Following the trap question, participants encountered $10$ additional questions. Each question presented the output generated by our proposed method alongside a set of five baseline images, all arranged randomly, as depicted in \cref{fig:survey_question}. Once again, participants were instructed to identify and select the image that exhibited the most realistic coloring.\\
By employing this study design, we aimed to assess participants' ability to differentiate the output generated by our method from the baselines, thus facilitating an evaluation of the effectiveness of our proposed approach.\\
The Surveymonkey platform\footnote{\url{https://www.surveymonkey.com}}, through its anti-bot tests, played a vital role in ensuring the integrity of our user study. These tests helped to verify that participants engaging with the survey were human respondents rather than automated bots.\\
To further enhance the credibility of our study, we implemented an additional filtering criteria based on the total filling time of the surveys and on the trap question response.
\subsection{Filtering Surveys}
Our trap question effectively recognized and excluded $13$ surveys, eliminating them from the final analysis. This reduction in the total number of surveys from $402$ to $389$ ensures that the remaining data set accurately represents the participants who correctly responded to the trap question. The exclusion of these $13$ surveys strengthens the reliability and validity of our findings by ensuring the integrity of the data used for analysis.\\
\begin{figure}

\begin{tabular}{c}
\includegraphics[width=0.5\textwidth]{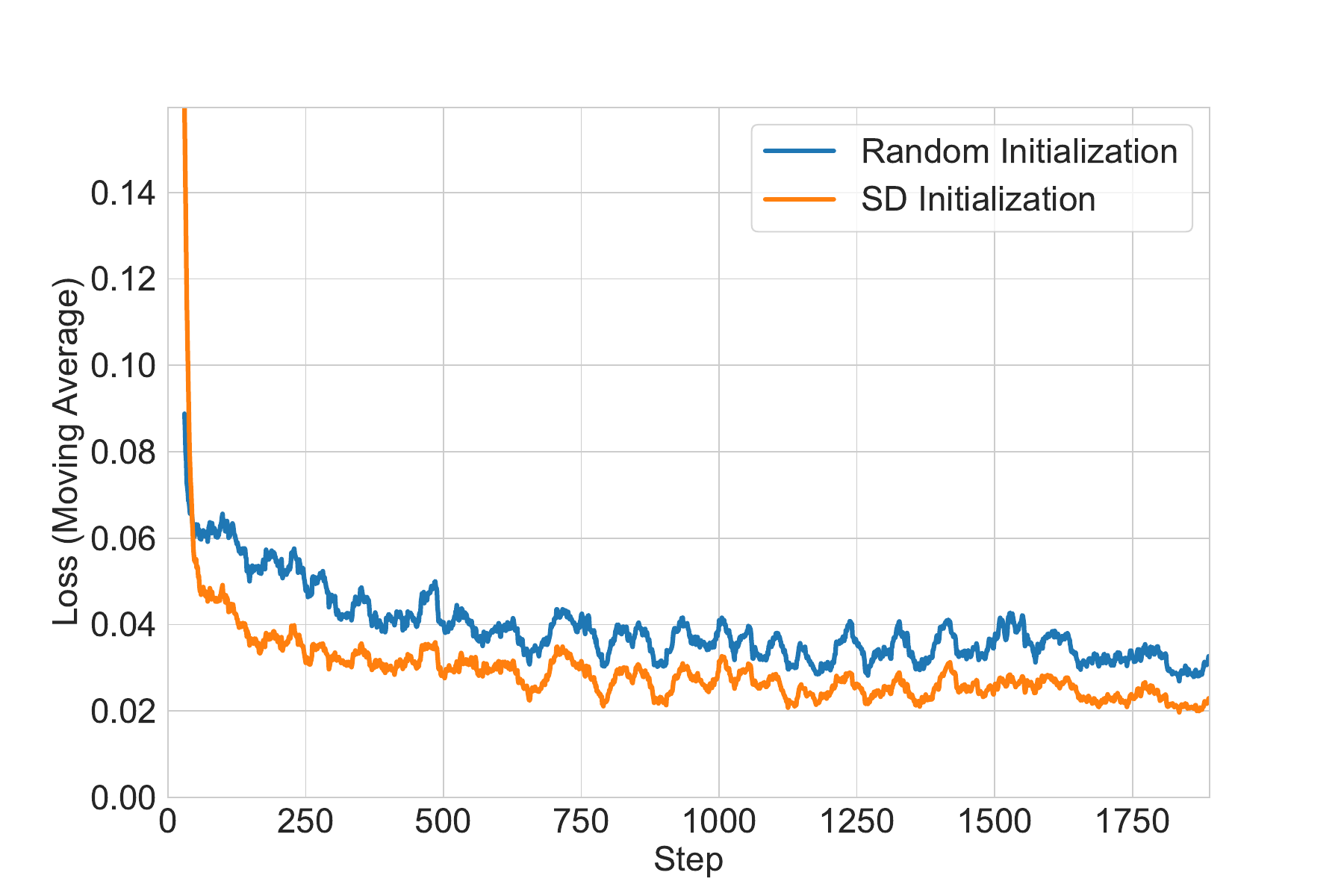} \\
\end{tabular}

\caption{
\textbf{Impact of Stable diffusion (SD) Weight Initialization.} A pretrained SD weight was used to initialize the UNet for quicker convergence. Both SD's and ours use the frozen clip text encoder, even though SD's output is residual noise and ours is residual color. With SD initialization, the UNet converged approximately 30\% faster, as demonstrated in our experiment.  }
\label{fig:convergance_model_fig}
\end{figure}

\begin{figure}

\begin{tabular}{cc@{\hspace{1pt}}c@{\hspace{1pt}}c@{\hspace{1pt}}c@{\hspace{1pt}}c@{\hspace{1pt}}c@{\hspace{1pt}}c}
  \centering

&
\includegraphics[width=\linewidth]{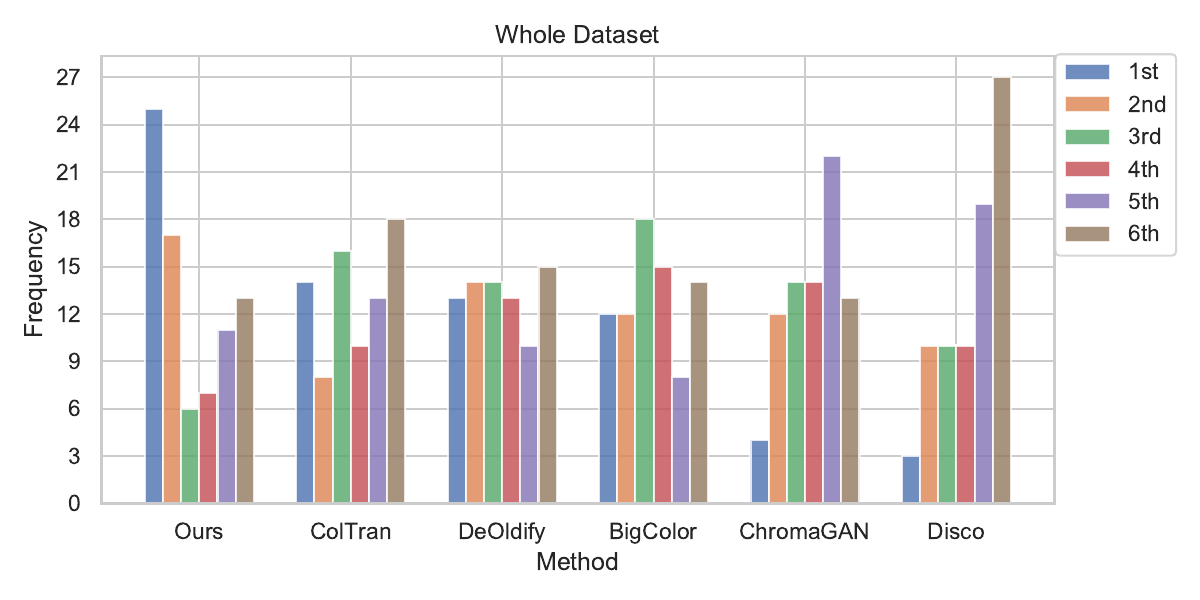}\\

&
\includegraphics[width=\linewidth]{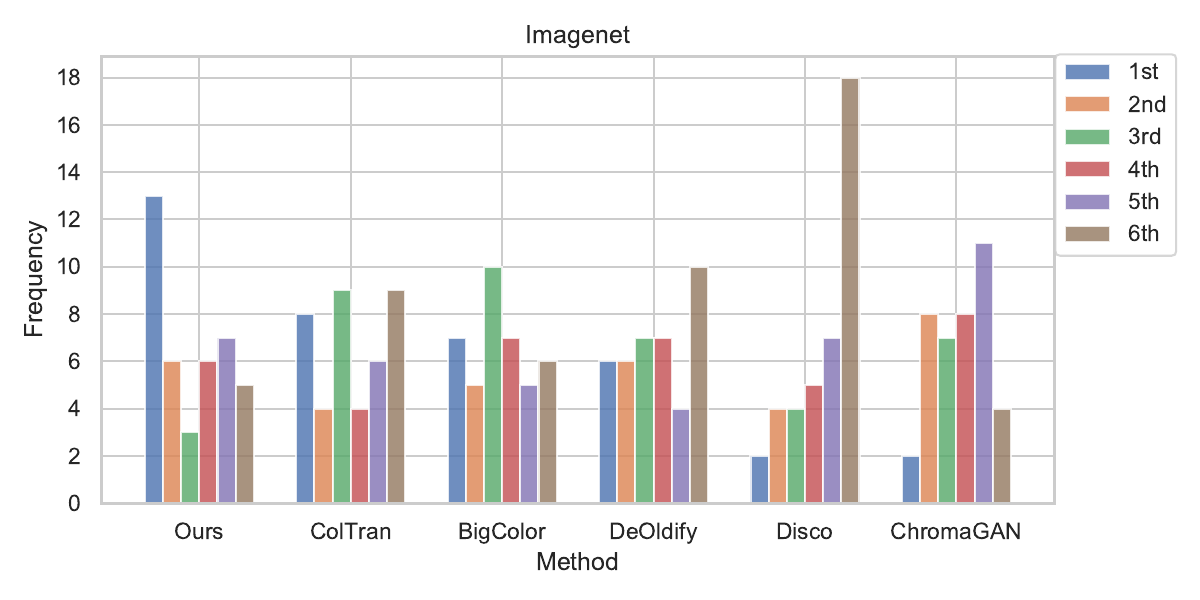}\\

&
\includegraphics[width=\linewidth]{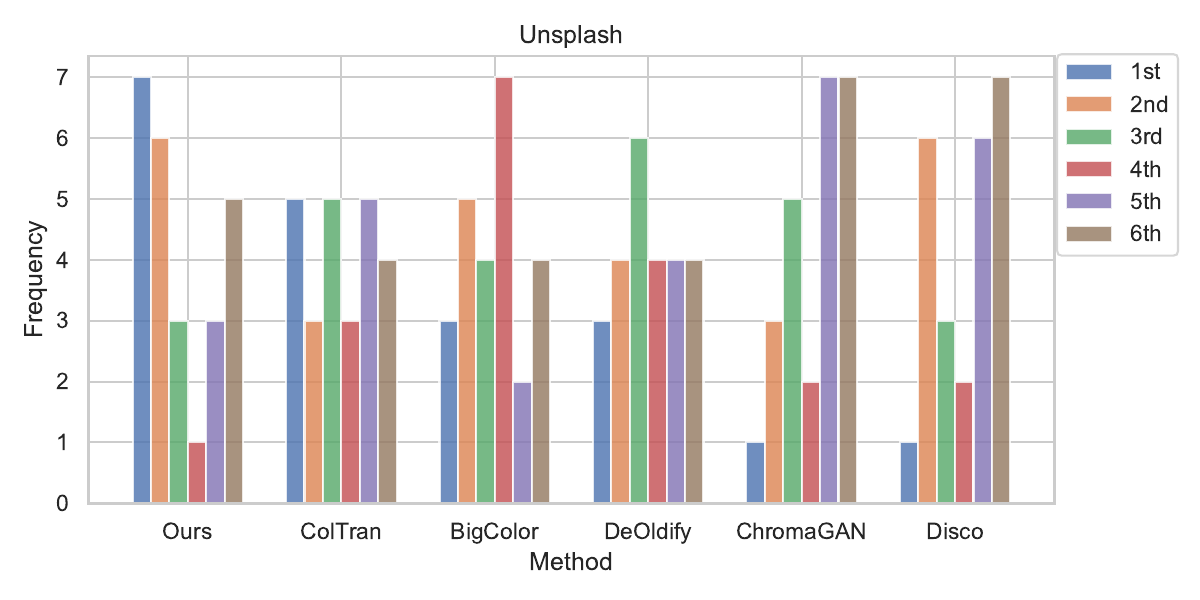}\\

&
\includegraphics[width=\linewidth]{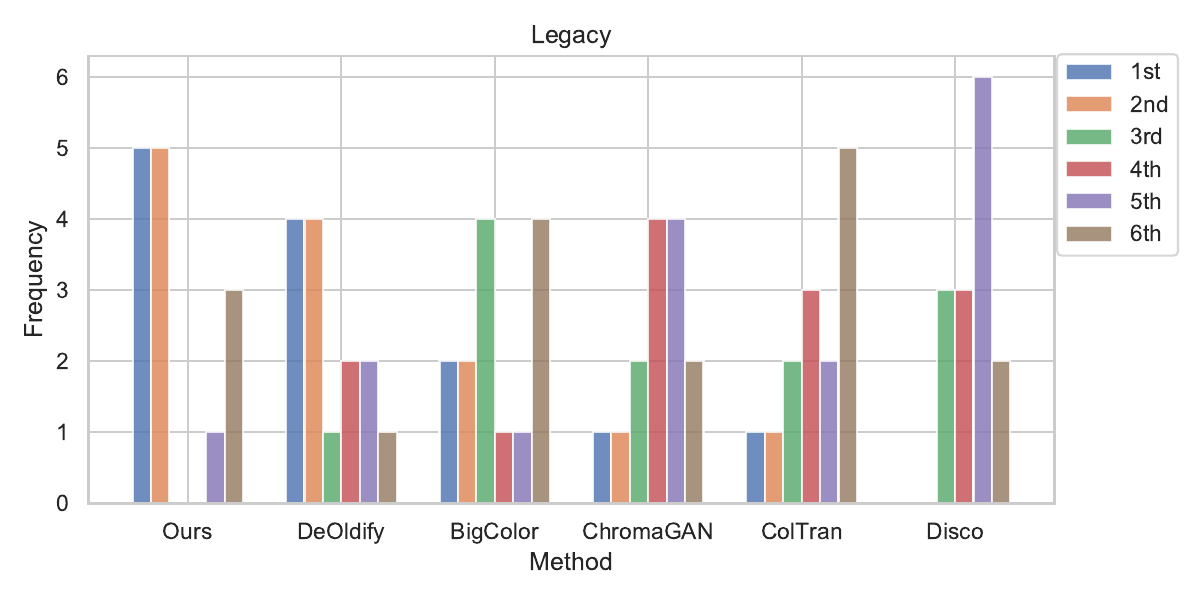}\\

\end{tabular}

\caption{
\textbf{Rankings Across Datasets.} We evaluated participants preferences by ranking the methods individually for each data set as well as for the entire data set. The x-axis illustrates the range of methods, organized from left to right in decreasing order of frequency with which each method achieved a "1st" place status (blue column).}
\label{fig:all_rankings}
\end{figure}

We assumed that participants who completed the survey in a legitimate manner would require a minimum of $70$ seconds. This time allowance allotted several seconds per question, thereby promoting thoughtful responses and eliminating hasty submissions. As a result of this filtering process, we observed a reduction in the number of acceptable surveys from $389$ to $332$. By implementing this criterion, we aimed to ensure that our analysis included only surveys completed with sufficient time and attention, thereby enhancing the validity and credibility of our study findings. The user study unequivocally demonstrated the superiority of our method compared to all the baselines, as depicted in \cref{fig:all_rankings}. We specifically focused our analysis on the most preferable method, as well as the second-best performing approach. However, in order to gain a comprehensive understanding of the rankings and ascertain the comparative performance across all methods, we extended our investigation to include the remaining rankings.\\
In certain instances, we encountered situations where two or even three methods resulted in a tie for a particular rank. To address these ties, we adopted the following approach: assume we have a tie on rank $x$, if two methods are involved, we'll assign both of them to rank $x-1$, if three are in tie, we assign them to rank $x-2$.\\
\begin{table*}[t]
  \setlength{\tabcolsep}{3pt}
  \caption{\textbf{Constant Color Scales and Ranking Approach.} we compare the performance of different metrics, contrasting the use of constant color scales with the ranking approach. Our ranking approach produces the best colorization in terms of natural colorfulness, while achieving comparable results in terms of perceptual realism as measured with FID.  Abbreviations: Grayscale (GS) and Colorfulness (CLR).}
  \begin{tabular}{lcccccccccccc}
    \toprule
    \textbf{Method} & \multicolumn{6}{c}{ImageNet (10k)} & \multicolumn{6}{c}{COCO-Stuff (5k)} \\
    \cmidrule(lr){2-7} \cmidrule(lr){8-13}
    Scale &  FID ($\downarrow$) & CLR ($\uparrow$) & $\Delta$ CLR ($\downarrow$) & PSNR ($\uparrow$) & SSIM ($\uparrow$ )& LPIPS ($\downarrow$) &  FID ($\downarrow$)& CLR ($\uparrow$) & $\Delta$ CLR ($\downarrow$) & PSNR ($\uparrow$)& SSIM ($\uparrow$) & LPIPS ($\downarrow$)\\
    \midrule
    GT &  0 & 41.614 & 0 & $\infty$ & 1 & 0 &  0 & 41.028 &0 & $\infty$ & 1 & 0 \\
    \midrule
    0.0 (GS) &  16.67 & 1.676 & 39.938 & 22.672 &  0.909 &  0.226 &  26.126 &  1.201 & 39.827 &  23.121 &  0.908 &  0.224 \\
    0.2 &  10.225 & 4.647 & 36.967 &  23.121 &  0.912 &  0.207 &  18.813 &  4.119 & 36.909 & 23.425 &  0.91 &  0.211 \\
    0.4 &  5.057 &  12.104 & 29.510 & 23.635 &  \textbf{0.9148} &  \textbf{0.184} & 12.381 & 11.269 &  29.759 &  23.846 &  0.912 &  0.190 \\
    0.6 &  3.772 &  21.816 & 19.798  &  \textbf{23.744} &  0.9142 &  0.175 & 9.35 &  20.501 & 20.527  &  \textbf{24.024} &  \textbf{0.913} &  \textbf{0.180}\\
    0.8 &  \textbf{3.527} &  33.26 & 8.354  &  23.133 &  0.9053 &  0.184 & 8.175 & 31.437 & 9.595 &   23.556 &  0.907 &  0.186\\
    1.0 &  3.653 &  45.429 &  3.815 & 21.824 &   0.8841 &  0.205 & \textbf{7.907}&  43.24 & 2.212 &  22.355 &  0.891 &  0.203\\
    1.2 &  3.965 &  57.167 &  15.553 & 20.280 &  0.852 &  0.231 & 8.145 &  54.789 & 13.761 & 20.859 &  0.865 &  0.227\\
    1.4 &  4.417 &  \textbf{67.896} & 26.282 & 18.864 & 0.815 &  0.258 & 8.733 &  \textbf{65.322} & 24.294 &  19.449 &  0.833 &  0.253\\
    \cmidrule(lr){2-7} \cmidrule(lr){8-13}
    Ranking &  3.69 & 41.404 & \textbf{0.21} & 22.042 & 0.889 & 0.201 &  8.05 & 39.34& \textbf{1.688} & 22.57 & 0.895 & 0.200\\
    \bottomrule
  \end{tabular}
  \label{tab:color_scale_ranking}
\end{table*}

\begin{table}[h]
\caption{We run grid search on the CFG guidance scale and scaling in the latent space on imagenet-cval1k. We find the best performance with scale=$0.8$ and guidance scale=$1.6$.}
\centering
\begin{tabular}{cc|ccccc}
\multicolumn{2}{c}{} & \multicolumn{5}{c}{\textbf{Scale}} \\  
\multirow{3}{*}{\rotatebox[origin=c]{90}{\textbf{CFG Scale}}}
 & & \textbf{0.6} & \textbf{0.8} & \textbf{1.0} & \textbf{1.2} & \textbf{1.4} \\ \cline{2-7}
& \textbf{1.2} & 25.977 & 24.522 & 24.077 & 24.114 & 24.317 \\ 
& \textbf{1.4} & 24.939 & 23.971 & 23.913 & 24.137 & 24.641 \\ 
& \textbf{1.6} & 24.261 & \textbf{23.741} & 23.922 & 24.457 & 25.147 \\ 
\end{tabular}

\label{tab:grid_search_cfg}
\end{table}

\subsection{Elo Rating Calculation}
To calculate the Elo rating, we need to calculate the expected outcome (EO) of a game given current Elo rating for the two competeing methods.
\begin{equation}
\text{{EO}}^{12} = \frac{1}{1 + \left(10^{\frac{{\text{{method2\_elo}} - \text{{method1\_elo}}}}{400}}\right)}
\end{equation}

We then sample comparisons between our 6 methods, and update the Elo for each method until convergence based on the following update rule:
\begin{equation}
\begin{aligned}
\text{{new\_elo\_method1}} &= \text{{old\_elo\_method1}} \\
&\quad + \text{{lr}} \cdot (\text{{game\_result}} - \text{{EO}}^{12})
\end{aligned}
\end{equation}

\begin{equation}
\begin{aligned}
\text{{new\_elo\_method2}} &= \text{{old\_elo\_method2}} \\
&\quad + \text{{lr}} \cdot (\text{{game\_result}} - \text{{EO}}^{21})
\end{aligned}
\end{equation}
Where $lr$ was set to be $0.1$, and \text{{game\_result}} is $0$ for a win and $1$ for a loss.
All methods were initialized at a rating of 1500 ELO.

\begin{figure*}
  \centering
  \setlength{\tabcolsep}{5pt}

  \newlength{\imgwidth}
  \newlength{\imgheight}
  \setlength{\imgwidth}{4cm}
  \setlength{\imgheight}{7cm}

  \begin{tabular}{cccc}
    \includegraphics[width=0.23\linewidth, height=3.5cm, frame]{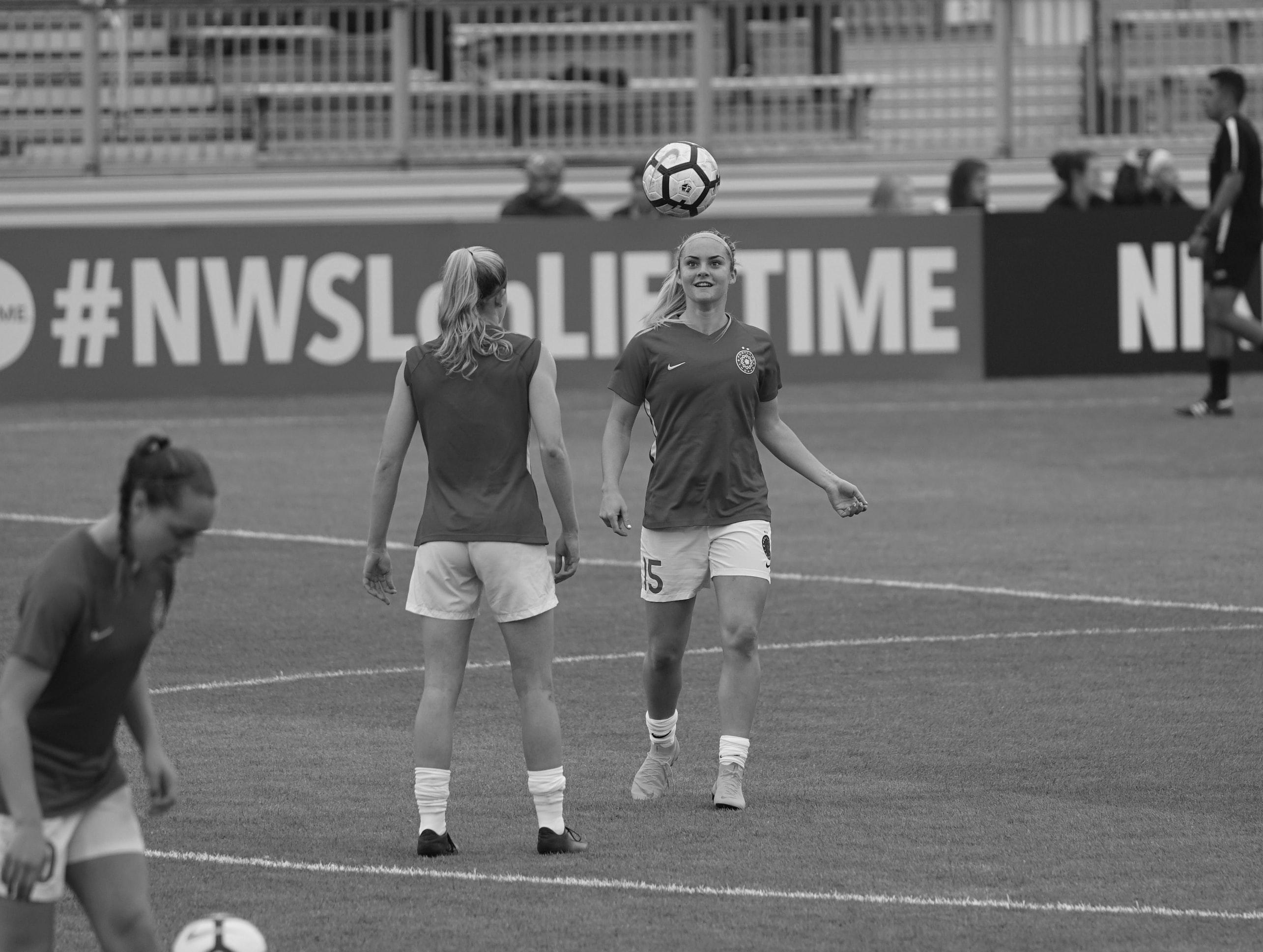} \hspace{-10pt} &
    \includegraphics[width=0.23\linewidth, height=3.5cm, frame]{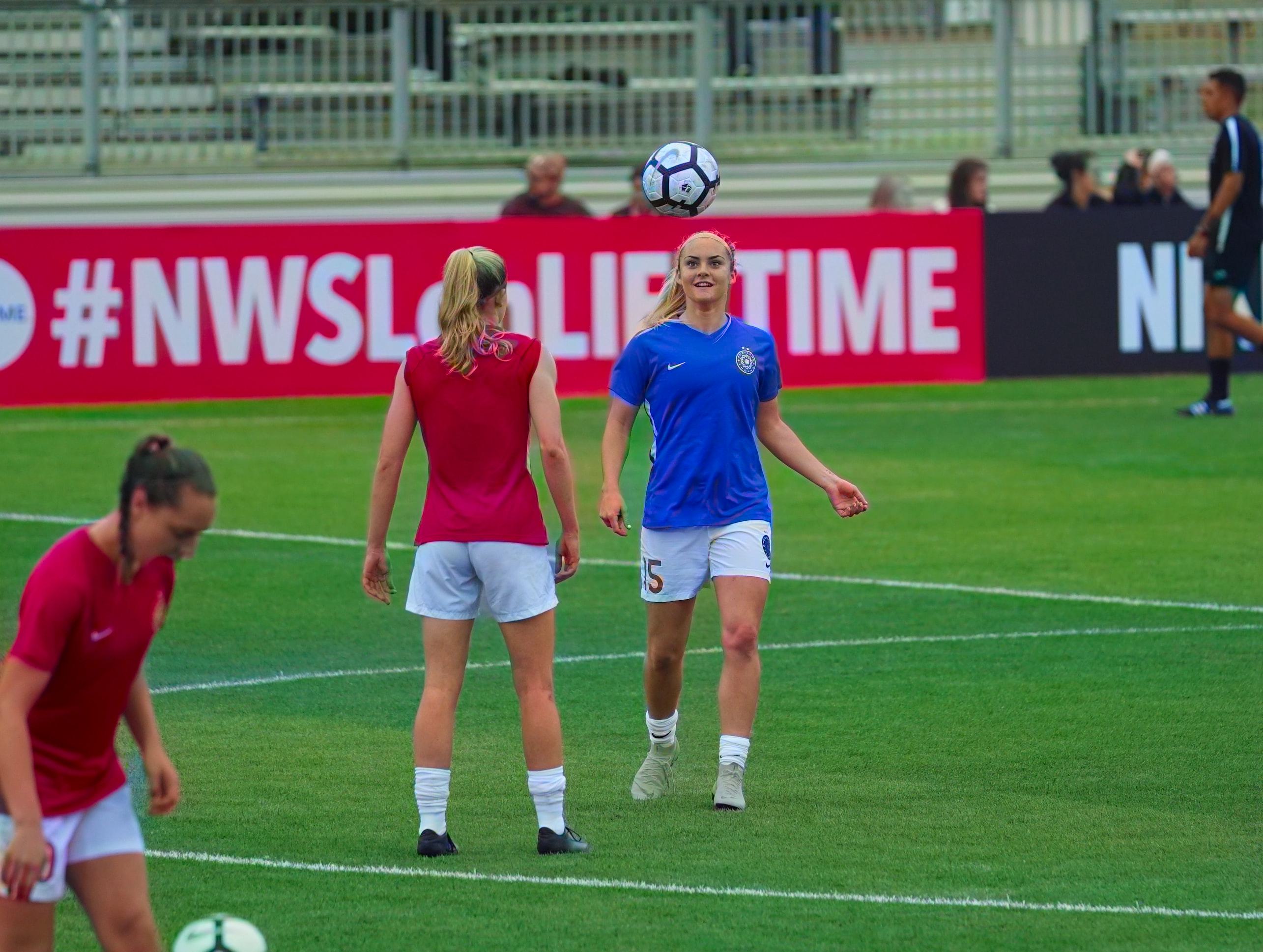} &
    \includegraphics[width=0.23\linewidth, height=3.5cm, frame]{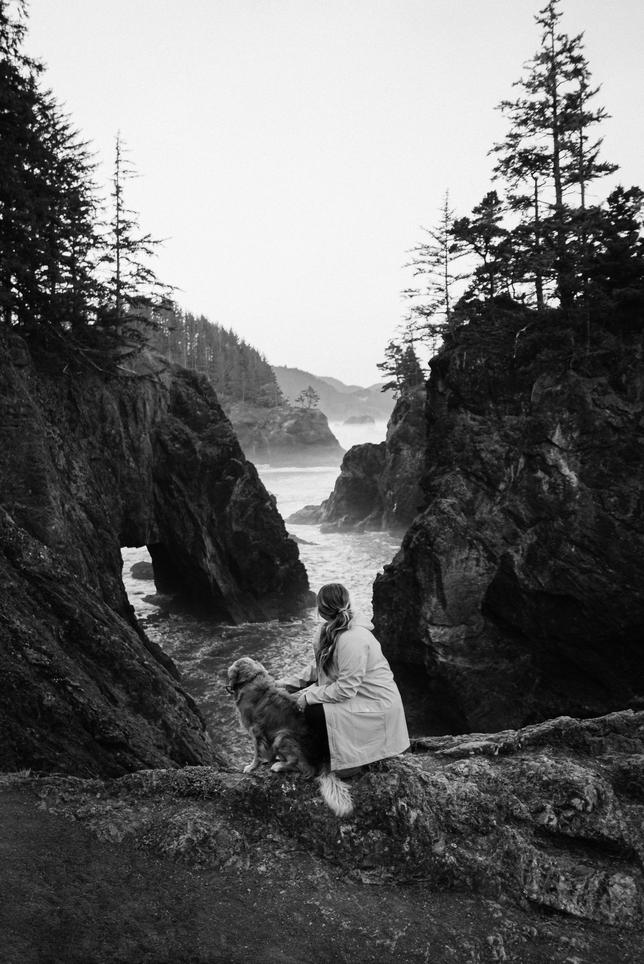} \hspace{-10pt} &
    \includegraphics[width=0.23\linewidth, height=3.5cm, frame]{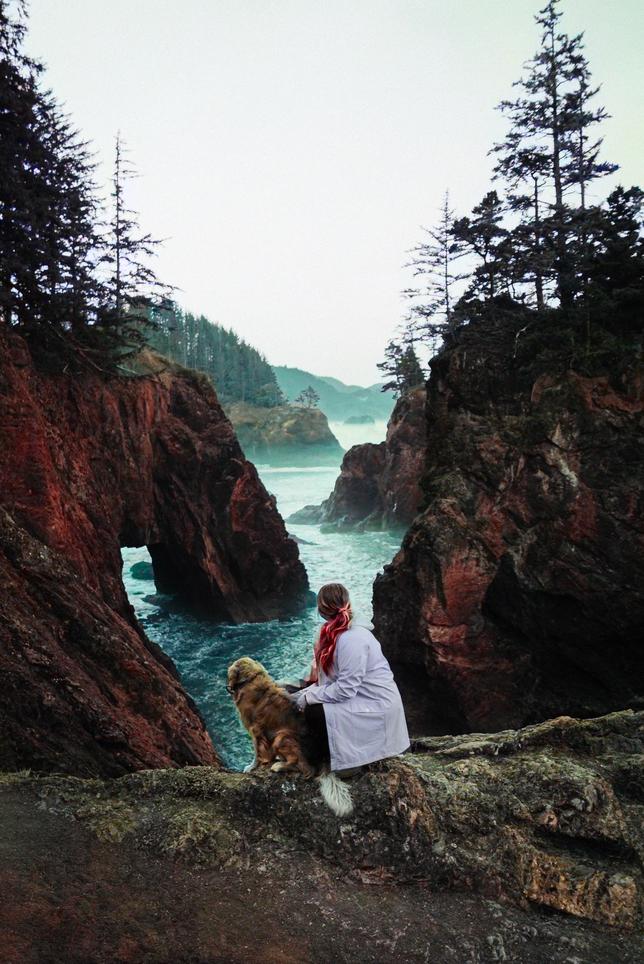} \\
    
    \multicolumn{2}{c}{\makecell{Left player with red shirt, middle player in white \\shirt, right player in blue shirt}} &
    \multicolumn{2}{c}{a woman with a red coat in a rainy day} \\

    \includegraphics[width=0.23\linewidth, height=6cm, frame]{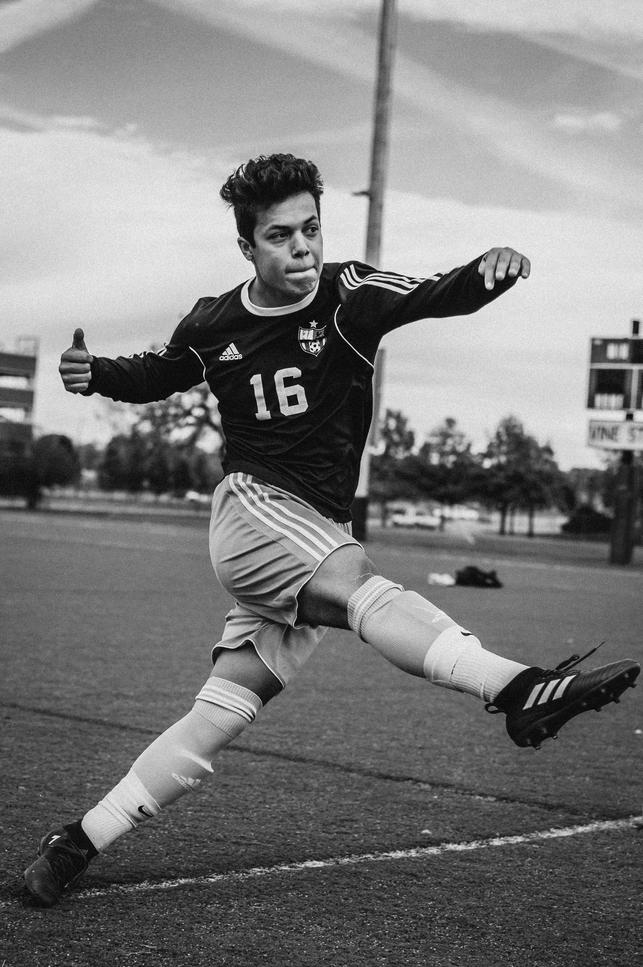} \hspace{-10pt} &
    \includegraphics[width=0.23\linewidth, height=6cm, frame]{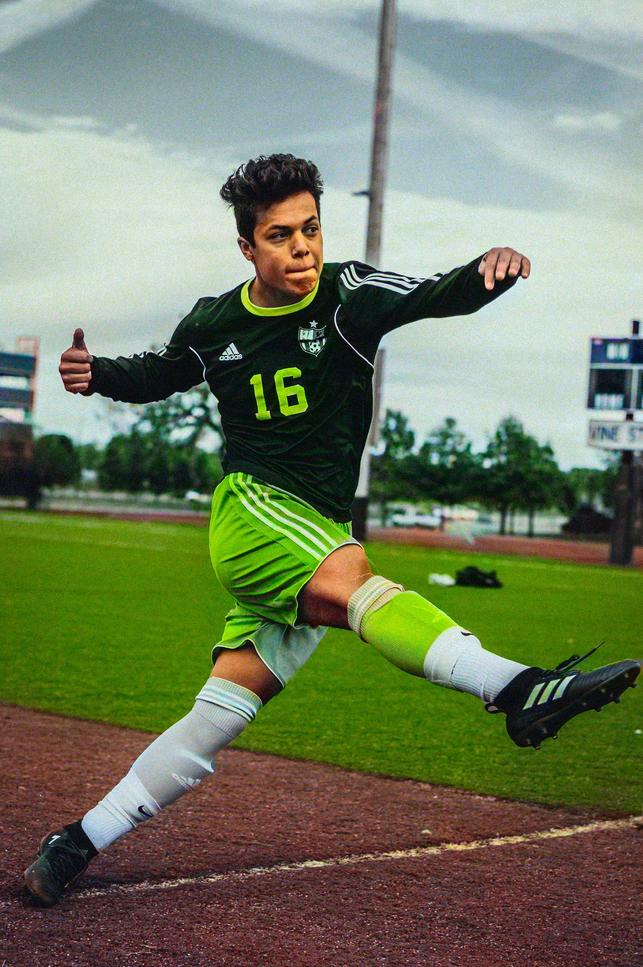} &
    \includegraphics[width=0.23\linewidth, height=6cm, frame]{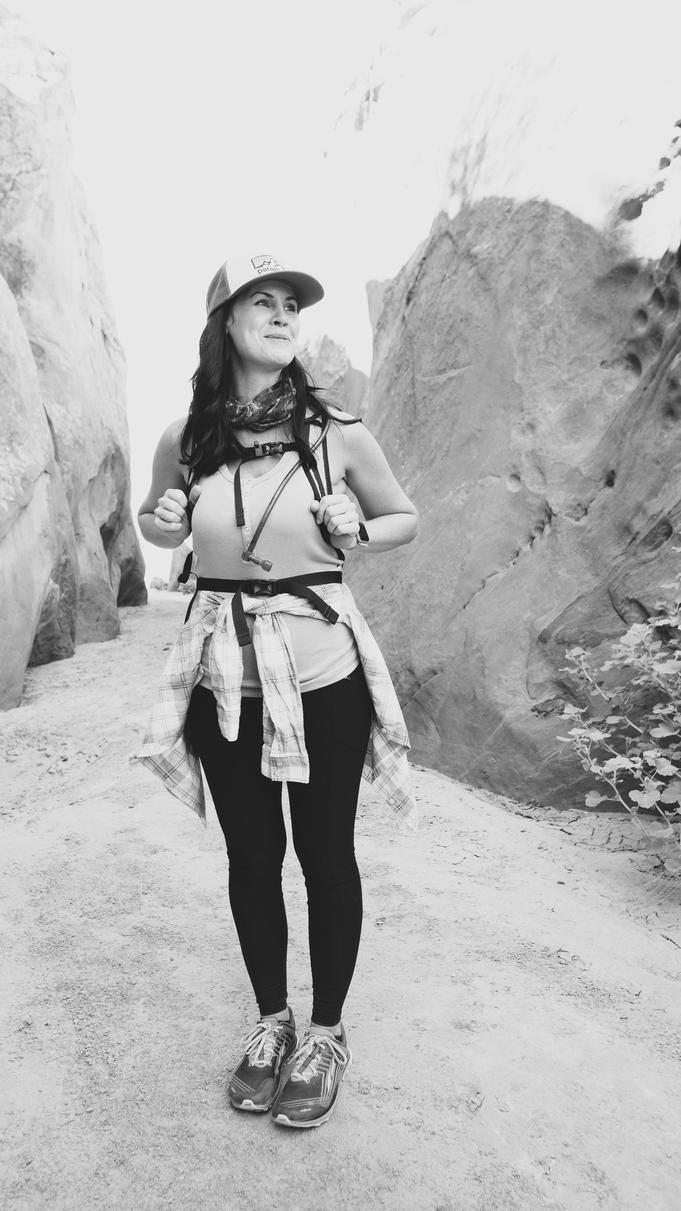} \hspace{-10pt} &
    \includegraphics[width=0.23\linewidth, height=6cm, frame]{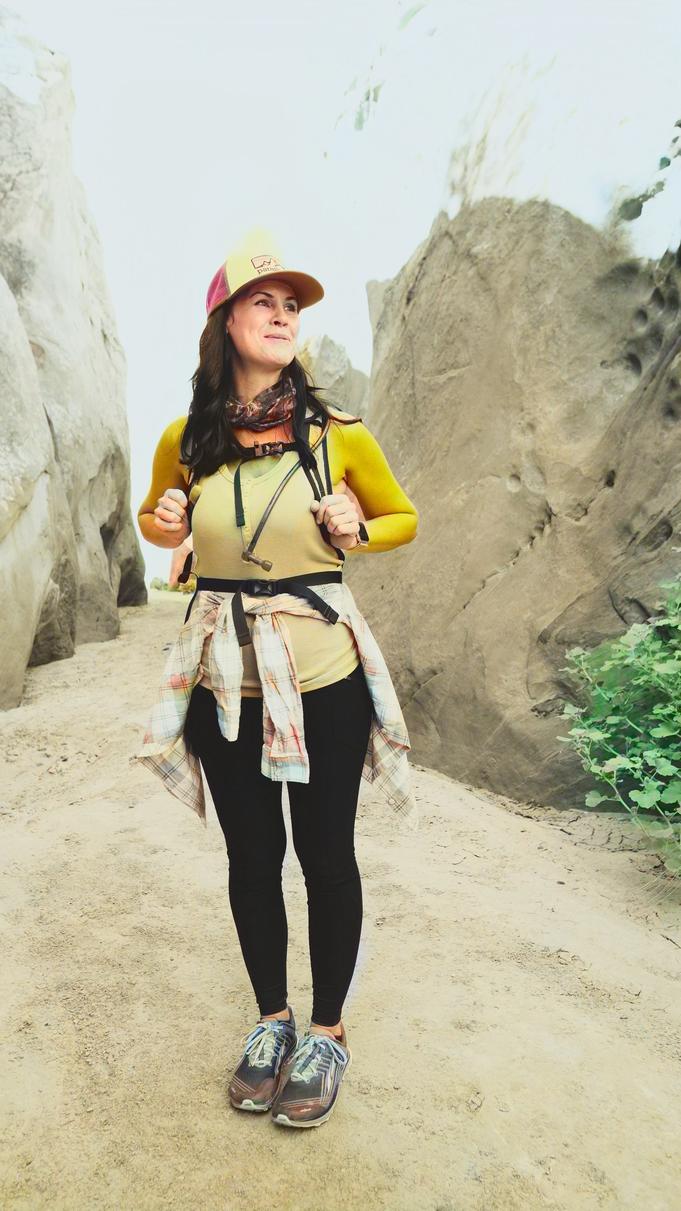} \\

    \multicolumn{2}{c}{\makecell{A football player on a yellow field with\\ black shirt green pants and red socks}} &
    \multicolumn{2}{c}{Hiker with red pants, yellow shirt and blonde hair} \\
  \end{tabular}

  \caption{\textbf{Limitations.} Our methods are challenged in colorizing images where the prompt directs for a complex spatial layout (i.e., 1st and 3rd pair). It might also not produce the requested colors for specific objects when their grayscale value is in stark contrast to the requested color, as evident in the bright coat of the 2nd pair and the black pants in the 4th pair. Image credits: Unsplash ©Jeffrey F Lin, Unsplash ©Karsten Winegeart, Unsplash ©Ben Weber, Unsplash ©Joshua Gresham.}
  \label{fig:text_limitations}
\end{figure*}

\begin{table}[b]
\setlength{\tabcolsep}{5pt}
\caption{\textbf{Effect of Prompts.} We measured the CLIP similarity between  images and their auto generated captions, to establish a reference (1st row). Next, we re-colorized using a fixed prompt (2nd row), and the auto generated captions (3rd row). This demonstrated the effectiveness of text prompts to guide the colorization process.}

\begin{center}
\begin{tabular}{ccc}
\toprule
 &   \textbf{ImageNet (10K)} & \textbf{COCO-Stuff (5k)} \\
\midrule
Original &  $0.328$ & $0.323$\\
Ours (fixed prompt) &  $0.314$ & $0.300$ \\
Ours (GT prompts) & $0.324$ & $0.309$ \\
\bottomrule
\end{tabular}
\end{center}

\label{table:clip_similarity}
\end{table}

\section{Additional Evaluation Details}

As demonstrated in \Cref{table:clip_similarity}, using prompts improves the cosine similarity between the output image and text, indicating the level of controllability provided by the prompts. 

\begin{figure*}
\setlength{\tabcolsep}{1pt}

\newlength{\wwa}
\setlength{\wwa}{0.159\linewidth}

\begin{tabular}{cccccc}
  \centering

\includegraphics[width=\wwa,frame]{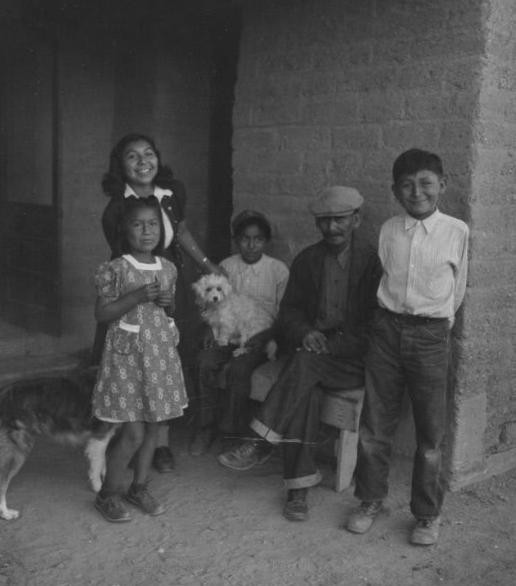} &
\includegraphics[width=\wwa,frame]{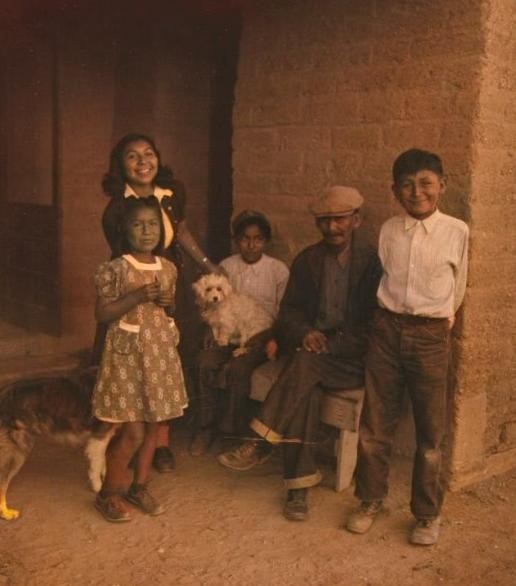}&
\includegraphics[width=\wwa,frame]{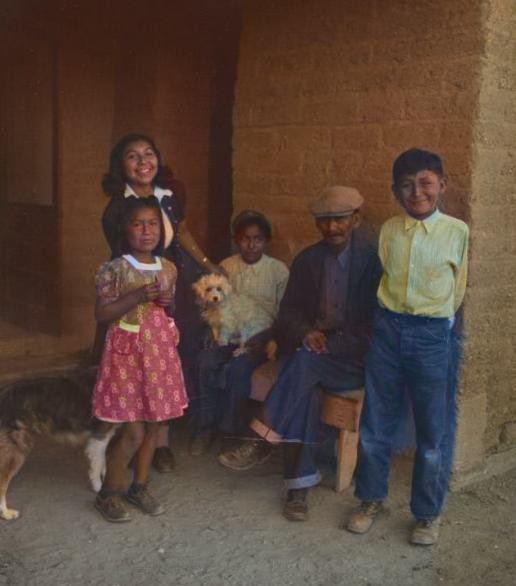} &
\includegraphics[width=\wwa,frame]{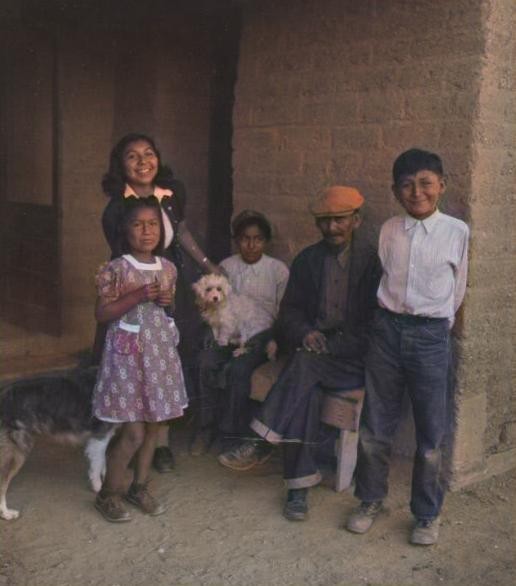} &
\includegraphics[width=\wwa,frame]{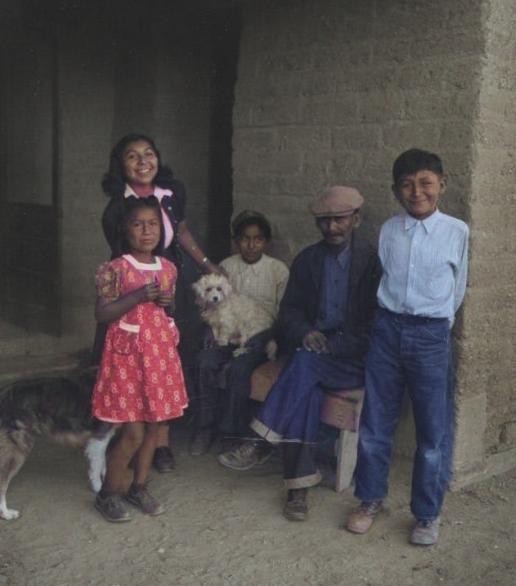} &
\includegraphics[width=\wwa,frame]{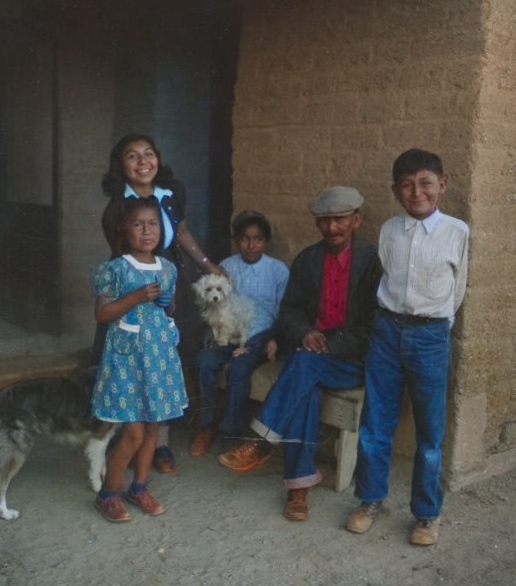}
\\

\includegraphics[width=\wwa,frame]{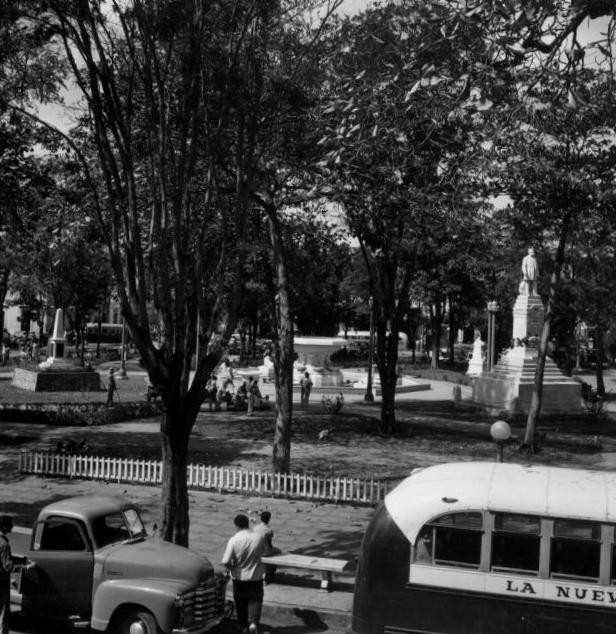} &
\includegraphics[width=\wwa,frame]{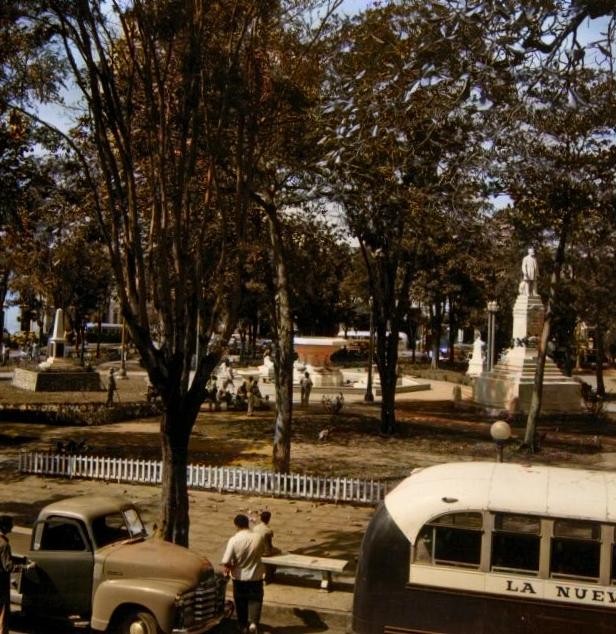} &
\includegraphics[width=\wwa,frame]{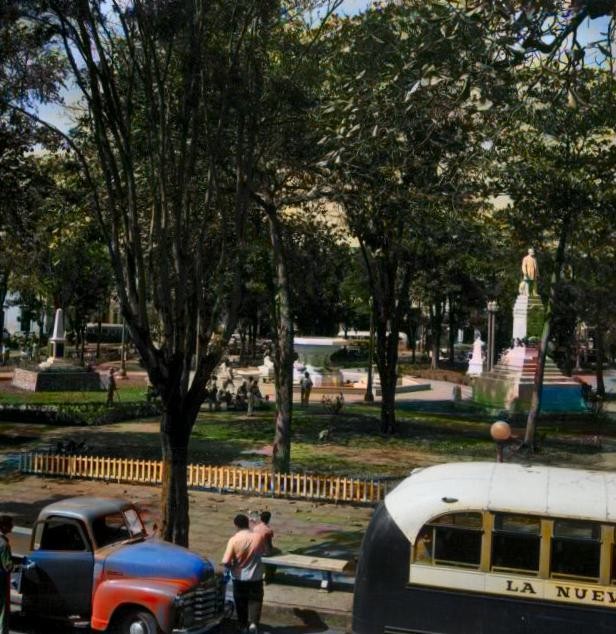} &
\includegraphics[width=\wwa,frame]{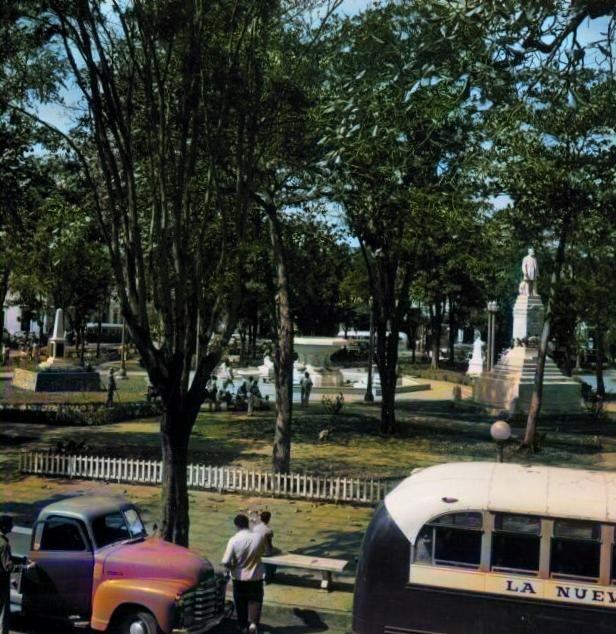} &
\includegraphics[width=\wwa,frame]{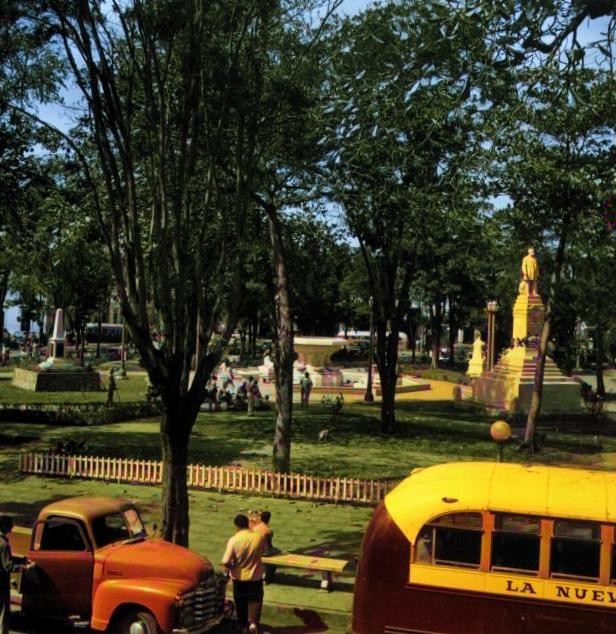}  &
\includegraphics[width=\wwa,frame]{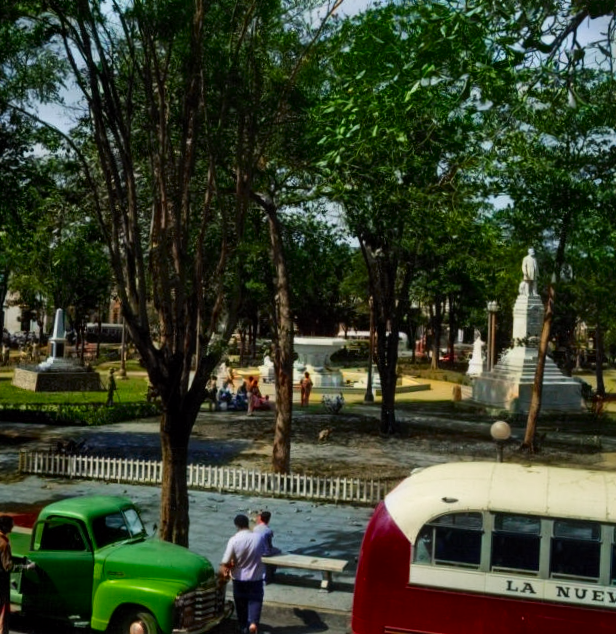}
\\

\includegraphics[width=\wwa,frame]{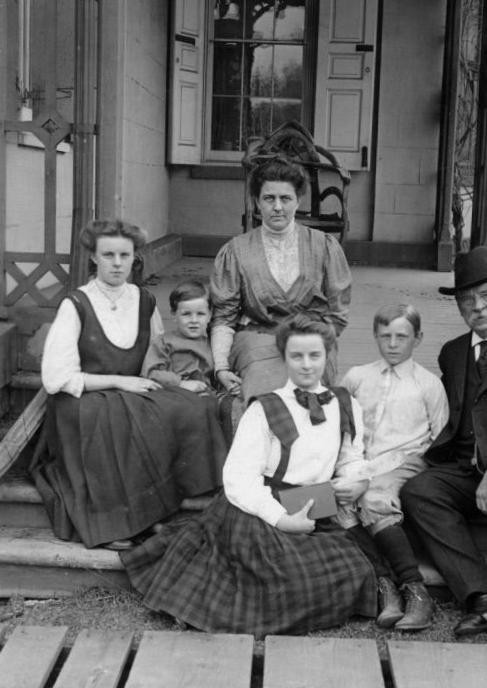} &
\includegraphics[width=\wwa,frame]{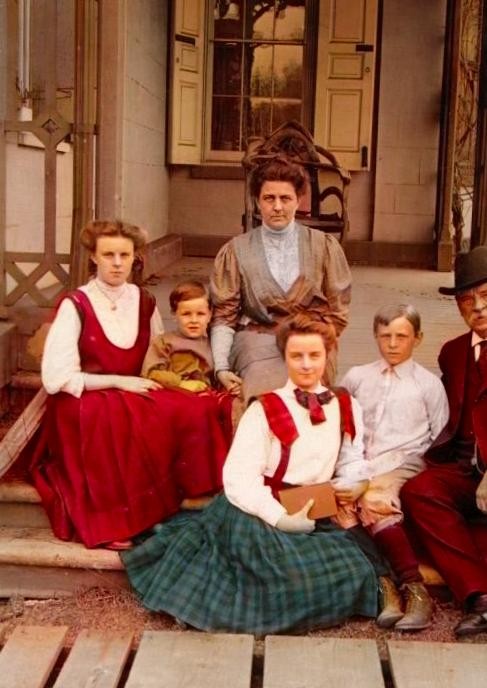} &
\includegraphics[width=\wwa,frame]{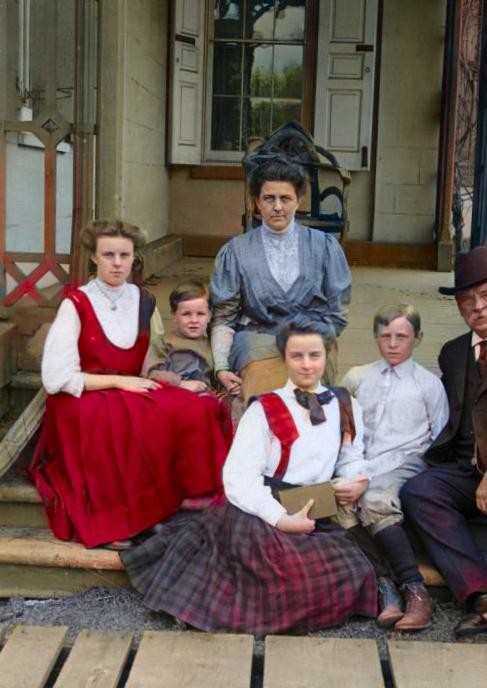} &
\includegraphics[width=\wwa,frame]{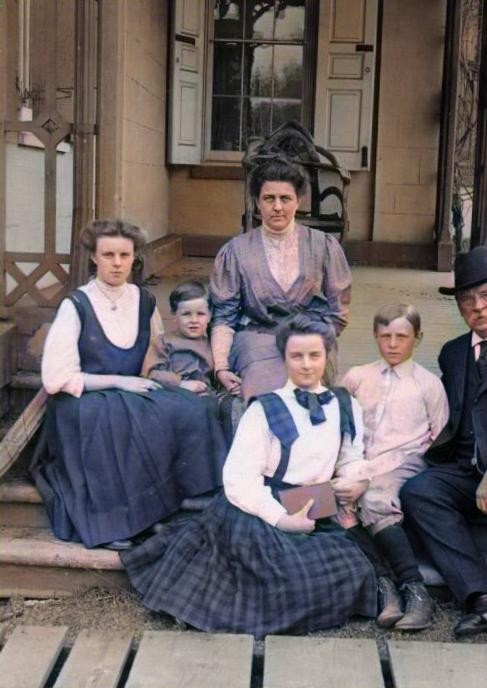} &
\includegraphics[width=\wwa,frame]{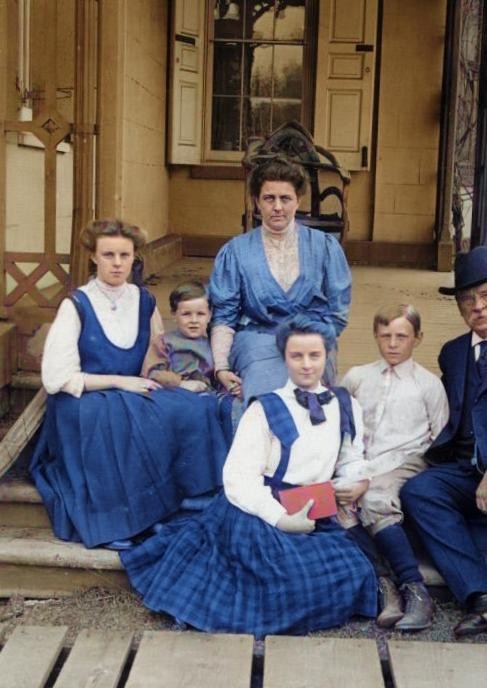} &
\includegraphics[width=\wwa,frame]{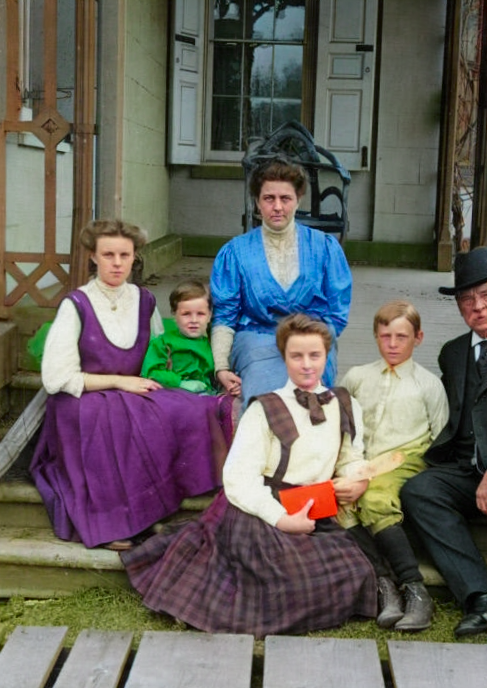}
\\

 (a) Input &  (b) UniColor &(c)  BigColor &(d)  DeOldify &(e)  Disco & (f) Ours \\ 

\end{tabular}

\caption{
\textbf{Visual comparison on legacy photos.} Images sourced from the KeystoneDepth dataset.
}
\label{fig:legacy_qualitative_comparison}
\end{figure*}

\end{document}